\documentclass{article}

\usepackage{sections/general}

\usepackage{natbib}
\usepackage[margin=1in]{geometry}
\usepackage[utf8]{inputenc} 
\usepackage[T1]{fontenc}    
\usepackage{hyperref}       
\usepackage{url}            
\usepackage{booktabs}       
\usepackage{amsfonts}       
\usepackage{nicefrac}       
\usepackage{microtype}      
\usepackage{xcolor}         
\usepackage{titletoc}

\usepackage{autonum}

\title{SGD Finds then Tunes Features in Two-Layer Neural Networks with Near-Optimal Sample Complexity: A Case Study in the XOR problem}

\author{%
  Margalit Glasgow \\
  Department of Computer Science\\
  Stanford University\\
  \texttt{mglasgow@stanford.edu} \\
}

\begin{document}

\maketitle

\begin{abstract}
    In this work, we consider the optimization process of minibatch stochastic gradient descent (SGD) on a 2-layer neural network with data separated by a quadratic ground truth function. We prove that with data drawn from the $d$-dimensional Boolean hypercube labeled by the quadratic ``XOR'' function $y = -x_ix_j$, it is possible to train to a population error $o(1)$ with $d \pl(d)$ samples. Our result considers simultaneously training both layers of the two-layer-neural network with ReLU activations via standard minibatch SGD on the logistic loss. To our knowledge, this work is the first to give a sample complexity of $\tilde{O}(d)$ for efficiently learning the XOR function on isotropic data on a standard neural network with standard training. Our main technique is showing that the network evolves in two phases: a \em signal-finding \em phase where the network is small and many of the neurons evolve independently to find features, and a \em signal-heavy \em phase, where SGD maintains and balances the features. We leverage the simultaneous training of the layers to show that it is sufficient for only a small fraction of the neurons to learn features, since those neurons will be amplified by the simultaneous growth of their second layer weights.
\end{abstract}

\section{Introduction}
Stochastic gradient descent (SGD) is the primary method of training neural networks in modern machine learning. Despite the empirical success of SGD, there are still many questions about why SGD is often able to efficiently find good local minima in the non-convex optimization landscape characteristic of training neural networks. 

A growing body of work aims to theoretically understand the optimization dynamics and sample complexity of learning natural classes of functions via SGD on neural networks. A particularly well-understood regime in this regard is the neural tangent kernel (NTK)\citep{jacot2021neural}, where the network only moves a small distance from its initialization. However, in many cases, the NTK provably requires a poor sample complexity to generalize~\citep{abbe2022merged}.



More recent work aims to prove convergence guarantees for SGD on neural networks with tight sample complexity guarantees. A natural test-bed for this, which has garnered a lot of attention, is learning target functions that are inherently low-dimensional, depending only on a constant number of dimensions of the data~\citep{chen2020learning, chen2020towards, nichani2022identifying, barak2022hidden, bietti2022learning, mousavi2022neural, refinetti2021classifying, abbe2021staircase, abbe2022merged, abbe2023sgd}. Such functions, often called \em sparse \em or \em multi-index \em functions, can be written as $f(x) := g(Ux)$, where $U \in \mathbb{R}^{k \times d}$ has orthogonal rows, and $g$ is a function on $\mathbb{R}^k$. 
Many works have shown that learning such target functions via SGD on neural networks is possible in much fewer samples than achievable by kernel methods \citep{chen2020towards, bai2019beyond, damian2022neural, abbe2021staircase, abbe2022merged, abbe2023sgd}. The results in these papers apply to a large class of ground truth functions, and have greatly enhanced our understanding of the sample complexity necessary for learning via SGD on neural networks. 

The limitation of the aforementioned works is that they typically modify the SGD algorithm in ways that don't reflect standard training practices, for example using layer-wise training, changing learning rates, or clipping. While providing strong guarantees on certain subclasses of multi-index functions, such modifications may limit the ability of SGD to learn broader classes of multi-index functions with good sample complexity. We discuss this more in the context of related work in Section~\ref{sec:relatedwork}.

The goal of this paper is to show that for a simple but commonly-studied problem, standard minibatch SGD on a two-layer neural network can learn the ground truth function in near-optimal sample complexity. In particular, we prove in Theorem~\ref{thm:main} that a polynomial-width ReLU network trained via online minibatch SGD on the logistic loss will classify the boolean XOR function $f(x) := -x_i x_j$ with a sample complexity of $\tilde{O}(d)$.\footnote{We consider this near-optimal in the sense that for algorithms that are rotationally invariant, $\tilde{\Theta}(d)$ samples are required. See Section~\ref{sec:stat} for details.} We study the XOR function because it one of the simplest test-beds for a function which exhibits some of the core challenges of analyzing SGD on neural networks: a random initialization is near a saddle point, 
and the sample complexity attainable by kernel methods is suboptimal (see further discussion in Section~\ref{sec:relatedwork}). 

Despite its simplicity, the prior theoretical understanding of learning the XOR function via SGD on standard networks is lacking. It is well-known that the NTK requires $\Theta(d^2)$ samples to learn this function~\citep{wei2019regularization, ghorbani2021linearized, abbe2023sgd}. \citet{wei2019regularization} showed that $\tilde{O}(d)$ samples statistically suffice, either by finding the global optimum of a two-layer network, or by training an infinite-width network, both of which are computationally intractable. Similar guarantees of $\tilde{O}(d)$ are given by \citet{bai2019beyond} and \citet{chen2020towards}; however, such approaches rely on drastically modifying the network architecture and training algorithm to achieve a quadratic neural tangent kernel. \citet{abbe2023sgd} proves a sample complexity of $\tilde{O}(d)$ for the XOR problem, but uses an algorithm which assumes knowledge of the coordinate system under which the data is structured, and is thus not rotationally invariant. It is also worth noting that several works have studied the XOR problem with non-isotropic data, where the cluster separation grows to infinity~\citep{frei2022random, ben2022high}, in some cases yielding better sample complexities. 

The main approach in our work is showing that while running SGD, the network naturally evolves in two phases. In the first phase, which we call the \em signal-finding \em phase, the network is small, and thus we can show that a sufficient fraction of the neurons evolve independently, similarly to how they would evolve if the output of the network was zero. Phase 1 is challenging because it requires moving away from the saddle near where the network is initialized, which requires super-constant time (here we use ``time'' to mean the number of iterations times step size). This rules out using the mean field model approach as in \citet{mei2018mean, mei2019mean}, or showing convergence to a lower-dimensional SDE as in \citet{ben2022high}, which both break down after constant time when directly applied to our setting. 

After the signal components in the network have become large enough to dominate the remaining components, the network evolves in what we call the \em signal-heavy \em phase. In this phase, we show inductively that throughout training, the signal components stay significantly larger than their counterparts. This inductive hypothesis allows us to approximate the output of the network on a sample $x$ by its \em clean \em approximation, given by a network where all the non-signal components have been removed. Under this approximation, the dynamics of the network are easier to compute, and we can show that the signal components will grow and rebalance until all four of the clusters in the XOR problem have sufficiently small loss.  

Our Phase 2 analysis leverages the simultaneous training of both layers to show that the dominance of the signal components will be maintained throughout training. In particular, we show once individual neurons become signal heavy, their second layer weights become large, and thus a positive feedback cycle between the first and second layer weights of that neuron causes it to grow faster than non-signal-heavy neurons. This allows us to maintain the signal-heavy inductive hypothesis. If we only trained the first layer, and all second layer weights had equal absolute value, then unless we have strong control over the balance of the clusters, it would be possible for the non-signal components to grow at a rate which is on the same order as the rate of the signal components (see Remark~\ref{sec:simul}).

\subsection{Related Work}\label{sec:relatedwork} 
\paragraph{Learning Multi-Index Functions via Neural Networks}
Most related to our work is a body of work aiming to understand the sample complexity of learning multi-index functions via SGD on neural networks~\citep{bietti2022learning, refinetti2021classifying, chen2020towards, abbe2021staircase, abbe2022merged, abbe2023sgd, damian2022neural, barak2022hidden, daniely2020learning, mousavi2022neural, nichani2022identifying, ge2017learning, mahankali2023beyond}. Such functions are typically studied in either the Gaussian data setting where $x \sim \mathcal{N}(0, I_d)$, or in the Boolean hypercube setting, where $x \sim \on{Uniform}(\{\pm 1\}^d)$. In both cases, we have $f(x) := g(Ux)$, where $U$ projects $x$ onto a lower dimensional space of dimension $k$, and $g$ is an arbitrary function on $k$ variables. In the Boolean setting, $U$ projects onto a subset of $k$ coordinates of $x$, so in the case of the XOR function we study, $k = 2$ and $g$ is a quadratic function. 

Chen and Meka~\citep{chen2020learning} showed when $k$ is constant, and $g$ is a degree-$D$ polynomial for constant $D$, there exists a polynomial-time algorithm which learns such multi-index functions on Gaussian covariates in $\tilde{O}(d)$ samples. Such algorithms can also be emulated in the same sample complexity via SGD on neural networks designed to emulate arbitrary Statistical Query algorithms~\citep{abbe2020universality, abbe2021power}, though these networks bear little similarity to standard neural networks used in practice. 

The sample complexity of learning multi-index functions via SGD on standard neural networks is an open and active area of research. It is known that the neural tangent kernel (and more generally, kernel methods) require $\Omega(d^D)$ samples~\citep{hsu2021dimension}. A line of work by Abbe et al.~\citep{abbe2021staircase, abbe2022merged, abbe2023sgd} has conjectured that the sample complexity required for SGD is $\tilde{\Theta}(d^{\max(L - 1, 1)})$, where $L$ denotes the ``leap complexity'', a measure of hierarchical structure upper bounded by $D$, and which equals $2$ for the XOR function. If true, this conjecture would place the sample complexity of SGD on standard neural networks squarely between that of kernel methods and arbitrary polynomial-time algorithms. When $L = 1$, \citet{abbe2022merged} showed via a mean-field analysis that is possible to learn with $\Theta(d)$ samples via layer-wise training, where the first layer is trained until it learns the subspace $U$, and then the second layer is trained as a linear model. For $L > 1$, \citet{abbe2023sgd} provided a layer-wise SGD algorithm achieving the conjectured complexity, but which assumes knowledge of the coordinate system under which the data is structured. This means the algorithm is not-rotationally invariant, barring the network from learning more general multi-index functions. Other works have also used layer-wise training to give similar results for subclasses of multi-index functions~\citep{damian2022neural, mousavi2022neural, barak2022hidden}; \citet{mousavi2022neural} studies a setting where $k = 1$ and $L = 1$, while \citet{damian2022neural, barak2022hidden} study settings where $L \geq 2$, and use just a single gradient step on on the first layer, which requires $\Omega(d^{L})$ samples. Numerous other works~\citep{tan2019online, bietti2022learning,  wu2023learning} have made progress in the setting of single-index functions ($k = 1$) when $L > 1$, in some cases achieving tight guarantees that depend on a quantity called the ``information exponent'' of $g$, though these methods require training only a single neuron in $\mathbb{R}^d$. The recent work \citet{mahankali2023beyond} considers training a single-index target function with $k = 2$ and degree $4$ on a 2-layer neural network via vanilla gradient descent, and shows a sample complexity of $O(d^{3 + \epsilon})$, which improves over kernel methods.

The above discussion highlights a gap in our understanding when $k \geq 2$ and $L \geq 2$. Indeed, such a setting is challenging because it requires learning multiple neurons, and escaping one (or more) saddles~\citep{abbe2023sgd}. For this reason, we believe the XOR function (with $k, L = 2$) is a good stepping stone for understanding the behaviour of SGD on neural networks for more general functions with $k \geq 2, L \geq 2$. We note that several other works~\citep{bai2019beyond, chen2020towards} have achieved a near-optimal sample complexity of $\tilde{O}(d)$ for the XOR problems; these works use a non-standard architecture and training algorithm which puts SGD into a quadratic NTK regime. While such a regime can often attain sample complexities beating the standard (linear) NTK, in general this method yields complexities of $\tilde{O}(d^{D-1})$, which is larger than the rate achieved by \citet{abbe2022merged} whenever $L = 1$ and $D \geq 3$. We emphasize that our work achieves the near-optimal sample complexity $\tilde{O}(d)$ with a standard two-layer neural network, trained with standard minibatch SGD. 

We note that many more works have explored both empirically~(eg. \citep{woodworth2020kernel, chizat2019lazy}) and theoretically~(eg.\citep{li2020learning, allen2020backward, suzukibenefit, telgarsky2022feature}) the sample-complexity advantages of ``rich'' SGD training over the ``lazy'' NTK regime. 

\paragraph{Simultaneous Training of Layers.}
While many of the works mentioned above use layer-wise training algorithms, the standard empirical practice is to train all layers simultaneously. Several theoretical works explore this setting, uncovering implicit biases of ReLU (or other homogeneous) networks trained simultaneously~\citep{wei2019regularization, chizat2020implicit, lyu2019gradient, lyu2021gradient, maennel2018gradient}. Under a variety of assumptions, these works have related the solutions found via gradient descent to margin-maximizing solutions. A much finer understanding of the implicit bias of simultaneous training is provided for a line of work on diagonal neural networks~\citep{pesme2023saddle, even2023s}.

\subsection{Organization of Paper}
In Section~\ref{sec:setup}, we formally describe the data and training model we study. In Section~\ref{sec:results} we state our result. In Section~\ref{sec:overview}, we give an overview of proof technique. In Section~\ref{sec:discussion}, we discuss the limitations of our work, takeaways, and open questions. All proofs are given in the Appendix.

\subsection{Notation}
For a vector $v$, we use $\|v\|$ to denote the $\ell_2$ norm, and $\|v\|_1$ to denote the $\ell_1$ norm. We use $\|M\|_2$ to denote the spectral norm of a matrix $M$. All big-O notation is with respect to $d \rightarrow \infty$, and we use $\tilde{O}$ to suppress log factors in big-O notation. $\omega(1)$ denotes growing to infinity with $d$. We use $\mathbb{S}^{d-1}(r)$ to denote the sphere of radius $r$ in $d$ dimensions, and $\mathbf{1}(\cdot)$ to denote the indicator variable of an event.

\section{Model and Setting}\label{sec:setup}

\subsection{Data.}\label{sec:data}

We study the setting where the data comes from the Boolean hypercube $x \sim \text{Uniform}(\{-1, 1\}^d)$, and the label $y$ is given by $y(x) = \on{XOR}(x_1, x_2) := -x_1x_2$.

Note that with $\mu_1 := e_1 - e_2$, and $\mu_2 := e_1 + e_2$, we can model the distribution as

\begin{equation}
    (x, y) = \begin{cases}(\mu_1 + \xi , 1) & w.p. \: 1/4 \qquad (-\mu_1 + \xi , 1) \quad \:\:\: w.p. \: 1/4\\
    (\mu_2 + \xi , -1)  & w.p. \: 1/4 \qquad (-\mu_2 + \xi , -1)  \quad w.p. \: 1/4
    \end{cases},
\end{equation}
where $\xi \sim \text{Uniform}(0^2\times\{-1, 1\}^{d-2})$ so that $\xi \perp \{\mu_1, \mu_2\}$.  We will often write
\begin{equation}
    x = z + \xi,
\end{equation}
where $z$ is the projection of $x$ onto the space spanned by $e_1$ and $e_2$, and $\xi$ is the projection of $x$ orthogonal to $e_1$ and $e_2$. We denote this distribution by $P_d$, and throughout, it is implicitly assumed that all probabilities and expectations over $x$ are for $x \sim P_d$.

\begin{remark}
While for simplicity, we  state our results for the setting where the data comes from an axis-aligned Boolean hypercube, and where ground truth depends on the first two dimensions, the minibatch SGD algorithm and the initialization of the network will be  rotationally invariant. Thus all our results hold for a Boolean hypercube with any basis.
\end{remark}

\subsection{Training.}\label{sec:training}
\paragraph{Model.}
We train both layers of a two-layer ReLU network with width $p$:
\begin{align}
    \frac{1}{p}\sum_{j = 1}^p a_j\sigma(w_j^Tx),
\end{align}
where $\sigma(\alpha) = \max(0, \alpha)$ is the ReLU function. We will use the variable $\rho := \frac{1}{p}\sum_{j = 1}^p \mathbf{1}_{(w_j, a_j)}$ to denote the empirical distribution of the neurons and their second layer weights. Thus we denote
\begin{align}
       f_{\rho}(x) := \mathbb{E}_{w, a \sim \rho} a \cdot \sigma(w^Tx),
\end{align}

We will often abuse notation and write probabilities and expectations using $w \sim \rho$, and use $a_w$ to denote its associated second layer weight. We note that it is not necessarily the case the second layer weight $a_w$ is a \em function \em of $w$; we do this for the convenience of not indexing each pair as $(w_j, a_j)$. 

\paragraph{Initialization.} We initialize the network with $w_j \sim \on{Uniform}(\mathbb{S}^{d - 1}(\theta))$ for a scale parameter $\theta$, such that $\|w_j\| = \theta$. We initialize the second layer as $a_j = \epsilon_j \|w_j\|,$ where $\epsilon_j \sim \on{Uniform}(\pm 1)$.

\paragraph{Minibatch SGD.} We train using minibatch SGD on the logistic loss function
\begin{align}
    \ell_{\rho}(x) := -2\log\left(\frac{1}{1 + \exp(-y(x)f_{\rho}(x))}\right),
\end{align}
and define the population loss $L_{\rho} := \mathbb{E}_{x \sim P}\ell_{\rho}(x)$.
We will use the shorthand $\ell'_{\rho}(x)$ to denote the derivative of $\ell_{\rho}(x)$ with respect to $f_{\rho}(x)$:
\begin{align}
    \ell'_{\rho}(x) := -\frac{2y(x)\exp(-y(x)f_{\rho}(x))}{1 + \exp(-y(x)f_{\rho}(x))}.
\end{align}

We use $\rho_t$ to denote the empirical distribution of the $p$ neurons $(w^{(t)}, a_w^{(t)})$ at iteration $t$. At each step, we perform the minibatch SGD update
\begin{align}
    w^{(t + 1)} := w^{(t)} - \eta \nabla \hat{L}_{\rho}(w^{(t)}) \qquad  a_w^{(t + 1)} := a_w^{(t)} - \eta \nabla \hat{L}_{\rho}(a_w^{(t)}).
\end{align}
Here $\hat{L}_{\rho} = \frac{1}{m} \sum_{x^{(i)} \in M_t}\ell_{\rho}(x^{(i)})$ denotes the empirical loss with respect to a minibatch $M_t$ of $m$ random samples chosen i.i.d. from $P_d$ at step $t$, and for a loss function $L$ and a parameter $u$ in the network, $\nabla_u L := p\frac{\partial L}{\partial u}$ denotes the scaled partial derivative of the loss with respect to $u$, defined in particular for a neuron $(w, a_w)$, as follows: \footnote{Since the ReLU function is non-differentiable at zero, we define $\sigma'(0) = 0$. }\footnote{For convenience, we scale this derivative up by a factor of $p$ to correspond to the conventional scaling in the mean-field model. Of course if we didn't perform this scaling, we would achieve the same result by scaling the learning rate $\eta$.}
\begin{align}
    \nabla_w \hat{L}_{\rho} &= \frac{1}{m}\sum_{x^{(i)} \in M_t}\frac{\partial}{\partial w} p \ell_{\rho}(x^{(i)}) = \frac{1}{m}\sum_{x^{(i)}\in M_t} a_w \ell'_{\rho_t}(x^{(i)})\sigma'(w^Tx^{(i)})x^{(i)}; \\
    \nabla_{a_w} \hat{L}_{\rho} &= \frac{1}{m}\sum_{x^{(i)}\in M_t}\frac{\partial}{\partial a_w} p \ell_{\rho}(x^{(i)}) = \frac{1}{m}\sum_{x^{(i)} \in M_t}\ell'_{\rho_t}(x^{(i)})\sigma(x_i^Tw).
\end{align}

\section{Main Result}\label{sec:results}

The following theorem is our main result. 
\begin{theorem}\label{thm:main}
There exists a constant $C > 0$ such that the following holds for any $d$ large enough. Let $\theta := 1/\log^C(d)$. Suppose we train a 2-layer neural network with minibatch SGD as in Section~\ref{sec:training} with a minibatch size of $m \geq d/\theta$, width $1/\theta \leq p \leq d^C$, step size $d^{-C} \leq \eta \leq \theta$, and initialization scale $\theta/\sqrt{p}$. Then for some $t \leq C\log(d)/\eta$, with probability $1 - d^{-\omega(1)}$, we have
\begin{align}
    \mathbb{E}_{x \sim P_d}[\ell_{\rho_t}(x)] \leq o(1).
\end{align}
\end{theorem}
By setting $\eta = \theta$ and $m = d/\theta$, Theorem~\ref{thm:main} states that we can learn the XOR function up to $\eps$ population loss in $\Theta\left(d 
 \pl(d)\right)$ samples and iterations on a polynomial-width network.

\section{Proof Overview}\label{sec:overview}
Throughout the following section, and in our proofs, we will use the following shorthand to refer to the components of a neurons $w$. We decompose $w = w_{1:2} + \wpe$, where $w_{1:2}$ is the projection of $w$ in the direction spanned $e_1$ and $e_2$ (and equivalently by $\mu_1 = e_1 - e_2$ and $\mu_2 = e_1 + e_2$), and $\wpe$ is the component of $w$ in the orthogonal subspace. We further decompose $w_{1:2} = \wsig + \wopp$ as follows:
\begin{align}
    \wsig = \begin{cases}\frac{1}{2}\mu_1\mu_1^Tw & a_w \geq 0; \\
    \frac{1}{2}\mu_2\mu_2^Tw & a_w < 0.\end{cases}
    \qquad \wopp = \begin{cases}\frac{1}{2}\mu_2\mu_2^Tw & a_w \geq 0;\\
    \frac{1}{2}\mu_1\mu_1^Tw & a_w < 0.\end{cases}
\end{align}
Intuitively, we want the neurons to grow in the $\wsig$ direction, but not the $\wopp$ direction; in a network achieving the maximum normalized margin, we will have $w = \wsig$ exactly, and $\wopp = \wpe = 0$. 
We summarize this notation in Table~\ref{table:notation}, along with future shorthand we will introduce in this section. 
\begin{table}
  \caption{Summary of Notation used in Proof Overview and Proofs}
  \label{table:notation}
  \centering
  \begin{tabular}{c|c|c}
    \toprule
    $\wsig = \begin{cases}\frac{1}{2}\mu_1\mu_1^Tw & a_w \geq 0 \\
    \frac{1}{2}\mu_2\mu_2^Tw & a_w < 0\end{cases}$      &
    $\wopp = \begin{cases}\frac{1}{2}\mu_2\mu_2^Tw & a_w \geq 0\\
    \frac{1}{2}\mu_1\mu_1^Tw & a_w < 0\end{cases}$      &
    $\begin{cases}w_{1:2} = \wsig + \wopp\\ \wpe = w - w_{1:2}\end{cases}$      \\
    \midrule
    $\bmu = f_{\rho}(\mu)y(\mu)$ &
    $\bmin = \min_{\mu \in \{\pm \mu_1, \pm \mu_2\}}\bmu$ & $\bmax = \max_{\mu \in \{\pm \mu_1, \pm \mu_2\}}\bmu$\\
    \midrule
    $\gmu = |\ell'_{\rho}(\mu)|$ &
    $\gmin = \min_{\mu \in \{\pm \mu_1, \pm \mu_2\}}|\ell'_{\rho}(\mu)|$ & $\gmax = \max_{\mu \in \{\pm \mu_1, \pm \mu_2\}}|\ell'_{\rho}(\mu)|$\\
\bottomrule
  \end{tabular}
\end{table}

The main idea of our proof is to break up the analysis of SGD into two main phases. In the first phase, the network is small, and thus we have (for most $x$) that the loss $\ell_{\rho}(x)$ is well approximated by a first order approximation of the loss at $f_{\rho} = 0$, namely 
\begin{align}
    \ell_0(x; \rho) := -2\log(1/2)-y(x)f_{\rho}(x).
\end{align}
As long as this approximation holds, the neurons of the network evolve (approximately) independently, since $\ell'_0(x) := \frac{\partial \ell_0(x; \rho)}{\partial f_{\rho}(x)}= -y(x)$ does not depend on the full network $\rho$. We will show under this approximation that for many neurons, $\|\wsig\|$ grows exponentially fast. Thus we will run this first phase for $\Theta(\log(d)/\eta)$ iterations until for all four clusters $\mu \in \{\pm \mu_1, \pm \mu_2\}$, there exists a large set of neurons $S_{\mu}$ on which $\wsig^T\mu > 0$, and the ``margin'' from this set of neurons is large, ie.
\begin{align}\label{eq:simple_invariant}
    \hmu := \mathbb{E}_{\rho}[\mathbf{1}(w \in S_{\mu})a_w \sigma(w^T\mu)] \gg \mathbb{E}_{\rho}\|\wpe + \wopp\|^2.
\end{align}


In the Phase 2, we assume that Eq.~\ref{eq:simple_invariant} holds, and we leverage the dominance of the signal to show that (1) The signal components $\wsig$ grow faster that $\wopp + \wpe$, and thus Eq.~\ref{eq:simple_invariant} continues to hold; and (2) SGD balances the signal components in the 4 cluster directions such that the margins $\hmu$ balance, and become sufficiently large to guarantee $o(1)$ loss.

We proceed to describe the analysis in the two phases in more detail. Full proofs are in the Appendix.
\subsection{Phase 1}
In Phase 1, we approximate the evolution of the network at each gradient step by the gradient step that would occur for a network with output $0$. The main building blocks of our analysis are estimates of the $L_0 := \mathbb{E}_x \ell_0(x; \rho)$ population gradients, and bounds on the difference $\nabla L_0 - \nabla L_{\rho}$.

\paragraph{$L_0$ population gradients.} Since the primary objective of this phase is to grow the neurons in the signal direction, we sketch here the computation of the gradient $\nabla_{w_{1:2}} L_{0}$ in the subspace spanned by $\mu_1, \mu_2$. The remaining estimates of $\nabla L_0$ are simpler, and their main objective is to show that $\nabla_{\wpe} L_{0}$ and $\nabla_{a_w} L_{0}$ are sufficiently small, such that $\|\wpe\|$ doesn't change much throughout Phase 1, and $|a_w|$ stays approximately the same as $\|w\|$. For convenience, the reader may assume that $|a_w| = \|w\|$ exactly.\footnote{When $\eta \rightarrow 0$ as in gradient flow, this equivalence holds exactly for ReLU networks, as long as the initialization satisfies $|a_w| = \|w\|$.},

For a data sample $x \sim \Rad^d$, we denote $x = z + \xi$, where $z \in \on{Span}(\muset)$, and $\xi \perp \on{Span}(\muset)$. By leveraging the symmetry of the data distribution and the fact that $y(z) = y(-z)$, we can compute
\begin{equation}\label{eq:gradwsig}
\begin{split}
    \nabla_{w_{1:2}} L_{0} &= -a_w\mathbb{E}_{x = z + \xi} y(x)\sigma'(w^Tx)z\\
    &=  -a_w\mathbb{E}_{\xi}\frac{1}{2}\mathbb{E}_{z} y(z)\left(\sigma'(w^T\xi + w^Tz) - \sigma'(w^T\xi - w^Tz)\right)z\\
    &= -a_w\mathbb{E}_{\xi}\frac{1}{2}\mathbb{E}_{z} y(z)\mathbf{1}(|w^Tz| \geq |w^T\xi|)\on{sign}(w^Tz)z \\
    &= -\frac{1}{2}a_w\mathbb{E}_{z}y(z)\on{sign}(w^Tz)z \mathbb{P}_{\xi}[|w^Tz| \geq |w^T\xi|]\\
    &\approx -\frac{1}{2}a_w\mathbb{E}_{z}y(z)\on{sign}(w^Tz)z \mathbb{P}_{G \sim \mathcal{N}(0, \|\wpe\|^2)}[G \leq |w^Tz|]\\
    &\approx  -\frac{1}{2}a_w\mathbb{E}_{z}y(z)\on{sign}(w^Tz)z \sqrt{\frac{2}{\pi}}\frac{|w^Tz|}{\|w\|}.
\end{split}
\end{equation}
Here the two approximations come from the fact that $\xi$ has boolean coordinates and not Gaussian, and from an approximation of the Gaussian distribution, which holds whenever $\frac{|w^Tz|}{\|\wpe\|}$ is small. By taking the expectation over $z \in \{\pm \mu_1, \pm \mu_2\}$, the last line of Eq~\ref{eq:gradwsig} can be shown to evaluate to
\begin{align}\label{eq:gradwsig2}
-\frac{|a_w|}{\|w\|\sqrt{2\pi}}\wsig + \frac{|a_w|}{\|w\|\sqrt{2\pi}}\wopp.
\end{align}
Observe that near initialization, this gradient is quite small, since $\frac{\|\wsig\|}{\|w\|}$ is approximately $\frac{1}{\sqrt{d}}$ for a random initialization. Nevertheless, this gradient suggests that $\wsig$ will grow exponentially fast.


\paragraph{Bounding the difference $\nabla L_0 - \nabla L_{\rho}$.}
To bound $\|\nabla_w L_{\rho} -  \nabla_w L_0\|_2$, first recall that 
\begin{align}
   \nabla_w L_{0} - \nabla_w L_{\rho} &= \mathbb{E}_x a_w(\ell'_{\rho}(x) - \ell'_0(x))\sigma'(w^Tx)x.
\end{align}
Defining $\Delta_x := (\ell'_{\rho}(x) - \ell'_0(x))\sigma'(w^Tx)$,  we can show using routine arguments (see Lemma~\ref{lemma:graddiff1} for the details) that:
\begin{align}\label{eq:weak_approx}
    \|\nabla_w L_{\rho} -  \nabla_w L_0\|_2 = |a_w|\|\mathbb{E}_x \Delta_x x\| &\leq |a_w|\sqrt{\mathbb{E}_x \Delta_x^2} \\
    &\approx |a_w|\sqrt{\mathbb{E}_x f_{\rho}(x)^2} \\
    &\lessapprox |a_w|\mathbb{E}_{\rho}[\|a_w w\|] \approx \frac{|a_w|}{\pl(d)}.
\end{align}

While this deviation bound is useful for showing that $\wpe$ doesn't move too much, this bound far exceeds the scale of the gradient in the $\wsig$, which is on the scale $\frac{|a_w|}{\sqrt{d}}$ near initialization. Fortunately, we can show in Lemma~\ref{lemma:approxsig} that the deviation is much smaller on the first two coordinates, namely,
\begin{align}\label{eq:strong_approx}
    \|\nabla_{w_{1:2}} L_{\rho} -  \nabla_{w_{1:2}} L_{0}\|_2 \leq |a_w|O(\log^2(d))\left(\mathbb{E}_{\rho}[\|a_w w_{1:2}\|] + \mathbb{E}_{\rho}[\|a_w w\|] \frac{\|w_{1:2}\|}{\|w\|}\right)
\end{align}
Note that since near initialization $\|w_{1:2}\| \ll \|w\|$ for all neurons, this guarantee is much stronger than Eq.~\ref{eq:weak_approx}. In fact, since throughout this phase we can show that $a_w$ and $\|w\|$ change relatively little, staying at the scale $1/\pl(d)$, the approximation error in Eq.~\ref{eq:strong_approx} is smaller than the gradient in the $\wsig$ direction (Eq.~\ref{eq:gradwsig2}) whenever say $\|\wsig\| \geq 100\mathbb{E}_{\rho}[\|a_w w_{1:2}\|]$, which occurs on a substantial fraction of the neurons. 

Lemma~\ref{lemma:approxsig} is the most important lemma in our Phase 1 analysis. At a high level, it shows that the approximation error $\|\nabla_{w_{1:2}} L_{\rho} -  \nabla_{w_{1:2}} L_{0}\|_2$ can be coupled with the growth of the signal, $-(\nabla_{w} L_0)^T\frac{\wsig}{\|\wsig\|}$. This is because we use a symmetrization trick with the pairs $z + \xi$ and $-z + \xi$ to show that both the error and the signal gradient only grow from samples $x = z + \xi$ where $|z^Tw| \geq |\xi^Tw|$.

In more detail, to prove Eq.~\ref{eq:strong_approx}, we also need to leverage the fact that for any $\xi \in \{\mu_1, \mu_2\}^{\perp}$ and $z \in \{\pm \mu_1, \pm \mu_2\}$, we have $|\ell'_{\rho}(\xi +z) - \ell'_{\rho}(\xi - z')| \leq 4p\mathbb{E}_{\rho}[\|a_w w_{1:2}\|]$, much smaller than we can expect $|\ell'_{\rho}(x) - \ell'_{0}(x)|$ to be. Thus $|\Delta_{\xi + z} -  \Delta_{\xi - z}| \leq 4p\mathbb{E}_{\rho}[\|a_w w_{1:2}\|]$ whenever $|\xi^Tw| \geq |z^Tw|$ (such that $\sigma'(w^T(\xi + z)) = \sigma'(w^T(\xi - z))$). Following the symmetrization trick in Eq.~\ref{eq:gradwsig}, we have
\begin{align}
    \left\|\frac{1}{a_w}\left(\nabla_{w_{1:2}} L_{\rho} - \nabla_{w_{1:2}} L_{0}\right)\right\| &= \|\mathbb{E}_x \Delta_x z\| \\
    &= \|\mathbb{E}_{\xi} \mathbb{E}_z \Delta_{\xi + z} z\| \\
    &= \frac{1}{2}\|\mathbb{E}_{\xi} \mathbb{E}_z (\Delta_{\xi + z} - \Delta_{\xi - z})z\| \\
    &\leq 2\sqrt{2}\mathbb{E}_{\rho}[\|a_w w_{1:2}\|] + \sqrt{2}\mathbb{E}_{\xi}\mathbb{E}_z\mathbf{1}(|\xi^Tw| \leq |z^Tw|)|\Delta_x|.
\end{align}
A careful computation comparing $w^T\xi$ to a Gaussian distribution then shows that $$\mathbb{E}_z\mathbf{1}(|\xi^Tw| \leq |z^Tw|)|\Delta_x| \approx \left(\mathbb{P}_x[|\xi^Tw| \leq |z^Tw|]\right)\left(\mathbb{E}_x|\Delta_x|\right) \lessapprox \frac{\|w_{1:2}\|}{\|w\|}\mathbb{E}_{\rho}[\|a_w w\|] .$$

\paragraph{Putting Phase 1 Together}
The building blocks above, combined with standard concentration bounds on $\nabla \hat{L}_{\rho}$, suffice to show that a substantial mass of neurons will evolve according to Eq~\ref{eq:gradwsig2}, leading to exponential growth in $\wsig$. After $\Theta(\log(d)/\eta)$ iterations, for these neurons, we can achieve $\|\wsig\| \gg \|\wpe + \wopp\|$. Formally, we show the following for some $\zeta \leq 1/\pl(d)$:
\begin{lemma}[Output of Phase 1: Informal; See Lemma~\ref{lemma:phase1formal} for formal version]\label{lemma:phase1informal}
    With high probability, for $\eta \leq \tilde{O}(1)$, after some $T = \Theta(\log(d)/\eta)$ iterations of minibatch SGD, with $m = \tilde{\Theta}(d)$ samples in each minibatch, the network $\rho_T$ satisfies:
    \begin{enumerate}
        \item $\mathbb{E}_{\rho_T}[\|\wpe + \wopp\|^2] \leq \theta$.
        \item For each $\mu \in \muset$, on at least a $0.1$ fraction of all the neurons, we have $\wsig^T\mu > 0$, and $\|\wsig\|^2 \geq \zeta^{-1}\theta$.
    \end{enumerate}
\end{lemma}
We remark that the analysis to prove Lemma~\ref{lemma:phase1informal} is somewhat subtle, since the tight approximation in Eq~\ref{eq:gradwsig} breaks down when $\|\wsig\|$ approaches $\|\wpe\|$. The details are given in Appendix~\ref{apx:phase1}.

\subsubsection{Phase 2}
The conclusion of Lemma~\ref{lemma:phase1informal} is a sufficient condition of the network to begin the second phase. In the second phase, we have that (for most $x$)
\begin{align}\label{eq:clean}
    \ell'_{\rho}(x) \approx \ell'_{\rho}(z),
\end{align}
where we recall that $z$ is the component of $x$ in the space spanned by $\mu_1$ and $\mu_2$. We refer to this as the \em clean \em loss derivative, 
and our main tool will be analyzing the evolution of SGD under this clean surrogate for the loss derivative. Namely, we define:
\begin{align}\label{eq:cldef}
      &\ncl_w L_{\rho} := a_w\mathbb{E}_{x} \ell_{\rho}'(z)\sigma'(w^Tx)x \quad \text{and} \quad \ncl_{a_w} L_{\rho} := \mathbb{E}_{x} \ell_{\rho}'(z)\sigma(w^Tx).
\end{align}

Before proceeding, we introduce the following definitions, which will be useful in Phase 2 (summarized in Table~\ref{table:notation}):
\begin{align}
    \bmin &:= \min_{\mu \in \{\pm \mu_1, \pm \mu_2\}} \bmu \qquad \gmin := \min_{\mu \in \{\pm \mu_1, \pm \mu_2\}} |\ell'_{\rho}(\mu)| = \frac{\exp(-\bmax)}{1 + \exp(-\bmax)}\\
    \bmax &:= \max_{\mu \in \{\pm \mu_1, \pm \mu_2\}} \bmu \qquad \gmax := \max_{\mu \in \{\pm \mu_1, \pm \mu_2\}} |\ell'_{\rho}(\mu)| = \frac{\exp(-\bmin)}{1 + \exp(-\bmin)}
\end{align}
To ensure the approximation in Eq.~\ref{eq:clean} holds throughout the entire the second phase, we will maintain a certain inductive hypothesis, which ensures the the scale of the signal-direction components of the network continue to dominate the scale of the non-signal-direction components of the network. Formally, we consider the following condition.
\begin{restatable}[Signal-Heavy Inductive Hypothesis]{definition}{heavydef}\label{def:heavy}
For parameters $\zeta = o(1)$ and $H > 1$ with $\zeta \leq \exp(-10H)$, we say a network is $(\zeta, H)$-\em signal-heavy \em if there exists some set of \em heavy \em neurons $S$ on which $\exp(6H)\|\wpe\| + \|\wopp\| \leq \zeta\|\wsig\|$, and 
$$\mathbb{E}_{\rho}\mathbf{1}(w \notin S)\|w\|^2 \leq \zeta \hmin.$$
Here we have defined $\hmu := \mathbb{E}[\mathbf{1}(w \in S, \wsig^T > 0)a_w\sigma(w^T\mu)]$ and $\hmin := \min_{\mu \in \muset} \hmu$.

Further, $$\mathbb{E}_{\rho}[\|w\|^2] \leq \mathbb{E}_{\rho}[|a_w|^2] + \zeta H \leq 2H,$$ and for all neurons, we have $|a_w| \leq \|w\|$.
\end{restatable}
We show via a straightforward argument in Lemma~\ref{lemma:reduction} that if the conclusion of Lemma~\ref{lemma:phase1informal} (from Phase 1)
holds for some $\zeta$, then the network is $(\Theta(\zeta^{1/3}),H)$-signal-heavy, for $H = \Theta(\log\log(d))$.

Assuming that the network is $(\zeta,H)$-signal-heavy, using a similar approach to Eq.~\ref{eq:weak_approx}, we can show (see Lemma~\ref{claim:approxerror} for the precise statement) that for any neuron $(w, a_w)$,
\begin{align}
    \frac{1}{|a_w|}\|\nabla_w L_{\rho} - \ncl_w L_{\rho} \|_2 &\lessapprox \sqrt{\mathbb{E}_x (f_{\rho}(x) - f_{\rho}(z))^2} \lessapprox  \mathbb{E}_{\rho}[\|a_w\wpe\|] \leq \zeta \bmax, 
\end{align}
and similarly $\|\nabla_{a_w} L_{\rho} - \ncl_{a_w} L_{\rho} \|_2 \lessapprox \|w\| \zeta \bmax$.

By working with the clean gradients, it is possible to approximately track (or bound) the evolution of $\wsig$, $\wpe$, and $\wopp$ on neurons in $S$, the set of neurons for which $\|\wsig\| \gg \|\wpe + \wopp\|$. In Lemmas~\ref{claim:clean}, \ref{lemma:cleanns}, and \ref{cor:clean} we show the following for any $w \in S$ (let $\mu$ be the direction of $\wsig$):

\begin{enumerate}
    \item \textbf{The signal component $\wsig$ grows quickly.} We have $-\wsig^T \ncl_w L_{\rho} \approx |a_w\ell'_{\rho}(\mu)|\tau$, where $\tau := \frac{\sqrt{2}}{4}$. Also $a_w$ grows at a similar rate. This growth is due to the fact that points with $z = -\mu$ will almost never activate the ReLU, while points with $z = \mu$ almost always will.
    \item \textbf{A linear combination of $\|\wpe\|^2$ and $\|\wopp\|^2$ decreases.} The argument here is more subtle, but the key idea is to argue that if $|\wpe^T\xi| \geq |\wopp^Tz|$ frequently, then $\|\wpe\|^2$ will decrease. Meanwhile, if $|\wpe^T\xi| \leq |\wopp^Tz|$ frequently, then $\wopp$ will decrease (and there is a sizeable event on which they both decrease).
\end{enumerate}
Since most of the mass of the network is in $S$, this shows that the signal will grow at the exponential rate $\tau |\ell'_{\rho}(\mu)|$ --- or for the ``weakest'' cluster, that is, in the direction $\mu$ that maximizes $\hmu$, we will have $\hmin^{(t + 1)}  \gtrapprox \left(1 + 2\eta \tau\gmax\right)\hmin^{(t)}$.

On neurons outside of $S$, we show in Lemma~\ref{claim:allneuron} that they grow \em at most \em as fast as the rate of the weakest clusters, meaning we can essentially ignore these neurons.

\begin{remark}\label{sec:simul}
If we did not train the second layer weights (and for instance they all had norm $1$), then our tools would not suffice to maintain the signal-heavy hypothesis in Definition~\ref{def:heavy}. Indeed, the neurons in $S$ would grow at a \em linear \em rate of $\tau |\ell'_{\rho}(\mu)|$, and at (up to) an equal linear rate outside of $S$. Thus the neurons outside of $S$ might eventually attain a non-negligible mass. However, because the layers are trained simultaneously, this leads to positive feedback between the growth of $\|\wsig\|$ and $|a_w|$, leading to exponential growth, which maintains the mass ratios between the neurons in and out of $S$.

\end{remark}

Combining the ideas above, we prove the following lemma, which shows that after one SGD step, the network stays signal-heavy (with a slightly worse parameter), the behavior of the weakest margin improves, and the network (measured by the size of the largest margin $\bmax$) doesn't become too big.
\begin{lemma}[Phase 2 Inductive Step: Informal; See Lemma~\ref{lemma:main2} for formal version]\label{lemma:main2informal}
If a network $\rho_t$ is $(\zeta, H)$-signal heavy with heavy set $S$, then after one minibatch gradient step, with probability $1 - d^{-\omega(1)}$,
\begin{enumerate}
    \item $\rho_{t + 1}$ is $(\zeta(1 + 10\eta \zeta H), H)$-signal heavy.
    \item $\hmin^{(t + 1)}  \geq \left(1 + 2\eta \tau (1 - o(1))\gmax\right)\hmin^{(t)}$
    \item $\hmax^{(t + 1)}  \leq \left(1 + 2\eta \tau (1 + o(1))\gmin\right)\hmax^{(t)}$, where $\hmax^{(t)} := \max_{\mu \in \muset} \hmu^{(t)}$.
\end{enumerate}
\end{lemma}
Theorem~\ref{thm:main} is proved by iterating this lemma for $\Theta(\log\log(d)/\eta)$ steps, yielding $\bmin \approx \hmin = \omega(1)$.
\section{Conclusion and Discussion}\label{sec:discussion}

In this work, we showed that in $\tilde{O}(d)$ samples, it is possible to learn the XOR function on Boolean data on a 2-layer neural network. Our results shows that by a careful analysis that compares that dynamics to the dyamincs under the surrogate $L_0$ loss, we can show that SGD find the signal features, and escape the region of the saddle where it was initialized. Then, after learning the feature direction, we show that SGD will enlarge and balance the signal components to learn well-classify points from all 4 clusters.

We now discuss some of the limits and possible extensions of our techniques.

\paragraph{Minibatch SGD vs SGD vs GD.}
In this work, we study minibatch SGD, with a batch size of $m \geq d \pl(d)$. This affords us enough samples at each iteration to have strong enough convergence to the population loss. Extending our results to SGD with a batch size of $1$ is an interesting open question, and it is possible that this could be achieved using the drift-martingale techniques in \citet{tan2019online, arous2021online, abbe2023sgd}. Such methods allow larger fluctuations from the population loss at each step, but show that the fluctuations concentrate over time, even when SGD is run for $T = \omega(1/\eta)$ steps, enough time to escape a saddle.

We remark that in this problem, using minibatch SGD with fresh samples can achieve stronger sample complexities than that required to show uniform convergence of the empirical gradient to the population gradient (as in \citet{ge2017learning,mei2018landscape}), which in our setting, is $\Omega(d^2)$ samples. This means proving the convergence of GD on the empirical loss would require tools beyond uniform convergence.

\paragraph{Boolean Data vs Gaussian Data.}
One limitation of this work is that our results only hold for boolean data, and not gaussian data $x \sim \mathcal{N}(0, I_d)$. As a matter of convenience, it is easier to compute the population gradients $\nabla_w L_0$ and $\ncl_w L_{\rho}$ with Boolean data, and the gradient does not depend on interactions between $\wsig$ and $\wopp$. With some willingness to perform various Gaussian integrals, we believe the analysis in Phase 1 could be extended to the Gaussian setting. This would require changing Lemma~\ref{lemma:pop1a} to reflect the population gradients, and modifying the definition of ``strong'' neurons (Def.~\ref{def:strong}) to be a more restrictive set that only includes neurons where $\|\wopp\| \ll \|\wsig\|$, such that $\wsig$ grows at the maximum possible rate. We do not know of any way to directly extend Phase 2 to the Gaussian case. This is because if the cluster margins $\gamma_u$ become very imbalanced, it is possible $\wsig$ could grow in the wrong direction.


\paragraph{Classification vs Regression.} In our classification setting, it suffices to show that the margin on each cluster grows large. We accomplish this in our Phase 2 analysis by showing that there is a large mass of neurons primarily in the $\mu$-direction for each $\mu \in \muset$. Adapting this strategy may be possible for XOR regression on Boolean data, but on Gaussian data, representing the ground truth function would require more specialization among the neurons. To see this, consider the following simpler example: to represent the single-index function $f^*(x) = (e_1^Tx)^2$ on Gaussian data on a ReLU network without biases, the neurons cannot all be oriented in the $\pm e_1$ direction, otherwise the output would be $a\sigma(x_1) + b\sigma(-x_1)$ for scalars $a, b$. Studying the power of SGD to perform this specialization is an exciting open direction. We believe that our Phase 1 analysis may be a useful first step in this regard to show that the network can become signal-heavy. More powerful techniques would need to be developed to show specialization once the network contains sufficient signal.



\bibliographystyle{plainnat}
\bibliography{ref}
\appendix
\pagebreak

\section{Notation and Organization}

We use $\Rad^{\ell}$ to denote the uniform distribution on $\{\pm 1\}^{\ell}$. Any expectations or probabilities over $x$ are over the distribution $x \sim \Rad^d$. Whenever the variable $x$ is used, we use $z = z(x)$ to denote the projection of $x$ onto its first two coordinates (the space spanned by $\mu_1$ and $\mu_2$), and we use $\xi = \xi(x)$ to denote the projection of $x$ orthogonal to $z$. We use expectations over $z$ and $\xi$ to mean expectation over $z(x)$ and $\xi(x)$, where $x \sim \Rad^{\ell}$. For a vector $v \in \mathbb{R}^d$, we use the notation $v_{\setminus i}$ to denote the vector $v - e_iv_i$, which sets the $i$th coordinate of $v$ to $0$.

Throughout, we often remove the superscript $(t)$ and use $w$ to denote $w^{(t)}$. Sometimes for emphasis, or when we are comparing $w^{(t + 1)}$ and $w^{(t)}$, we will include the superscripts. However, the reader should not be alarmed if we use $w^{(t)}$ and $w$ interchangeably, even in such calculations. The same holds for all other neural network parameters.

Throughout the appendix, will use the notation $x \in a \pm b$ to mean that $|x - a| \leq b$. 

We recall that much helpful notation used throughout the appendix is summarized in Table~\ref{table:notation} in the main body.

In Section~\ref{sec:losslemmas}, we derive several auxiliary lemmas which will be used throughout the appendix. In Section~\ref{apx:phase1}, we prove our results for Phase 1. In Section~\ref{apx:phase2}, we prove our results for Phase 2.  In Section~\ref{apx:thmpf}, we prove the main theorem.  In Section~\ref{sec:stat}, we sketch why $\tilde{\Theta}(d)$ samples are needed for learning the XOR function with a rotationally invariant algorithm.

\section{Auxiliary Lemmas}\label{sec:losslemmas}

\subsection{From Booleans to Gaussians}\label{sec:boolean}
The following section provides some lemmas to handle the fact that our data is drawn from the Boolean hypercube, and not Gaussian. At a high level, our goal is to show that for all neurons $w$, for any $a > 0$ we have that 
$$\mathbb{P}_{\xi \sim \Rad^d}[|\xi^Tw| \leq a] \approx \mathbb{P}_{X \sim \mathcal{N}(0, I_d)}[|X^Tw| \leq a].$$

Of course, this is not true for all vectors $w$ (eg. if $w$ is sparse, these two probabilities may differ significantly.)
Fortunately, the neurons $w$ we care about will be equal to their initialization (which is a random vector on the sphere) plus a small arbitrary perturbation. Thus we will define a notion of a ``well-spread vector'', claim that at initialization with high probability all neurons are well-spread (Lemma~\ref{lemma:spread_init}), and then use Lemmas~\ref{lemma:infinity} and \ref{lemma:boolean2} to relate the Boolean probabilities to Gaussian probabilities, for all neurons $w$ that are near well-spread vectors. Later on in training, the neurons may not be close to their initialization, but we will be able to bound their $\ell_{\infty}$ norm, and thus use the well known Berry-Esseen Lemma to relate the Boolean and Gaussian probabilities.
\begin{definition}[Well-Spread Vector]\label{def:wellspread}
    We say a vector $v \in \mathbb{R}^d$ is \em $c$-well-spread \em if:
    \begin{enumerate}
        \item $\|v\|_3^3 \leq 20\|v\|_2^3d^{-1/2}$ and $\|v\|_{\infty} \leq \frac{\log(d)}{\sqrt{d}}\|v\|_2$.
        \item Let $S$ index the set of $d/c^2$ coordinates of $v$ with smallest absolute value (break ties arbitrarily). Then $\sum_{i \in S}|v_i| \geq \frac{\|v\|\sqrt{d}}{c^5}$, and $\max_{i \in S} |v_i| \leq \frac{\|v\|}{c\sqrt{d}}$.
    \end{enumerate}
\end{definition}
\begin{lemma}\label{lemma:spread_init}
For any constant $c$ large enough, the following holds. For any $r > 0$, if $w \sim \mathcal{S}^{d-1}(r)$, then with probability $1 - d^{-\omega(1)}$, $w$ is $c$-well-spread.
\end{lemma}
\begin{proof}
Without loss of generality we may assume $r = 1$. We can write $w = u/X$, where $u \sim \mathcal{N}(0, \frac{1}{d}I_d)$, and $X = \|u\|$. With probability $1 - d^{-\omega(1)}$, the following events hold:
\begin{enumerate}
    \item We have $X \in [0.9, 1.1]$. This holds by Bernstein's concentration inequality for sub-exponential random variables.
    \item $\|u\|_{\infty} \leq \frac{\log(d)}{2\sqrt{d}}$. This holds by a union bound over all $d$ coordinates, and using the Gaussian CDF.
    \item $\|u\|_3^3 \leq 10 d^{-1/2}$. This holds by applying Lipshitz concentration of Gaussians to the function $f(u) := \|u\|_3$. Indeed, this function is 1-Lipshitz since $\|u\|_3 \leq \|u\|_2$. We have $\mathbb{E}[\|u\|_3] = \mathbb{E}[(\sum_i |u_i^3|)^{1/3}] \leq \left(\sum_i\mathbb{E}[\sum |u_i^3|]\right)^{1/3} \leq \left(\sum_i\sqrt{\mathbb{E}[\sum u_i^6]}\right)^{1/3} = \left(d\sqrt{d^{-3}15}\right)^{1/3} \leq 2d^{-1/6}$, so Lipshitz concentration yields that $\|u\|_3 \leq 2.1d^{-1/6}$ with probability $1 - d^{-\omega(1)}$, and such, that $\|u\|_3^3 \leq 10d^{-1/2}$.
    \item For $c$ large enough, there are at least $d/c^2$ indices $i$ for which $|u_i| \leq \frac{1}{2c\sqrt{d}}$. This holds by a Chernoff bound for Bernoulli random variables, since for any $i$, $\mathbb{P}\left[|u_i| \leq \frac{1}{2c\sqrt{d}}\right] \geq \frac{1}{4c}$.
    \item There are no more than $d/(2c^2)$ coordinates $i$ for which $|u_i| \leq \frac{10}{c^3\sqrt{d}}$. This again holds by a Chernoff bound for Bernoulli random variables, since for any $i$, $\mathbb{P}\left[|u_i| \leq \frac{10}{c^3\sqrt{d}}\right] \leq \frac{20}{c^3}$.
\end{enumerate}
Combining the first, second, and third items above yields the first property of a well-spread vector for $w$. Combining the first, fourth, and fifth items above yields the second property of a well-spread vector for $w$.
\end{proof}
\begin{lemma}\label{lemma:infinity}
There exists a universal constant $C_0$ such that the following holds for any $C$-well-spread vector $v$ for $C \geq C_0$. For any $\Delta \in \mathbb{R}^d$ with $\|\Delta\|_2 \leq \zeta\|v\|_2$ if $\zeta \leq \frac{1}{C^{10000}}$, we have for $d$ large enough:
\begin{align}\mathbb{P}_{\xi}\left[|\xi^T(v + \Delta)| \leq \frac{\|v\|}{\sqrt{d}}\right] \geq \frac{1}{2}\exp(-100C^8)\frac{1}{\sqrt{d}}.
\end{align}
\end{lemma}
\begin{lemma}\label{lemma:boolean2}
There exists a universal constant $C_0$ such that the following holds for any $C$-well-spread vector $v$ for $C \geq C_0$. For any $\Delta \in \mathbb{R}^d$ with $\|\Delta\|_2 \leq \zeta\|v\|_2$ if $\zeta \leq \frac{1}{C^{10000}}$, we have for $d$ large enough,
\begin{align}
    \left|\mathbb{P}_{\xi \sim \Rad^d}[\xi^T(v + \Delta) \in [a\|v\|, b\|v\|]] - P_{\frac{|b - a|}{2}}\right| \leq 2P_{\frac{|b - a|}{2}}(\sqrt{\zeta} + max(|a|, |b|)^2) + 200\cbe d^{-1/2},
\end{align}
where 
\begin{align}
    P_{c} := \mathbb{P}_{G \sim \mathcal{N}(0, 1)}[|G| \leq c] = \mathbb{P}_{X \sim \mathcal{N}(0, I_d)}[|X^Tv| \leq c\|v\|],
\end{align}
and $\cbe$ is the constant from Theorem~\ref{thm:be}.
\end{lemma}

Our main tool in proving these two lemmas is the Berry-Esseen Inequality.
\begin{theorem}[Berry-Esseen Inequality]\label{thm:be}
There exists a universal constant $\cbe$ such that for independent mean $0$ random variables $X_1, \cdots X_n$, with $\mathbb{E}[X^2] = \sigma_i^2$, and $\mathbb{E}[|X_i|^3] = \rho_i$, we have
\begin{align}
    \sup_x \left|\Phi(x) - F_n(x)\right| \leq \cbe\frac{\sum \rho_i}{(\sum \sigma_i^2)^{3/2}},
\end{align}
where $\Phi$ is the CDF of $\mathcal{N}(0, 1)$, and $F_n$ is the CDF of $\frac{\sum X_i}{\sqrt{\sum \sigma_i^2}}$. Thus if $u \in \mathbb{R}^{\ell}$, then
\begin{align}
    \sup_x \left|\mathbb{P}_{\xi \sim \Rad^{\ell}}[u^T\xi \geq x] - \mathbb{P}_{G \sim \mathcal{N}(0, 1)}[G \geq x/\|u\|]\right| \leq \cbe\frac{\|u\|^3}{\|u\|_2^3}.
\end{align}
\end{theorem}
We will use the following lemma which follows from Chebychev's inequality and Berry-Esseen.
\begin{lemma}\label{lemma:RW}
The following holds for any constant $C$ large enough. Suppose $u \in \mathbb{R}^{\ell}$ satisfies $\|u\|_{\infty} \leq 1$. Then for $\xi \sim \Rad^{\ell}$, we have 
\begin{align}
\mathbb{P}[|\xi^Tu|\leq C] \geq \frac{1}{C\sqrt{\ell}}.
\end{align}
\end{lemma}
\begin{proof}
We need to do casework on the size of $\|u\|$. Applying Berry-Esseen, if $C \geq 8\sqrt{\pi}\cbe$ and $\|u\|_2 \geq 8\sqrt{\pi}\cbe$, we have
\begin{align}
    \mathbb{P}_{\xi}[|\xi^Tu|\leq C] &\geq \mathbb{P}_{G \sim \mathcal{N}(0, 1)}[|G|\leq C/\|u\|_2] - 2\cbe \frac{\|u\|_3^3}{\|u\|_2^3}\\
    &\geq \mathbb{P}_{G \sim \mathcal{N}(0, 1)}[|G|\leq C/\|u\|_2]  - 2\cbe \frac{\|u\|_2^2\|u\|_{\infty}}{\|u\|_2^3}\\
    &\geq\mathbb{P}_{G \sim \mathcal{N}(0, 1)}[|G|\leq C/\|u\|_2]  - \frac{2\cbe}{\|u\|_2}\\
    &\geq \sqrt{\frac{1}{\pi}}\frac{C}{C + \|u\|_2} - \frac{2\cbe}{\|u\|_2}\\
    &\geq \frac{1}{C\sqrt{\ell}}
\end{align}
Here in the second inequality we used Holder's inequality (with $p = 1$, $q = \infty$), and in the final inequality we used the bounds $8\sqrt{\pi}\cbe \leq \|u\|_2 \leq \sqrt{\ell}\|u\|_{\infty} \leq \sqrt{\ell}$ and $C \geq 8\sqrt{\pi}\cbe$.

If $\|u\| \leq 4\sqrt{\pi}\cbe$, then by Chebychev's inequality, we have for $C$ large enough:
\begin{align}
    \mathbb{P}_{\xi}[|\xi^Tu|\geq C] \leq \frac{\mathbb{E}(\xi^Tu)^2}{C^2} = \frac{\|u\|^2_2}{C^2} \leq 1 - \frac{1}{C\sqrt{\ell}}.
\end{align}
\end{proof}

\begin{proof}[Proof of Lemma~\ref{lemma:infinity}]
Let $w := v + \Delta$. Let $B$ be the set of ``bad'' coordinates on which $|\Delta_i| \geq \frac{\|v\|}{3C^3\sqrt{d}}$. Thus since $\|\Delta\|  \leq \zeta \|v\|$, we have $|B| \leq (3C^3)^2d\zeta^2 \leq \frac{d}{3C^4}$ for $C$ large enough. Let $S$ be the set of $d/C^2$ coordinates of $v$ with smallest absolute value (as in the definition of well-spread). Thus letting $S' := S \setminus B$, we have:
\begin{enumerate}[{\bfseries{D\arabic{enumi}}}]
    \item\label{D1}
    \begin{align}
        \sum_{i \in S'}|w_i| &\geq \sum_{i \in S}|v_i| - |S'|\max_{i \notin B}|\Delta_i| - |B|\max_{i \in S}|v_i|\\
        &\geq \frac{\|v\|\sqrt{d}}{C^5} - \frac{d}{C^2}\left(\frac{\|v\|}{3C^3\sqrt{d}}\right) - \frac{d}{3C^4}\left(\frac{\|v\|}{C\sqrt{d}}\right)\\
        &\geq \frac{\|v\|\sqrt{d}}{3C^5}.
    \end{align}
    \item\label{D2} $\max_{i \in S'} |w_i| \leq \frac{\|v\|_2}{C\sqrt{d}} + \frac{\|v\|}{3C^3\sqrt{d}} \leq \frac{2\|v\|_2}{C\sqrt{d}}$.
\end{enumerate}

Without loss of generality, assume the coordinates of $w$ are ordered with the with the indices not in $S'$ first, followed by the indices in $S'$.


Let $X_t$ for $t = 1 \ldots d$ be the random walk $X_t = \sum_{i = 1}^t \xi_i w_i$. Let $\tau$ be the first time at which the random walk crosses zero after step $d - |S'|$, such that $|X_{\tau}| \leq |w_{\tau}|$. If it never crosses zero after this time, let $\tau := d + 1$.

Let $A$ be the event that $\tau \neq d + 1$, and let $B_{t}$ be the event that $\left|\sum_{i = t + 1}^d \xi_i w_i\right| \leq \frac{\|v\|}{2\sqrt{d}}$. We proceed with a sequence of claims.



\begin{claim}\label{claim:1}
    If $A$ and $B_{\tau}$ occur, then $ \left|\sum_{i = 1}^d \xi_i w_i\right| \leq \frac{\|v\|}{\sqrt{d}}$.
\end{claim}
\begin{proof}
If $A$ and $B_{\tau}$ occur, then
    \begin{align}
         \left|\sum_{i = 1}^d \xi_i w_i\right| &\leq |X_{\tau}| + \left|\sum_{i = \tau + 1}^d \xi_i w_i\right|\\
         &\leq |w_{\tau}| + \frac{\|v\|}{2\sqrt{d}}\\
         &\leq \max_{i \in S'}|w_i| + \frac{\|v\|}{2\sqrt{d}} \leq \frac{\|v\|}{\sqrt{d}},
    \end{align}
    for $C$ large enough.
\end{proof}
\begin{claim}
\begin{align}
        \mathbb{P}_{\xi}\left[A \text{ and } B_{\tau}\right] \geq  \mathbb{P}_{\xi}\left[A\right]\min_{t \in [d - |S'|, d]}\mathbb{P}_{\xi}\left[B_t\right]
\end{align}
\end{claim}
\begin{proof}
\begin{align}
    \mathbb{P}_{\xi}\left[A \text{ and } B_{\tau}\right] &= \sum_{t = d - |S'|}^{d} 
    \mathbb{P}_{\xi}[\tau = t]\mathbb{P}_{\xi}[B_{t} | \tau = t]\\
    &= \sum_{t = d - |S'|}^{d}\mathbb{P}_{\xi}[\tau = t]\mathbb{P}_{\xi}[B_{t}]\\
    &\geq \left(\sum_{t = d - |S'|}^{d}\mathbb{P}_{\xi}[\tau = t]\right)\min_{t \in [d - |S'|, d]}\mathbb{P}_{\xi}\left[B_t\right]\\
    &= \mathbb{P}_{\xi}[A]\min_{t \in [d - |S'|, d]}\mathbb{P}_{\xi}\left[B_t\right].
\end{align}
\end{proof}
\begin{claim}
    For any $t \in [d - |S'|, d]$, $\mathbb{P}_{\xi}\left[B_t\right] \geq \frac{1}{\sqrt{d}}$.
\end{claim}
\begin{proof}
We will apply Lemma~\ref{lemma:RW}. Fix $t$ and let $u' \in \mathbb{R}^{d - t}$ be the vector with $u'_i = w_{i + t}$. Then $\|u'\|_{\infty} \leq \frac{2\|v\|_2}{C\sqrt{d}}$ by \ref{D2}. Let $u := \frac{u'}{\|u'\|_{\infty}}$. Applying Lemma~\ref{lemma:RW} to $u$ yields (so long as $C \geq 4C_{\ref{lemma:RW}}$, where $C_{\ref{lemma:RW}}$ is the constant in Lemma~\ref{lemma:RW}):
\begin{align}
\mathbb{P}\left[|\xi^Tu'| \leq \frac{\|v\|}{2\sqrt{d}}\right] &= \mathbb{P}\left[|\xi^Tu|\leq \frac{\|v\|}{2\sqrt{d}\|u\|_{\infty}}\right]\\
&\geq \mathbb{P}\left[|\xi^Tu|\leq \frac{C}{4}\right] \\
&\geq \frac{1}{C_{\ref{lemma:RW}}\sqrt{d - t}}\\
&\geq \frac{1}{C_{\ref{lemma:RW}}\sqrt{|S'|}}\\
&\geq \frac{1}{C_{\ref{lemma:RW}}\sqrt{d/C^2}} \geq \frac{1}{\sqrt{d}}.
\end{align}
Here the first inequality in the last line follows from the fact that $|S'| \leq |S| = \frac{d}{C^2}$.
\end{proof}
\begin{claim}\label{claim:4}
\begin{align}
    \mathbb{P}_{\xi}[\tau \geq d - |S'|] \geq \frac{1}{2}\exp(-100C^8).
\end{align}
\end{claim}
\begin{proof}
Because for any scalar $b$ and any set of coordinates $P$ we have $\mathbb{P}_{\xi} \left[\sum_{i \in P} \xi_i w_i = b\right] = \mathbb{P}_{\xi} \left[\sum_{i \in P} \xi_i w_i = -b\right]$, we have
\begin{align}\label{eq:prod}
    \mathbb{P}_{\xi}[\tau \geq d - |S'|] &\geq \mathbb{P}_{\xi}\left[|X_{d - |S'|}| \leq 2\|v\|\right]\mathbb{P}\left[\sum_{i = d - |S'| + 1}^d\xi_i^Tw_i \geq 2\|v\|\right].
\end{align}
We will show that these two probabilities are sufficiently large via Berry-Esseen. For the second probability, let $u \in \mathbb{R}^{d - |S'|}$ be the vector with $u_i = w_{i + d - |S'|}$. Then we have by \ref{D1},
\begin{align}\|u\|_2 \geq \frac{1}{\sqrt{|S'|}}\|u\|_1 \geq \frac{\|v\|\sqrt{d}}{3C^5\sqrt{|S'|}} \geq  \frac{\|v\|\sqrt{d}}{3C^5\sqrt{|S|}} = \frac{\|v\|}{3C^4},
\end{align}
while by Holder's inequality and \ref{D2}, we have
\begin{align}
    \|u\|_3^3 \leq \|u\|_2^2\|u\|_{\infty} \leq \|u\|_2^2 \frac{2\|v\|}{C\sqrt{d}}.
\end{align}
Thus by Berry-Esseen, we have 
\begin{align}\label{eq:be}
    \mathbb{P}\left[\sum_{i = d - |S'| + 1}^d\xi_i^Tw_i \geq 2\|v\|\right] &\geq \mathbb{P}_{G \sim \mathcal{N}(0, 1)}[G \geq 2\|v\|/\|u\|] - \cbe \frac{2\|v\|}{C\sqrt{d}\|u\|}\\
    &\geq  \mathbb{P}_{G \sim \mathcal{N}(0, 1)}[G \geq 6C^4] - \cbe \frac{6C^3}{\sqrt{d}}\\
    &\geq \exp(-100C^8),
\end{align}
for a constant $C$ large enough (and $d$ sufficiently large).

Now consider the probability $\mathbb{P}_{\xi}\left[|X_{d - |S'|}| \leq 2\|v\|\right]$. By Chebychev's inequality, we have 
\begin{align}
  \mathbb{P}_{\xi}\left[|X_{d - |S'|}| \geq 2\|v\|\right] &\leq   \frac{\mathbb{E}[X_{d - |S'|}^2]}{4\|v\|^2}\\
  &= \frac{\mathbb{E}[\sum_{i = 1}^{d - |S'|}w_i^2]}{4\|v\|^2}\\
  &\leq \frac{\|w\|^2}{4\|v\|^2} \leq \frac{1}{2}.
\end{align}
Combining this with Eq~\ref{eq:be} and Eq~\ref{eq:prod} yields the claim.
\end{proof}
Combining Claims~\ref{claim:1}-\ref{claim:4} yields
\begin{align}
    \mathbb{P}_{\xi}\left[|\xi^Tw| \leq \frac{\|v\|}{\sqrt{d}}\right] \geq \frac{1}{2}\exp(-100C^8)\frac{1}{\sqrt{d}},
\end{align}
which proves the lemma.
\end{proof}

\begin{proof}[Proof of Lemma~\ref{lemma:boolean2}]
We can without loss of generality assume $b \geq |a|$, since the variable $\xi$ is symmetric. Define $B := \{i : |\Delta_i| \geq \sqrt{\zeta}\frac{\|v\|}{\sqrt{d}}\}$. Observe that since $\|\Delta\|_2 \leq \zeta \|v\|$, we have $|B| \leq \zeta d$. Let $w = v + \Delta$. We can write
\begin{align}\label{eq:xint}
    \mathbb{P}_{\xi}\left[\xi^Tw \in [a, b]\|v\|\right] &= \int_{x = -\infty}^{\infty} \mathbb{P}_{\xi}\left[\sum_{i \in B}\xi_iw_i = x\|v\| \right] \mathbb{P}_{\xi}\left[\sum_{i \in [d] \setminus B} \xi_i w_i \in [a - x, b - x]\|v\| \right]
\end{align}
We use the following claim to bound this integral.
\begin{claim}
    \begin{align}
        \left|\mathbb{P}_{\xi}\left[\sum_{i \in [d] \setminus B} \xi_i w_i \in [a - x, b - x]\|v\| \right] - P_{\frac{b - a}{2}}\right| \leq P_{\frac{b - a}{2}}\left(\sqrt{\zeta} + (|x| + b)^2/2\right) + 200 \cbe d^{-1/2}.
    \end{align}
\end{claim}
\begin{proof}
We use Berry-Esseen. Let $u \in \mathbb{R}^{d - |B|}$ with coordinates equal to the set $\{w_i\}_{i \in [d] \setminus S}$. Then
\begin{align}
    |\|u\|^2 - \|v\|^2| &\leq |B|\max_{i \in B}|v_i|^2 + \|\Delta\|^2 + 2\|v\|\|\Delta\|\\
    &\leq \zeta \log(d)^2\|v\|^2 + 3\zeta \|v\|^2\\
    &\leq \sqrt{\zeta}\|v\|^2,
\end{align}
where the last line follows because $\zeta \leq 1/\log^5(d)$.
Further 
\begin{align}
    \|u\|_3^3 \leq \sum_{i \notin B}w_i^3  \leq \sum_{i \notin B}4\left(v_i^3 + \Delta_i^3\right) \leq \leq 4\|v\|_3^3 + 4 \|\Delta\|^2 \sup_{i \notin B}|\Delta_i| \leq 4\|v\|_3^3 + 4\zeta^{3/2}\|v\|_2^3 d^{-1/2} \leq 100\|v\|_2^3d^{-1/2}.
\end{align}
In both these computations we have used the fact that $v$ is well-spread to bound $\|v\|_{\infty}$ and $\|v\|_3^3$.

By Berry-Esseen (Theorem~\ref{thm:be}), we have 
\begin{align}\label{eq:offset}
    \mathbb{P}_{\xi}\left[\sum_{i \in [d] \setminus B} \xi_i w_i \in [a - x, b - x]\|v\| \right] \in \mathbb{P}_{G \sim \mathcal{N}(0, 1)}\left[G \in [a - x, b - x]\frac{\|v\|}{\|u\|}\right] \pm 2\cbe \frac{\|u\|_3^3}{\|u\|_2^3}
\end{align}
Now we have
\begin{align}
     \mathbb{P}_{G \sim \mathcal{N}(0, 1)}&\left[G \in [a - x, b - x]\frac{\|v\|}{\|u\|}\right]\\
     &\in \mathbb{P}_{G \sim \mathcal{N}(0, 1)}\left[G \in [a - x, b - x]\right]\left[1 - \sqrt{\zeta}, 1 + \sqrt{\zeta}\right]\\
     &\in \mathbb{P}_{G \sim \mathcal{N}(0, 1)}\left[G \in \left[-\frac{b - a}{2}, \frac{b - a}{2}\right]\right]\left[\frac{\phi(b + |x|)}{\phi(0)} - \sqrt{\zeta}, 1 + \sqrt{\zeta}\right].\\
\end{align}
where we have used the fact that $\frac{\|v\|}{\|u\|} \in \left[\frac{1}{\sqrt{1 - \sqrt{\zeta}}}, \frac{1}{\sqrt{1 + \sqrt{\zeta}}}\right] \in [1 - \sqrt{\zeta}, 1 + \sqrt{\zeta}]$, and that $a \leq |b|$.

Now 
\begin{align}
    \frac{\phi(b + |x|)}{\phi(0)} = \exp(-(b + |x|)^2/2) \geq 1 - (b + |x|)^2/2,
\end{align}
so
\begin{align}
     \mathbb{P}_{G \sim \mathcal{N}(0, 1)}&\left[G \in [a - x, b - x]\frac{\|v\|}{\|u\|}\right] \in P_{\frac{b - a}{2}}\left[1 - \sqrt{\zeta} - (b + |x|)^2/2, 1 + \sqrt{\zeta}\right]\\
\end{align}

Thus returning to Eq~\ref{eq:offset}, we have
\begin{align}
     \mathbb{P}_{\xi}\left[\sum_{i \in [d] \setminus B} \xi_i w_i \in [a - x, b - x]\|v\| \right] \in P_{\frac{b - a}{2}}\left[\left(1 - (|x| + b)^2/2 - \sqrt{\zeta}\right), 1 + \sqrt{\zeta}\right] \pm 200\cbe d^{-1/2},
\end{align}
which yields the desired result.
\end{proof}
Returning to Eq~\ref{eq:xint}, we have
\begin{align}
 \left|\mathbb{P}_{\xi}\left[\xi^Tw \in [a\|v\|, b\|v\|]\right] - P_{\frac{b - a}{2}}\right| &\leq P_{\frac{b - a}{2}}\sqrt{\zeta} + 200\cbe d^{-1/2} + \frac{1}{2}P_{\frac{b - a}{2}} \int_{x = -\infty}^{\infty} \mathbb{P}_{\xi}\left[\sum_{i \in B}\xi_iw_i = x\|v\| \right](|x| + b)^2\\
 &\leq P_a(\sqrt{\zeta} + b^2) + 200 \cbe d^{-1/2} + P_{\frac{b - a}{2}} \int_{x = -\infty}^{\infty} \mathbb{P}_{\xi}\left[\sum_{i \in B}\xi_iw_i = x\|v\| \right]x^2\\
 &= P_a(\sqrt{\zeta} + b^2) + 200\cbe d^{-1/2} + P_{\frac{b - a}{2}} \frac{1}{\|v\|^2} \mathbb{E}_{\xi}\left[\left(\sum_{i \in B}\xi_iw_i\right)^2\right]\\
 &= P_a(\sqrt{\zeta} + b^2) + 200 \cbe d^{-1/2} + P_{\frac{b - a}{2}} \frac{1}{\|v\|^2}\left(\sum_{i \in B}w_i^2\right).
\end{align}
Now 
\begin{align}
    \sum_{i \in B}w_i^2 \leq \sum_{i \in B}2(\Delta_i^2 + v_i^2) \leq 2\|\Delta\|^2 + 2|B|\max_i v_i^2 \leq \sqrt{\zeta}\|v\|^2.
\end{align}
Thus we have 
\begin{align}
      \left|\mathbb{P}_{\xi}\left[|\xi^Tw| \in [a\|v\|, b\|v\|]\right] - P_{\frac{b - a}{2}}\right| &\leq 2P_{\frac{b - a}{2}}(\sqrt{\zeta} + b^2) + 200 \cbe d^{-1/2},
\end{align}
which proves the lemma.
\end{proof}

\subsection{Simultaneous Training of Two-Layer ReLU Networks}
We will make use of the following general-purpose lemma for training neural networks with ReLU activations and Lipshitz loss functions.

\begin{lemma}[Empirical Concentration]\label{lemma:emp}
If we train with data from $P_d$ on any loss $L$ that is 2-Lipshitz in the output of the network, with a minibatch size of $m \geq d\log(d)^2$, we have for any neuron $(w, a_w )$ with probability $1 - d^{-\omega(1)}$,
\begin{enumerate}
    \item $\|\nabla_w L - \nabla_w \hat{L}\|^2 \leq \frac{d\log^2(d)}{m} a_w^2$; and for any $i \in [d]$, $\|\nabla_{w_i} L - \nabla_{w_i} \hat{L}\|^2 \leq \frac{\log^2(d)}{m} a_w^2$;
    \item $\|\nabla_{a_w} L - \nabla_{a_w} \hat{L}\|^2 \leq \frac{d\log^2(d)}{m} \|w\|^2$; 
\end{enumerate}

\end{lemma}
\begin{proof}
    We can use the Generalized Hoeffding's inequality to prove this.
    
    For the first statement, each coordinate of $\ell'_{\rho}(x)\sigma'(w^Tx)x$ minus its expectation a random variable bounded by a constant, and is thus subgaussian. 
    
    For the second statements, $\ell'_{\rho}(x)\sigma(w^Tx)$ minus its expectation is subgaussian with parameter $\|w\|$. 

\end{proof}

\begin{lemma}\label{lemma:layer_balance}
If we train with data from $P_d$ on any loss $\ell$ that is 2-Lipshitz with respect the the output of the network, with a minibatch size of $m \geq d\log(d)^2$ and $\eta \leq 1/4$, we have for any neuron $(w, a_w )$ with probability $1 - d^{-\omega(1)}$,
    \begin{enumerate}[{\bfseries{S\arabic{enumi}}}]
        \item \label{Sw} $\|\nabla_w L\| \leq 2|a_w |$;
        \item \label{Sa} $\|\nabla_a L\| \leq 2\|w\|$;
        \item \label{Sb1} If $|a_w^{(t)}| \leq \|w^{(t)}\|$, then $|a_w^{(t + 1)}| \leq \|w^{(t + 1)}\|$.
        \item \label{Sb2} $\|w^{(t + 1)}\|^2 - |(a_w^{(t + 1)})|^2 \leq 4\eta^2 |a_w^{(t)}|^2 + \|w^{(t)}\|^2 - |(a_w^{(t)})|^2$.
    \end{enumerate}
\end{lemma}
\begin{proof}
Consider the first statement first. We have 
\begin{align}
    \frac{1}{|a_w|}\|\nabla_w L\| &= \sup_{v: \|v\| = 1}\mathbb{E}_x \ell'_{\rho}(x)\sigma'(w^Tx)x^Tv \\
    &\leq \mathbb{E}_x |\ell'_{\rho}(x)| |x^Tv|\\
    &\leq 2\mathbb{E}_x |x^Tv|\\
    &\leq 2.
\end{align}
For the second statement, we have
\begin{align}
    \|\nabla_a L\| &= \mathbb{E}_x \ell'_{\rho}(x)\sigma(w^Tx)\\
    &\leq 2\|w\|.
\end{align}
For the third and fouth statements, we use the shorthand $a = a_w^{(t)}$, $w = w^{(t)}$. To prove the third statement, we can write
\begin{align}
    (a^{(t + 1)})^2 &= \left(a - \eta \nabla_{a} \hat{L}\right)^2\\
    &= (a)^2 - 2\eta a\nabla_{a} \hat{L} + \eta^2 (\nabla_{a} \hat{L})^2
\end{align}
and 
\begin{align}
    \|w^{(t + 1)}\|^2 &= \|w - \eta \nabla_w \hat{L}\|^2\\
    &= \|w\|^2 - 2\eta w^T\nabla_w \hat{L} + \eta^2 \|\nabla_w \hat{L}\|^2.
\end{align}
Because we use the ReLu activation, we have 
\begin{align}
    (w^{(t)})^T\nabla_w \hat{L} = a^{(t)}\nabla_{a} \hat{L}.
\end{align}

Thus
\begin{align}
     (a^{(t + 1)})^2 -  \|w^{(t + 1)}\|^2 &=  (a^{(t)})^2 -  \|w^{(t)}\|^2 + \eta^2\left((\nabla_{a} \hat{L})^2 - \|\nabla_w \hat{L}\|^2\right)\\
     &\leq (a^{(t)})^2 -  \|w^{(t)}\|^2 + \eta^2\left((\nabla_{a} \hat{L})^2 -  \frac{1}{\|w\|^2}(w^T\nabla_w \hat{L})^2\right)\\
     &= (a^{(t)})^2 -  \|w^{(t)}\|^2 + \eta^2\left((\nabla_{a} \hat{L})^2 -  \frac{a^2}{\|w\|^2}(\nabla_{a} \hat{L})^2\right)\\
     &= (a^{(t)})^2 -  \|w^{(t)}\|^2 + \frac{\eta^2(\nabla_{a} \hat{L})^2}{\|w\|^2}\left(\|w\|^2 - a^2\right)\\
     &= \left((a^{(t)})^2 - \|w^{(t)}\|^2\right)\left(1 - \frac{\eta^2(\nabla_{a} \hat{L})^2}{\|w\|^2}\right)
\end{align}

By the previous conclusions of the lemma and Lemma~\ref{lemma:emp}, we have with probability $1 - d^{-\omega(1)}$,
\begin{align}
    (\nabla_{a} \hat{L})^2 \leq 2(\nabla_a L)^2 + 2\zeta \|w\|^2 \leq 2 \|w\|^2 + 2\zeta \|w\|^2,
\end{align}
where $\zeta = \log(d)^2\frac{d}{m}$. Assuming $\eta \leq 1/4$, this yields the desired statement.

For the fourth result, we have with probability $1 - d^{-\omega(1)}$,
\begin{align}
    \|w^{(t + 1)}\|^2- (a^{(t + 1)})^2  -  \|w^{(t)}\|^2 + (a^{(t)})^2 & = \eta^2\left(\|\nabla_w \hat{L}\|^2 - (\nabla_{a} \hat{L})^2\right)\\
    &\leq 2\eta^2 \left(\|\nabla_w L\|^2 + \|\nabla_w \hat{L} - \nabla_w L\|^2\right)\\
    &\leq 4\eta^2a_w^2.
\end{align}
    
\end{proof}

\section{Phase 1}\label{apx:phase1}
In this section, we prove the following lemma.
\begin{restatable}[Output of Phase 1; Formal]{lemma}{lemphase}\label{lemma:phase1formal}
For any constants $c$ sufficiently large, and $C$ sufficiently large in terms of $c$, for any $d$ large enough, the following holds. Let $\theta := 1/\log(d)^C$.

Suppose we train a 2-layer neural network with minibatch SGD as in Section~\ref{sec:training} with a minibatch size of $m \geq d/\theta^2$, width $1/\theta \leq p \leq d^C$, and step size $\eta \leq \theta$, and initialization scale $\theta$. Then with probability at least $1 - \theta$, after some $T_1 = \Theta(\log(d)/\eta)$ steps of minibatch SGD, the network $\rho_{T_1}$ satisfies:
    \begin{enumerate}
        \item $\mathbb{E}_{\rho_{T_1}}[\|a_w w\|] \leq 1$;
        \item $\mathbb{E}_{\rho_{T_1}}[\|\wpe + \wopp\|^2] \leq 4\theta^{2}$;
        \item For all $\mu \in \muset$, on at least a $0.1$ fraction of the neurons, we have $\|\wsig\| \geq \log(d)^{c}\theta$ and $\wsig^T\mu > 0$.
    \end{enumerate}
Additionally, $$\mathbb{E}_{\rho_{T_1}}[\|w\|^2] \leq \mathbb{E}_{\rho_{T_1}}[|a_w|^2] + \sqrt{\eta},$$ and for all neurons, we have $|a_w| \leq \|w\|$.
\end{restatable}

\subsection{Phase 1 Gradient Bounds}
The core ingredients of Phase 1 are the following three lemmas, which relate the gradient $\nabla L_{\rho}$ to $\nabla L_0$ and compute several properties of the $L_0$ population gradient. Recall that we have
\begin{align}\label{eq:L0}
    \nabla_w L_{0} &= \mathbb{E}_x\frac{\partial}{\partial w}\ell_{0}(x; \rho) = -\mathbb{E}_x a_w y(x)\sigma'(w^Tx)x; \\
    \nabla_{a_w} L_{0} &=  \mathbb{E}_x\frac{\partial}{\partial a_w}\ell_{0}(x; \rho) = -\mathbb{E}_x y(x)\sigma(w^Tx),
\end{align}
which is independent of the full distribution $\rho$.

\begin{lemma}\label{lemma:graddiff1}
For any neuron $(w, a_w )$, 
    \begin{enumerate}[{\bfseries{G\arabic{enumi}}}]
    \item\label{Ga}
    $|\nabla_{a_w} L_0 - \nabla_{a_w} L_{\rho}| \leq 2\|w\| \mathbb{E}_{\rho}[\|a_w w\|]$.
    \item\label{Gw} $\|\nabla_w L_0 - \nabla_w L_{\rho}\| \leq 2|a_w|\mathbb{E}_{\rho}[\|a_w w\|]$.
    \end{enumerate}
\end{lemma}

\begin{lemma}\label{lemma:approxsig}
Suppose $\mathbb{E}_{\rho}[\|a_w w\|] \leq d^{O(1)}$. For any neuron $(w, a_w)$, for any $i \in [d]$,
    \begin{align}
        &\|\nabla_{w_{i}} L_{\rho} - \nabla_{w_{i}} L_0\|\\
        &\qquad \leq  |a_w|\left(4(\mathbb{E}_{\rho}[\|a_w w_{i}\|]) + 2\log(d)\mathbb{E}_{\rho}[\|a_w w\|]\mathbb{E}_{x}\mathbf{1}(|\ix^Tw| \leq |w_i|) + d^{-\omega(1)}\right).
    \end{align}
\end{lemma}

\begin{lemma}[Signal Subspace $L_0$ Population Gradients]\label{lemma:pop1}
For any neuron $(w, a_w)$, we have
\begin{enumerate}[{\bfseries{B\arabic{enumi}}}]
    \item \label{Psig} $-\wsig^T \nabla_{w} L_0 = \frac{\sqrt{2}}{4}|a_w|\mathbb{E}_{\xi}\mathbf{1}(|w^T\xi| \leq \sqrt{2}\|\wsig\|)\|\wsig\|$
    \item \label{Popp} $-\wopp^T \nabla_{w} L_0 = -\frac{\sqrt{2}}{4}|a_w|\mathbb{E}_{\xi}\mathbf{1}(|w^T\xi| \leq \sqrt{2}\|\wopp\|)\|\wopp\|$
    \item \label{Pwpe}  $-\wpe^T\nabla_{w} L_0 \leq \begin{cases}
        \frac{|a_w|}{4}\mathbb{E}_{\xi}[\mathbf{1}(|w^T\xi| \in [\sqrt{2}\|\wsig\|, \sqrt{2}\|\wopp\|])|w^T\xi|] & \|\wopp\| > \|\wsig\|;\\
      -\frac{|a_w|}{4}\mathbb{E}_{\xi}[\mathbf{1}(|w^T\xi| \in [\sqrt{2}\|\wopp\|, \sqrt{2}\|\wsig\|])|w^T\xi|] & \|\wopp\| \leq \|\wsig\|.
    \end{cases}$
    \item \label{Pwinf} For $i \in [3, d]$, with $X = (\xi - e_i\xi_i)^Tw$, we have $$-w_i^T\nabla_{w_i} L_0 = \frac{|a_w||w_i|}{4}\left(\mathbb{P}\left[X \in [\sqrt{2}\|\wopp\| - |w_i|, \sqrt{2}\|\wopp\| + |w_i|]\right] - \mathbb{P}\left[X \in [\sqrt{2}\|\wsig\| - |w_i|, \sqrt{2}\|\wsig\| + |w_i|]\right]\right).$$
\end{enumerate}
\end{lemma}
\begin{proof}[Proof of Lemma~\ref{lemma:pop1}]
First consider \ref{Psig}. By symmetrizing over the pair $(z + \xi, -z +\xi)$, we have
\begin{align}
    -\wsig^T \nabla_{w} L_0 &= a_w\mathbb{E}_x y(x)\sigma'(w^Tx)z^T\wsig \\
    &= \frac{1}{2}a_w\mathbb{E}_{\xi} y(z)(\sigma'(w^T\xi + \wsig^Tz) - \sigma'(w^T\xi - \wsig^Tz))z^T\wsig\\
    &= \frac{1}{2}a_w\mathbb{E}_{\xi} y(z)\mathbf{1}(|w^T\xi| \leq |\wsig^Tz|)\on{sign}(\wsig^Tz)z^T\wsig\\
    &= \frac{\sqrt{2}}{4}|a_w|\mathbb{E}_{\xi}\mathbf{1}(|w^T\xi| \leq \sqrt{2}\|\wsig\|)\|\wsig\|
\end{align}
since $y(z)a_w > 0$ if $z \in \on{span}(\wsig)$.

Next consider \ref{Popp}. By a similar calculation via symmetrization, but using the fact that $y(z)a_w < 0$ if $z \in \on{span}(\wopp)$, we have 
\begin{align}
    -\wopp^T \nabla_{w} L_0 &= \frac{1}{2}a_w\mathbb{E}_{\xi} y(z)\mathbf{1}(|w^T\xi| \leq |\wopp^Tz|)\on{sign}(\wopp^Tz)z^T\wopp \\
    &= \frac{\sqrt{2}}{4}|a_w|\mathbb{E}_{\xi}\mathbf{1}(|w^T\xi| \leq \sqrt{2}\|\wopp\|)\|\wopp\|.
\end{align}

Next consider \ref{Pwpe}. Symmetrizing over the pair $(z + \xi, z - \xi)$, we have
\begin{align}
    -\wpe^T\nabla_{\wpe} L_0 &= a_w\mathbb{E}_x y(x)\sigma'(w^Tx)\wpe^T\xi \\
    &= a_w\frac{1}{2}\mathbb{E}_x y(z)\left(\sigma'(w^Tz + w^T\xi) - \sigma'(w^Tz - w^T\xi)\right)\wpe^T\xi \\
    &= a_w\frac{1}{2}\mathbb{E}_x y(z)\mathbf{1}(|w^T\xi| \geq |w^Tz|)|\wpe^T\xi|\\
    &= \begin{cases}
        |a_w|\frac{1}{4}\mathbb{E}_{\xi} \mathbf{1}(|w^T\xi| \in [\sqrt{2}\|\wopp\|, \sqrt{2}\|\wsig\|])|w^T\xi| & \|\wopp\| \leq \|\wsig\| \\
        -|a_w|\frac{1}{4}\mathbb{E}_{\xi} \mathbf{1}(|w^T\xi| \in [\sqrt{2}\|\wsig\|, \sqrt{2}\|\wopp\|])|w^T\xi| & \|\wsig\| \leq \|\wopp\|      
    \end{cases}
\end{align}

Finally consider \ref{Pwinf}. Recall that $\xii$ denotes $\xi - e_i \xi_i$. Then symmetrizing over the pair $(z + \xii + e_i\xi_i, z + \xii - e_i\xi_i)$, we have
\begin{align}
    -w_i^T\nlrz{w_i} &= -a_w \mathbb{E}_x y(x)\sigma'(w^Tx)w_i\xi_i \\
    &= -a_w \frac{1}{2}\mathbb{E}_{x} y(z)\left(\sigma'(w^Tz + w^T\xii + w_i\xi_i) - \sigma'(w^Tz + w^T\xii - w_i\xi_i)\right)w_i\xi_i\\
    &= -a_w \frac{1}{2}\mathbb{E}_{x} y(z)\mathbf{1}(|w^Tz + w^T\xii| \leq |w_i|)|w_i|
\end{align}

Now explicitly evaluating the expectation over $z$ and noting that the variable $\xii$ is symmetric, we have
\begin{align}
    a_w\mathbb{E}_{x} &y(z)\mathbf{1}(|w^Tz + w^T\xii| \leq |w_i|) \\
    &= |a_w|\frac{1}{2}\mathbb{P}_{\xii} \left[w^T\xii \in \left[\sqrt{2}\|\wsig\| - |w_i|, \sqrt{2}\|\wsig\| - |w_i|\right]\right]\\
    &\qquad - |a_w|\frac{1}{2}\mathbb{P}_{\xii} \left[w^T\xii \in \left[\sqrt{2}\|\wopp\| - |w_i|, \sqrt{2}\|\wopp\| - |w_i|\right]\right].\\
\end{align}

Thus with $X = w^T\xii$, we have
\begin{align}
    -w_i^T\nlrz{w_i} = \frac{|a_w||w_i|}{4}\left(\mathbb{P}_X\left[X \in [\sqrt{2}\|\wopp\| - |w_i|, \sqrt{2}\|\wopp\| + |w_i|]\right] - \mathbb{P}_X\left[X \in [\sqrt{2}\|\wsig\| - |w_i|, \sqrt{2}\|\wsig\| + |w_i|]\right]\right).
\end{align}
\end{proof}

To prove Lemma~\ref{lemma:graddiff1} and \ref{lemma:approxsig}, we will need the following lemma.
\begin{lemma}\label{lemma:delta}
    Suppose $x \sim \Rad^d$. Then
    \begin{align}
        \mathbb{E}_x (\ell'_{\rho}(x) - \ell'_0(x))^2 \leq 4(\mathbb{E}_{\rho}[\|a_w w\|])^2.
    \end{align}
    Further for any $x$ on the boolean hypercube and $i \in [d]$, 
    \begin{align}
        (\ell'_{\rho}(\ix + e_ix_i) - \ell'_{\rho}(\ix - e_ix_i))^2 \leq  16(\mathbb{E}_{\rho}[\|a_w w_i\|])^2.
    \end{align}

\end{lemma}
\begin{proof}
Recall that $\ell_{\rho}(x) = -2\log\left(\frac{1}{1 + \exp(-y(x)f_{\rho}(x))}\right)$, and so $\ell'_{\rho}(x) = -\frac{2y(x)\exp(-y(x)f_{\rho}(x))}{1 + \exp(-y(x)f_{\rho}(x))}$. Observe that $\ell'_{\rho}(x)$ is $2$-Lipshitz with respect to $f_{\rho}(x)$. Thus for the first statement, using Jensen's inequality, we have
\begin{align}
    \mathbb{E}_x (\ell'_{\rho}(x) - \ell'_0(x))^2 &\leq 4\mathbb{E}_x f_{\rho}(x)^2 \\
    &=4\mathbb{E}_{x}\left(\mathbb{E}_{\rho} a_w \sigma(w^Tx)\right)^2\\
    &=4\sup_{v: \|v\| = 1}\mathbb{E}_{x}\left(\mathbb{E}_{\rho}\|a_w w\| |v^Tx|\right)^2\\
    &= 4(\mathbb{E}_{\rho} \|a_w w\|)^2
\end{align}

For the second statement, by the $2$-Lipshitzness of $\ell'$, we have 
\begin{align}
     (\ell'_{\rho}(\ix + e_ix_i) - \ell'_{\rho}(\ix - e_ix_i))^2 &\leq 4  (f_{\rho}(\ix + e_ix_i) - f_{\rho}(\ix - e_ix_i))^2\\
     &= 4 \left(\mathbb{E}_{\rho} a_w (\sigma(w^T\ix + w_ix_i) - \sigma(w^T\ix - w_ix_i))\right)^2\\
    &\leq 4\left(\mathbb{E}_{\rho} 2 |a_w| |w_i|\right)^2\\
     &= 16(\mathbb{E}_{\rho}[\|a_w w_{i}\|])^2.
\end{align}
\end{proof}

\begin{proof}[Proof of Lemma~\ref{lemma:graddiff1}]
For convenience, define $\Delta_x := (\ell'_{\rho}(x) - \ell'_{0}(x))\sigma'(w^Tx)$. 
We consider item~\ref{Ga} first. By Cauchy Schwartz, we have
\begin{align}\label{eq:ga}
    \left|\nabla L_0(a_w ) -  \nabla L_{\rho}(a_w )\right| &= \left|\mathbb{E}_x(\ell'_{\rho}(x) - \ell'_{0}(x))\sigma(w^Tx)\right|\\
    &= \left|\mathbb{E}_x\Delta_x w^Tx\right|\\
    &\leq \sqrt{\mathbb{E}_x[\Delta_x^2]} \sqrt{\mathbb{E}_x (w^Tx)^2}\\
    &\leq \sqrt{\mathbb{E}_x[(\ell'_{\rho}(x) - \ell'_{0}(x))^2]} \|w\|_2.
\end{align}
Now by Lemma~\ref{lemma:delta}, we have 
$\mathbb{E}_x[(\ell'_{\rho}(x) - \ell'_{0}(x))^2] \leq 4p^2(\mathbb{E}_{\rho}[\|a_w w\|])^2$. Plugging this yields \ref{Ga} and \ref{Gw}.

For item~\ref{Gw}, we similarly have
\begin{align}\label{eq:gw}
    \|\nabla_w L_{\rho} -  \nabla_w L_0\|_2 & = |a_w|\|\mathbb{E}_x \Delta_x x\| \\
    &=|a_w|\sup_{v : \|v\| = 1} \mathbb{E}_x \Delta_x \langle{v, x\rangle} \\
    &\leq |a_w|\sup_{v : \|v\| = 1} \sqrt{\mathbb{E}_x \Delta_x^2} \sqrt{\mathbb{E}_x\langle{v, x\rangle}^2}\\
    &\leq |a_w|\sqrt{\mathbb{E}_x \Delta_x^2} \\
    &\leq |a_w|2\mathbb{E}_{\rho}[\|a_w w\|].
\end{align}

\end{proof}

\begin{proof}[Proof of Lemma~\ref{lemma:approxsig}]
Define $\Delta_x := (\ell'_{\rho}(x) - \ell'_{0}(x))\sigma'(w^Tx)$.  
Using the symmetry of the data for pairs $(\ix + e_ix_i, \ix - e_ix_i)$, we have
\begin{align}
    &\left\|\frac{1}{a_w}\left(\nabla_{w_{i}} L_{\rho} - \nabla_{w_{i}} L_{0}\right)\right\|\\
    &\qquad = \|\mathbb{E}_x \Delta_x x_i\| \\
    &\qquad = \frac{1}{2}\|\mathbb{E}_x (\Delta_{\ix + e_ix_i} - \Delta_{\ix - e_ix_i})x_i\| \\
    &\qquad \leq \frac{1}{2}\|\mathbb{E}_{x} \mathbf{1}(|\ix^Tw| \geq |w_i|)(\Delta_{\ix + e_ix_i} - \Delta_{\ix - e_ix_i})x_i\| + \frac{1}{2}\|\mathbb{E}_{x}\mathbf{1}(|\ix^Tw| \leq |w_i|)(\Delta_{\ix + e_ix_i} - \Delta_{\ix - e_ix_i})x_i\|\\
    &\qquad \leq \frac{1}{2}\sup_{x}|\ell'_{\rho}(\ix + e_ix_i) - \ell'_{\rho}(\ix - e_ix_i)| + \mathbb{E}_{x}\mathbf{1}(|\ix^Tw| \leq |w_i|)|\Delta_{x}|.
\end{align}
Here the last line follows from the fact that whenever $|\ix^Tw| \geq |w_i|$, we have $\sigma'(w^T(\ix + e_ix_i)) = \sigma'(w^T(\ix - e_ix_i)),$ and thus $|\Delta_{\ix + e_ix_i} - \Delta_{\ix - e_ix_i}| \leq |\ell'_{\rho}(\ix + e_ix_i) - \ell'_{\rho}(\ix - e_ix_i)|$. Note that the $\sup$ is over $x$ on the boolean hypercube.

Now by Lemma~\ref{lemma:delta}, we have $\sup_{x} |\ell'_{\rho}(\ix + e_ix_i) - \ell'_{\rho}(\ix - e_ix_i)| \leq 4(\mathbb{E}_{\rho}[\|a_w w_{i}\|])$. Further, by the $2$-Lipshitzness of $\ell'_{\rho}$ with respect to $f_{\rho}(x)$, (see the proof of Lemma~\ref{lemma:delta}), we have

\begin{align}\label{eq:diarmid}
   \mathbb{E}_{x}&\mathbf{1}(|x^T(w - w_ix_i)| \leq |w_i|)|\Delta_{x}|\\
   &\leq 2\mathbb{E}_{x}\mathbf{1}(|x^T(w - w_ix_i)| \leq |w_i|)|f_{\rho}(x)|\\
    &\leq 2\mathbb{E}_{x}\mathbf{1}(|x^T(w - w_ix_i)| \leq |w_i|)\log(d)\mathbb{E}_{\rho}[\|a_w w\|] + 2\left(
    \sqrt{d}\mathbb{E}_{\rho}[\|a_w w\|]\right)\mathbb{P}_x[|f_{\rho}(x)| \geq \log(d)\mathbb{E}_{\rho}[\|a_w w\|]]\\
    &\leq 2\mathbb{E}_x\mathbf{1}(|x^T(w - w_ix_i)| \leq |w_i|)\log(d)\mathbb{E}_{\rho}[\|a_w w\|] + d^{-\omega(1)},
\end{align}

where the last line follows from McDiarmid's inequality of bounded differences.
Thus putting these pieces together,
\begin{align}
    &\|\nabla_{w_i} L_{\rho} - \nabla_{w_i} L_0\|\\ &\quad\leq |a_w|\left(4\mathbb{E}_{\rho}[\|a_w w_{i}\|] + 2\log(d)\mathbb{E}_{\rho}[\|a_w w\|]\mathbb{E}_{x}\mathbf{1}(|\ix^Tw| \leq |w_i|) + d^{-\omega(1)}\right),
\end{align}
which yields the lemma.
\end{proof}
We additionally state and prove the following helper lemma that will be used in Phase 1.
\begin{lemma}[Helper Lemma]\label{lemma:helper}
Suppose for some vector $u_t$ and reals $0 \leq Q_t \leq B_t < 1$, we have $\|u_t\|^2 \leq Q_t^2 + \theta B_t^2$. Also suppose that for some vectors $G$ and $\hat{G}$ and some $\chi > \theta^{1/2}$:
\begin{enumerate}
    \item $-u_t^T G \leq \chi\left(Q_t^2 + \theta B_t^2\right)$
    \item $\|G - \hat{G}\| \leq \theta B_t$
    \item $\|G\| \leq O(B_t)$.
    \item $\eta \leq \theta^{2} = o(1)$.
\end{enumerate}
Then with $u_{t + 1} := u_t - \eta \hat{G}$, we have 
\begin{align}
    \|u_{t + 1}\|^2 \leq (Q_t^2 + \theta B_t^2)(1 + 5\eta \chi).
\end{align}
\end{lemma}
\begin{proof}
Define $W_t^2 :=  Q_t^2 + \theta^{1/2} B_t^2$, such that $B_t \leq W_t \theta^{-1/2}$.
\begin{align}
    \|u_{t + 1}\|^2 &= \|u_t - \eta \hat{G}\|^2\\
    &\leq \|u_t\|^2 - 2\eta u_t^TG + 2\eta \|u_t\|\|G - \hat{G}\| + 2\eta^2 \|G - \hat{G}\|^2 + 2\eta^2\|G\|^2\\
    &\leq  W_t^2(1 + 2\eta \chi) + 2\eta \|u_t\| \theta B_t + O(\eta^2B_t^2)\\
    &\leq  W_t^2(1 + 2\eta \chi) + 3\eta \|W_t\| \theta B_t\\
    &\leq W_t^2\left(1 + 2\eta \chi + 3\eta \theta^{1/2}\right)\\
    &\leq  W_t^2\left(1 + 5\eta \chi\right).
\end{align}
\end{proof}
    

\subsection{Inductive Lemmas, and Proof of Lemma~\ref{lemma:phase1formal} assuming Inductive Lemmass}
We now give a short sketch of the analysis in Phase 1 used to prove Lemma~\ref{lemma:phase1formal}.
Let $\zeta = 1/\log^c(d)$ and $\theta = 1/\log^C(d)$, where $c$ and $C$ are sufficiently large constants. While we will omit stating it explicitly, in all the lemmas henceforth in Section~\ref{apx:phase1}, it is assumed that first $c$ is chosen to be a sufficiently large constant, and then $C$ is chosen to be sufficiently large in terms of $c$. 

Phase 1 will be broken down into two sub-phases, 1a and 1b. The analysis in both sub-phases is quite similar, but our approximation of the gradients will be courser in Phase 1b than in 1a. Phase 1a will last for most of the time (some $\Ta = \Theta(\log(d))$ iterations), and and the end of the phase, we will guarantee the existence of a substantial set of ``strong'' neurons (see Definition~\ref{def:strong}) for which $\zeta^{1.5} \|w\| \leq \|\wsig\| \leq \|w\|$. Note that this is a very meaningful guarantee, since at initialization we have $\|\wsig\| \approx \frac{1}{\sqrt{d}}\|w\|$, and $\zeta$ is $1/\pl(d)$. Phase 1b will last for only some $\Tb = \Theta(\log\log(d))$ iterations, enough to guarantee that on some set of strong neurons, we have $\|\wsig\| \geq \|\wpe + \wopp\|/\zeta$. This will suffice to prove Lemma~\ref{lemma:phase1formal}.

To formalize this, we state three definitions will will be the basis of our inductive analysis for Phase 1. Our goal will to be to show that all neurons are ``controlled'' or ``weakly controlled'', meaning $\wopp$ and $\wpe$ don't grow too large, while a substantial fraction of neurons are ``strong'', and in these neurons, $\wsig$ grows quickly.

In what follows, we define the rate parameter $\tau := \frac{1}{\sqrt{2\pi}}$ to be the approximate rate at which the neurons near initialization would grow at under the $L_0$ population loss if $|a_w | = \|w\|$.
To see this, observe that from Lemma~\ref{lemma:pop1}, we have
\begin{align}\label{eq:tau}
    \frac{-\wsig^T\nabla_{\wsig} L_0 }{\|\wsig\|^2} &= \frac{|a_w |}{\|\wsig\|}\frac{\sqrt{2}}{4}\mathbb{P}_{\xi} \mathbf{1}(|w^T\xi| \leq \sqrt{2}\|\wsig\|) \\
    &\approx \frac{|a_w |}{\|\wsig\|}\frac{\sqrt{2}}{4}\frac{2\sqrt{2}\|\wsig\|}{\sqrt{2\pi}\|w\|}\\
    &= \frac{1}{\sqrt{2\pi}}.
\end{align}
Note that the ``$\approx$'' approximation step will hold under the conditions that $\|\wsig\| \ll \|\wpe\| \approx |a_w|$, and that the vector $\wpe$ is well-spread among its coordinates -- that is, none of its coordinates in the standard basis are too large, which could preclude the central limit theorem convergence of $w^T\xi \rightarrow \mathcal{N}(0, \|\wpe\|)$ in distribution. The details of the comparison of the probability over the boolean vector to the analogous probability over a Gaussian vector is fleshed out in Section~\ref{sec:boolean}.

This calculation gives some intuition for two conditions that we will maintain in our definition of controlled neurons for Phase 1a: we should have $|a_w | \approx \|\wpe\|$, and $\wpe$ should be well-spread in some sense, which we will enforce by requiring that $\wpe^{(t)} - \wpe^{(0)}$ and $\|\wpe\|_{\infty}$ are small. 
We define the following control parameters.
\begin{definition}[Control Parameters]\label{def:control}
Let $\Ta$ and $\Tb$ be as defined in Definition~\ref{def:T}. Define
\begin{align}\label{BQdef}
    B_t^2 &:= \begin{cases}
        \frac{\log^3(d)\theta^2}{d}(1 + 2\eta \tau(1 + 1/\log(d)))^t & t \leq \Ta; \\
         \frac{\log^3(d)\theta^2}{d}(1 + 2\eta \tau(1 + 1/\log(d)))^{\Ta}(1 + 4\eta)^{t - \Ta}\zeta^{-2} & \Ta \leq t \leq \Tb.
    \end{cases}\\
    Q_t^2 &:= \frac{\log^3(d)\theta^2}{d}\left(1 + \frac{50\eta}{\log(d)}\right)^t.
\end{align}  
Let $\cws$ be the universal constant which is the maximum of the constants in Lemmas~\ref{lemma:spread_init}, \ref{lemma:infinity}, and \ref{lemma:boolean2}.
\end{definition}

\begin{definition}[Phase 1 Length]\label{def:T}
Let $\Ta$ be the last time at which we have
$B_t^2 \leq \theta^2\zeta^2$, that is,
\begin{align}
    \Ta := \left\lfloor{\frac{\log(d) + 2\log(\zeta) - 3\log(\log(d))}{\log(1 + 2\eta \tau(1 + 1/\log(d)))}}\right\rfloor = \frac{1}{\eta}\Theta(\log(d)).
\end{align}

Let $\Tb$ to be the last time at which we have $B_t^2 \leq \theta^2\zeta^{-600}$, that is, 
\begin{align}\label{eq:T1b}
    \Tb := \Ta + \left\lfloor\frac{\log(\theta^2\zeta^{-598}/B_{\Ta}^2)}{\log(1 + 4\eta)}\right\rfloor = \Ta + \frac{1}{\eta}\Theta(\log\log(d)).
\end{align}
\end{definition}

\begin{definition}[Controlled Neurons]\label{def:controlled}
We say a neuron $(w, a_w )$ is \em controlled \em at iteration $t \leq \Tb$ if:
\begin{enumerate}[{\bfseries{C\arabic{enumi}}}]
    \item\label{Csig} $\|\wsig\|^2 \leq \min(B_t^2, \theta^2 \zeta^2)$.
    \item\label{Copp} $\|\wopp\|^2 \leq Q_t^2 + \theta B_t^2$ 
    \item\label{Ca} $|a_w | \in \theta(1 \pm t\eta\zeta)$, and $|a_w| \leq \|w\|$.
    \item\label{Cwpe} $\|\wpe - \wpe^{(0)}\| \leq \theta \zeta^{1/4} \eta t$ , and $\wpe^{(0)}$ is $\cws$-well spread (see Definition~\ref{def:wellspread}). 
    \item\label{Cwinf} $\|\wpe\|^2_{\infty} \leq  Q_t^2 + \theta B_t^2$.
\end{enumerate}
\end{definition}

In Phase 1b, we will need to consider the case where $\|\wsig\|$ grows larger than $\theta \zeta$ for some neurons. Thus we introduce the following definition of ``weakly controlled'' neurons.

\begin{definition}[Weakly Controlled Neurons]\label{def:weakcontrol}
We say a neuron $(w, a_w )$ is \em weakly controlled \em at iteration $t \in [\Ta, \Tb]$ if:
\begin{enumerate}[{\bfseries{W\arabic{enumi}}}]
    \item\label{Wwsig} $\theta^2\zeta^2 \leq \|\wsig\|^2 \leq B_t^2 \leq \theta^2 \zeta^{-600}$.
    \item\label{Wwopp} $\|\wopp\|^2 \leq 2\theta B_t^2 (1 + 3\eta \zeta)^{t} \leq 4\theta^2\zeta^2$.
    \item \label{Wa} $\|w\|^2 \geq |a_w|^2 \geq \|w\|^2 - \zeta^{1/2}\theta^2 - 8\eta^2 (t - \Ta)\theta^2 \zeta^{-600} $.
    \item \label{Wwpe} $\|\wpe\|^2 \leq 2\theta^2 (1 + 3\eta \zeta)^t \leq 3\theta^2$.
    \item\label{Winf} Either we have $\|\wpe\| \leq \|\wsig\|$, or $\|\wpe\|_{\infty} \leq \zeta^{\cbe 10000} \theta (1 + 21\cbe \eta)^{t - \Ta} \leq \zeta^{1000} \|\wpe\|_2$,
where $\cbe$ is the universal constant from Theorem~\ref{thm:be}.
\end{enumerate}
\end{definition}
We note the following simple claims which can be verified by plugging in the values $\Ta$ and $\Tb$, and recalling that $\zeta = \log^{-c}(d)$ and $\theta = \log^{-C}(d)$ for $c$ and $C$ sufficiently large.
\begin{claim}\label{claim:cont}
For any $t \leq \Tb$, conditions \ref{Copp}-\ref{Cwinf} of Definition~\ref{def:controlled} imply conditions \ref{Wwopp}-\ref{Winf} of Definition~\ref{def:weakcontrol}.
\end{claim}

\begin{claim}\label{claim:implication}
If all neurons are controlled or weakly controlled at time $t$, then
    \begin{enumerate}
        \item $\mathbb{E}_{\rho_t}[\|a_w w\|] \leq 2\max\left(\theta^2, B_t^2\right) \leq 2\theta^2 \zeta^{-600}$.
        \item For any $i \in [d]$, $\mathbb{E}_{\rho_t}[\|a_w w_{i}\|] \leq 3\max\left(\theta B_t, B_t^2\right)$.
    \end{enumerate}
\end{claim}

We now define strong neurons, which is the set of neurons on which $\|\wsig\|$ grows quickly.
\begin{definition}[Strong Neurons]\label{def:strong}
We say a neuron $(w, a_w )$ is \em strong \em at iteration $t$ if it is controlled or weakly controlled, $\wsig^T\wsig^{(0)} > 0$, and 
\begin{align}\|\wsig\|^2 \geq S_t^2 := \frac{\theta^2}{d}\prod_{s \leq t}\left(1 + 2\eta \tau (1 - \eps_s) \right)^{s},
\end{align}
where 
\begin{align}
    \eps_s := \begin{cases}
        1 - \frac{1}{\cs} & s \leq \Ts; \\
        5\zeta^{1/10} + \frac{200\cbe \sqrt{\pi}}{(1 + 2\eta \tau/\cs)^{s/2}} & \Ts < s \leq \Ta; \\
        1 - \frac{1}{20} & \Ta \leq s \leq \Tb,
    \end{cases}
\end{align}
and we have defined the universal constant $\cs := \frac{6400}{\sqrt{\pi}}\exp(100\cws^8)$.
\end{definition}

While the definition of a strong neuron is technical, the meaning is that $\|\wsig\|^2$ grows roughly at the rate of $(1 + 2\eta \tau)^{t}$. Indeed, this is the case when $\epsilon_s$ is small, which is true in the middle range of values $s$ above, which covers most of the iterations. (The fact that $\epsilon_s$ is constant for small $s$ comes from some errors derived in comparing probabilities of events on Boolean vectors to their Gaussian counterparts; see Section~\ref{sec:boolean}). 

We have the following implication of the definition of a strong neuron.

\begin{lemma}\label{lemma:St}
For $t \leq \Ta$, we have $S_t^2 \geq B_t^2/\log^4(d)$. Thus for a strong neuron, after $\Ta$ steps, we have $\|\wsig^{(\Ta)}\| \geq \zeta^{1.5}\theta$. Further, after $\Tb$ steps, we have $\|\wsig^{(\Tb)}\| \geq \zeta^{-1}\theta$.
\end{lemma}
\begin{proof}
We check the first statement first.
    \begin{align}
    \frac{B_t^2}{S_t^2} &\leq \log^3(d)\prod_{s = 1}^t \frac{1 + 2\eta\tau \left((1 + 1/\log(d)\right) }{1 + 2\eta \tau(1 - \epsilon_s)} \tagblue{Defs.~\ref{def:strong}, \ref{def:control}}\\
    &\leq  \log^3(d)\prod_{s = 1}^t \left(1 + 2\eta\tau \left((1 + 1/\log(d)\right)\right)\left(1 - 2\eta \tau(1 - \epsilon_s) + 4\eta^2\tau^2\right) \tagblue{$\frac{1}{1 + q} \leq 1 - q + q^2 \quad \forall q > 0$}\\
    &\leq \log^3(d)\prod_{s = 1}^t \left(1 + 4\eta \tau \left(1/\log(d) + \eps_s\right)\right) \tagblue{$\eta = o(1)$}\\
    &\leq \log^3(d) \exp\left(4\eta \tau \sum_{s = 1}^t 1/\log(d) + \epsilon_s \right) \tagblue{$1 + q \leq e^q \quad \forall q > 0$} \\
    &= \Theta\left(\log^3(d)\right) \leq \log^4(d).
\end{align}
Here the last line follows from the fact that $\epsilon_s$ is constant for $\Theta(1/\eta)$ iterations, and then it is exponentially decaying down to a minimum of $5\zeta^{1/10}$. Thus since $t \leq \Ta = \Theta(
\log(d)/\eta)$,the sum is $\Theta(1/\eta)$.

Thus after $\Ta$ steps, since $\zeta = o(\log^{-4}(d))$, we have 
\begin{align}
    \|\wsig^{(\Ta)}\| \geq \frac{B_t}{\log^2(d)} \geq \frac{\zeta \theta}{2\log^2(d)} \geq \zeta^{1.5}\theta.
\end{align}

For the second statement, we consider $\|\wsig^{(\Tb)}\|$. Observe that
\begin{align}
    \frac{B_{\Tb}^2}{S_{\Tb}^2} &= \frac{B_{\Ta}^2}{S_{\Ta}^2}\zeta^{-2}\prod_{s = \Ta + 1}^{\Tb}\frac{1 + 4\eta}{1 + \eta \tau/10} \tagblue{Defs.~\ref{def:strong}, \ref{def:control}}\\
    &\leq \frac{B_{\Ta}^2}{S_{\Ta}^2}\zeta^{-2}\prod_{s = \Ta + 1}^{\Tb} (1 + 4\eta)^{0.99}\\
    &= \frac{B_{\Tb}^2}{S_{\Ta}^2}\left(\frac{B_{\Tb}^2}{B_{\Ta}^2}\right)^{-0.01}\zeta^{-0.02} \tagblue{Def.~\ref{def:control}}\\
    &\leq \frac{B_{\Tb}^2}{S_{\Ta}^2} \zeta ^{5.5}, \tagblue{Def.~\ref{def:T}}
\end{align}
and thus $S^2_{\Tb} \geq S^2_{\Ta} \zeta^{-5.5} \geq (\theta^2 \zeta^{3})\zeta^{-5.5} \geq \theta^2 \zeta^{-2}$. It follows that $\|\wsig^{(\Tb)}\| \geq \theta \zeta^{-1}$.
\end{proof}

The first main lemma for Phase 1 is the following inductive step.
\begin{lemma}[Controlled Neurons Inductive Step]\label{lemma:inductive1aALT}
Suppose for some $t \leq \Tb$, all neurons are controlled or weakly controlled. Then with probability at least $1 - d^{-\omega(1)}$, for any neuron $(w^{(t)}, a_w^{(t)})$ which is controlled, at step $t + 1$:
\begin{enumerate}
    \item The neuron $(w^{(t + 1)}, a_w^{(t + 1)})$ is either controlled or weakly controlled.
    \item If $(w^{(t)}, a_w^{(t)})$ is strong, then $(w^{(t + 1)}, a_w^{(t + 1)})$ is strong.
\end{enumerate}
\end{lemma}

The following is our main inductive step for the second half of Phase 1.
\begin{lemma}[Weakly Controlled Neurons Inductive Step]\label{lemma:inductive1b}
Suppose for some $\Ta \leq t \leq \Tb$, all neurons are controlled weakly controlled. Then with probability at least $1 - d^{-\omega(1)}$, at step $t + 1$, all neurons are controlled or weakly controlled, and any strong neuron remains strong.
\end{lemma}

We defer the proofs of Lemmas~\ref{lemma:inductive1aALT} and \ref{lemma:inductive1b} to the following subsection. Assuming Lemma~\ref{lemma:inductive1b} we can now prove Lemma~\ref{lemma:phase1formal}, which we restate here for the readers convenience.

\lemphase*

\begin{proof}[Proof of Lemma~\ref{lemma:phase1formal}]
Choose $T_1 = \Ta + \Tb$. First we confirm that at initialization, all neurons are controlled (Definition~\ref{def:controlled}), and at least a $0.1$ fraction of neurons are strong (Definition~\ref{def:strong}). Indeed for any neuron $w$, by the Guassian CDF, with probability $1 - d^{-\omega(1)}$, we have $|w_i| \leq \theta\frac{\log(d)^2}{\sqrt{d}}$. Taking a union bound over all $i$ and over all at most polynomially many neurons yields properties \ref{Csig}, \ref{Copp}, and \ref{Cwinf} for all neurons. Property \ref{Ca} holds for all neurons since $a_w$ is initialized to have norm $1$. Finally, with probability $1 - d^{-\omega(1)}$, by Lemma~\ref{lemma:spread_init}, all neurons are well-spread at initialization, yielding \ref{Cwpe}. Now we can applying Chernoff's bound to bound the number of neurons for which $\|\wsig^{(0)}\| \geq \frac{\theta}{\sqrt{d}}$, yielding a $0.1$ fraction of neurons with $\|\wsig^{(0)}\| \geq \frac{\theta}{\sqrt{d}}$ and $\wsig^T\mu > 0$ for each $\mu \in \muset$.

Now with probability $1 - d^{-\omega(1)}$, after $\Ta$ steps, Lemma~\ref{lemma:inductive1aALT} guarantees that we have a network for which all of the neurons are controlled or weakly controlled  (\ref{def:weakcontrol}), and for each $\mu \in \muset$, at least a $0.1$ fraction of the neurons are strong and $\wsig^T\mu > 0$. 

Now applying Lemma~\ref{lemma:inductive1b} $\Tb - \Ta$ times yields that with probability $1 - d^{-\omega(1)}$, after $\Tb$ steps, for each $\mu$, we have at least a $0.1$ fraction of the neurons are strong and $\wsig^T\mu > 0$, and all neurons are controlled or weakly controlled. 

We can now conclude the first item of the lemma, which bounds $\mathbb{E}_{\rho_{T_1}}[\|a_w w\|]$ from properties \ref{Wwsig}, \ref{Wwopp}, and \ref{Wa}, \ref{Wwpe}. The second item, which bounds $\mathbb{E}_{\rho_{T_1}}[\|\wpe + \wopp\|^2]$ follows from properties \ref{Wwopp} and \ref{Wwpe}. Note, if any of the neurons are controlled (instead of weakly controlled), the same bounds hold from the respective properties of controlled neurons. The third item follows by considering the set of strong neurons, and observing that on strong neurons, by Lemma~\ref{lemma:St}, we have
\begin{align}
    \|\wsig\| \geq \zeta^{-1}\theta \geq \log(d)^c \theta.
\end{align}
Finally, the additional clause follows from properties \ref{Ca} and \ref{Wa}.
\end{proof}

\subsection{Proof of Inductive Lemmas (Lemmas~\ref{lemma:inductive1aALT} and \ref{lemma:inductive1b})}

To prove Lemma~\ref{lemma:inductive1aALT}, we will need to compute several bounds on the $L_0$ population gradients on controlled neurons. Recall that most of the gradients have been computed already in Lemma~\ref{lemma:pop1}; the following lemma just gives some additional bounds that hold for controlled neurons.
\begin{lemma}[Phase 1a $L_0$ Population Gradients Bounds]\label{lemma:pop1a}
If all neurons in the network are controlled or weakly controlled at some step $t \leq \Tb$, then for any controlled neuron $(w, a_w )$,
\begin{enumerate}[{\bfseries{A\arabic{enumi}}}]
\item \label{Awpe}
    $\|\nlrz{\wpe}\| \leq 3 \min\left(\sqrt{\theta B_t}, \theta \zeta^{1/2} \right)$.
\item \label{Aa}
   $|\nabla_{a_w} L_0| \leq 4\theta\min\left(\sqrt{\theta B_t}, \theta \zeta^{1/2}\right).$ 
\item \label{A12} For any $i \in [d]$, $\|\nabla_{w_{i}} L_0\| \leq \frac{1}{2}\min\left(B_t, \theta \zeta\right)$.
\item \label{Awinf} For any $i \in [d]$, $\|\nabla_{w_i} L_0 - \nabla_{w_i} L_{\rho}\| \leq \frac{1}{2}\theta B_t$.
\end{enumerate}

\end{lemma}
\begin{proof}
We begin with \ref{Awpe}. 
Recalling that $x = z + \xi$ for $z = x_{1:2}$, we have
\begin{align}
    \frac{1}{a_w }\nabla_{\wpe} L_0 &= \mathbb{E}_x -y(x)\sigma'(w^Tx)\xi \\
    &=  -\mathbb{E}_x y(z)\sigma'(w^T\xi)\xi + \mathbb{E}_x y(z)(\sigma'(w^Tx)-\sigma'(w^T\xi))\xi\\
    &= \mathbb{E}_x y(x)(\sigma'(w^Tx)-\sigma'(w^T\xi))\xi,
\end{align}
since $y(z)$ is independent of $\xi$. 

Now consider the norm of $\mathbb{E}_x y(x)(\sigma'(w^Tx)-\sigma'(w^T\xi))\xi$. We have 
\begin{align}
    \|\mathbb{E}_x y(x)(\sigma'(w^Tx)-\sigma'(w^T\xi))\xi\| &= \sup_{v : \|v\| = 1} \mathbb{E}_x y(x)(\sigma'(w^Tx)-\sigma'(w^T\xi))\xi^Tv \\
    &\leq \sqrt{\mathbb{E}_x (\sigma'(w^Tx)-\sigma'(w^T\xi))^2}\sqrt{\mathbb{E}_{\xi} (v^T\xi)^2}\\
    &= \sqrt{\mathbb{E}_x \mathbf{1}(|\xi^Tw| \leq |z^Tw|)}\\
    &\leq \sqrt{\mathbb{P}_{\xi}[|\xi^Tw| \leq \sqrt{2}\|w_{1:2}\|]}.
\end{align}
Now since $(w, a_w)$ is controlled, the property \ref{Cwpe} and Lemma~\ref{lemma:boolean2} (plugging in $v = \wpe^{(0)}$, $\Delta = \wpe^{(t)} - \wpe^{(0)}$, and $a = \frac{\sqrt{2}\|w_{1:2}\|}{\|v\|} \ll 1$) guarantees that 
\begin{align}
    \mathbb{P}_{\xi}[|\xi^Tw| \leq \sqrt{2}\|w_{1:2}\|] &\leq 2\mathbb{P}_{G \sim \mathcal{N}(0, 1)}[|G| \leq a] + 200\cbe d^{-1/2}\\
    &\leq \frac{\|w_{1:2}\|}{\theta} + 200\cbe d^{-1/2}.
\end{align}
Thus we have 
\begin{align}
    \|\nabla_{\wpe} L_0\| \leq 2\sqrt{\theta \|w_{1:2}\|} + 15|a_w|\sqrt{\cbe}d^{-1/4} \leq 3\min\left(\sqrt{\theta B_t}, \theta \zeta^{1/2}\right),
\end{align} where we have used the fact that since the neuron is controlled, $|a_w| \leq 2\theta$.

Next consider \ref{A12}. By Lemma~\ref{lemma:pop1}, we have
\begin{align}
    \|\nabla_{w_{i}} L_0\| &\leq \frac{1}{4}|a_w|\left(\mathbb{P}_{x}\mathbf{1}(|w^T\ix| \leq |w_i|)\right)\\
    &\leq \frac{|a_w|}{4}\left(\frac{|w_i|}{\|w^{(0)}\|} + 200\cbe d^{-1/2}\right)\\
    &\leq \frac{1}{3}|w_i| + 50\cbe d^{-1/2}\theta\\
    &\leq \frac{1}{2}\min\left(B_t, \theta \zeta\right),
\end{align}
where the third to last line follows from Lemma~\ref{lemma:boolean2} and plugging in the Gaussian density $P_a := \mathbb{P}_{G \sim \mathcal{N}(0, 1)}[|G| \leq a] \leq \sqrt{\frac{2}{\pi}}a$. The final inequality follows from \ref{Csig}, \ref{Copp}, and \ref{Cwinf}.

Next consider \ref{Aa}. Combining \ref{A12} and \ref{Awpe}, we have
\begin{align}
    |\nabla_{a_w} L_0| &= |w^T \nabla_{w} L_0| \\
    &\leq |\wpe^T \nabla_{\wpe} L_0| + |w_{1:2} \nabla_{\wpe} L_0|\\
    &\leq \|w\|\|\nabla_{\wpe} L_0\| + \|w_{1:2}\|\|\nabla_{w_{1:2}} L_0\| \\
    &\leq 3\|w\|\min\left(\sqrt{\theta B_t}, \theta \zeta^{1/2}\right) + \|w\|\min\left(B_t, \theta \zeta\right) \\
    &\leq 4\theta\min\left(\sqrt{\theta B_t}, \theta \zeta^{1/2}\right).
\end{align}

Finally consider \ref{Awinf}. Applying Lemma~\ref{lemma:approxsig} and then Claim~\ref{claim:implication} yields
\begin{align}\label{eq:as}
    &\|\nabla_{w_{i}} L_{\rho} - \nabla_{w_{i}} L_0\|\\
    &\qquad \leq  |a_w|\left(4(\mathbb{E}_{\rho}[\|a_w w_{i}\|]) + 2\log(d)\mathbb{E}_{\rho}[\|a_w w\|]\mathbb{E}_{\xi}\mathbb{E}_z\mathbf{1}(|\ix^Tw| \leq |w_i|) + d^{-\omega(1)}\right)\\
    & \qquad \leq 2\theta\left(12 B_t\max(\theta, B_t) + 4\log(d)\max(\theta^2, B_t^2)\mathbb{E}_{\xi}\mathbb{E}_z\mathbf{1}(|\ix^Tw| \leq |w_i|) + d^{-\omega(1)}\right).
\end{align}

Now $\wpe^{(0)}$ is well-spread, and $\|\wpe^{(t)} - \wpe^{(0)}\| \leq o(1)\|\wpe^{(0)}\|$ by the definition of controlled. Letting $v = \wpe^{(0)}$ and $\Delta = \wpe^{(t)} - \wpe^{(0)}$, plugging in Lemma~\ref{lemma:boolean2} we obtain
\begin{align}
    \mathbb{E}_{x}\mathbb{E}_z\mathbf{1}(|\ix^Tw| \leq |w_i|) &\leq \mathbb{P}_{G \sim \mathcal{N}(0, 1)}\left[|G| \leq \frac{|w_i|}{\|w - e_iw_i\|}\right](1 + o(1)) + 200\cbe d^{-1/2}\\
    &\leq \frac{|w_i|}{\|w\|} + 200\cbe d^{-1/2}\\
    &\leq \frac{2\min(B_t, \theta \zeta)}{\theta}\\
    &\leq \frac{2B_t}{\max(\theta, B_t)}
\end{align}
where the second to last line follows from the definition of controlled and of $B_t$ (eq.~\ref{BQdef}).
Thus returning to Equation~\ref{eq:as}, and using Claim~\ref{claim:implication}, we have 
\begin{align}
    \|\nabla_{w_{i}} L_{\rho} - \nabla_{w_{i}} L_0\| &\leq \Theta(\log(d))\left(\theta B_t \max(\theta, B_t) + \theta B_t \max(\theta, B_t) \right) \leq \theta B_t/2,
\end{align}
since $\max(\theta, B_t) = o(1/\log(d))$.

\end{proof}

We are now ready to prove the inductive step, Lemma~\ref{lemma:inductive1aALT}.
\begin{proof}[Proof of Lemma~\ref{lemma:inductive1aALT}]
Suppose that $(w^{(t)}, a_w^{(t)})$ is controlled. Our first goal will be to show that for $(w^{(t + 1)}, a_w^{(t + 1)})$, items \ref{Ca}, \ref{Cwpe}, \ref{Copp}, and \ref{Cwinf} hold. We will handle $\|\wsig\|$ at the end. 

Next we prove that \ref{Ca}. To prove that $|a_w| \in \theta (1 \pm t\eta \zeta)$ holds at the $(t + 1)$th step, we have with probability $1- d^{-\omega(1)}$
\begin{align}
    |a_w^{(t + 1)} - a_{w}^{(t)}| &=|\eta \nabla \hat{L}_{\rho}a_w^{(t)}|\\
    &\leq  \eta |\nabla_{a_w} L_0| + \eta |\nabla_{a_w} L_{\rho} - \nabla_{a_w} \hat{L}_{\rho}| + \eta|\nabla_{a_w} L_0 - \nabla_{a_w} L_{\rho}|\\
    &\leq \eta\left(4\theta^2 \zeta^{1/2}  + \|w\|\sqrt{\frac{d\log(d)^2}{m}} + 4\|w\|\mathbb{E}_{\rho}[\|a_w w\|] \right) \tagblue{Lemma~\ref{lemma:pop1a} (item \ref{Aa}), \ref{lemma:emp}, and \ref{lemma:graddiff1} respectively}\\
    &\leq \eta \theta \zeta. \tagblue{$\frac{d}{m} \leq \theta^2$, Claim~\ref{claim:implication}}
\end{align}
Here the final inequality we also used the definition of controlled to bound $\|w\|$. With probability $1 - d^{-\omega(1)}$, by Lemma~\ref{lemma:layer_balance} \ref{Sb1}, we have $|a_w^{(t + 1)}| \leq \|w^{(t + 1)}\|$.

To prove that \ref{Cwpe}, which states that $\|\wpe - \wpe^{(0)}\| \leq \theta \zeta^{1/4}\eta t$, continues to hold, we similarly have with probability $1 - d^{-\omega(1)}$,
\begin{align}
    \|\wpe^{(t + 1)} - \wpe^{(t)}\| &= \|\eta \nabla_{\wpe} \hat{L}_{\rho}\| \\
    &\leq \eta \|\nabla_{\wpe} L_0\| + \eta \|\nabla_{w} L_{\rho} - \nabla_{w} \hat{L}_{\rho}\| + \eta\|\nabla_{w} L_0 - \nabla_{w} L_{\rho}\|\\
    &\leq \eta\left(3\theta \zeta^{1/2} + |a_w| \sqrt{\frac{d\log(d)^2}{m}} + 2|a_w |\mathbb{E}_{\rho}[\|a_w w\|]\right) \tagblue{ Lemma~\ref{lemma:pop1a} (item \ref{Awpe}), \ref{lemma:emp}, and \ref{lemma:graddiff1} respectively}\\
    &\leq \eta \theta \zeta^{1/4} \tagblue{\ref{Ca}, $\frac{d}{m} \leq \theta^2$, Claim~\ref{claim:implication}}.
\end{align}

Next we check that item \ref{Copp}, which states that $\|\wopp\|^2 \leq Q_t^2 + \theta B_t^2$, continues to hold. We will use the helper lemma, Lemma~\ref{lemma:helper}. Observe that we have:
\begin{enumerate}
    \item $-\wopp^T\nabla_w L_0 \leq 0$ by Lemma~\ref{lemma:pop1}.
    \item $\|\nabla_{w_{1:2}} L_0 - \nabla_{w_{1:2}} \hat{L}_{\rho}\| \leq \theta B_t$ with probability $1 - d^{-\omega(1)}$. This follows from combining Lemma~\ref{lemma:pop1a} item \ref{Awinf} (applied twice to $i = 1, 2$) with Lemma~\ref{lemma:emp}.
    \item $\|\nabla_{\wopp} L_0\| \leq B_t$ by Lemma~\ref{lemma:pop1a} item \ref{A12}.
\end{enumerate}
Then calling Lemma~\ref{lemma:helper} with $u_t = \wopp$, $G = \nabla_{w_{1:2}} L_0$, $\hat{G} = \nabla_{w_{1:2}} \hat{L}_{\rho}$, and $Q_t$ and $B_t$ as in Definition~\ref{def:control}, we achieve 
\begin{align}
    \|\wopp^{(t+1)}\|^2 \leq (Q_t^2 + \theta B_t^2)(1 + 5\eta \theta^{1/2}) \leq Q_{t+1}^2 + \theta B_{t + 1}^2,
\end{align}
as desired.

Finally we check that \ref{Cwinf}, which states that $\|\wpe\|_{\infty}^2 \leq Q_t^2 + \theta B_t^2$, continues to hold by showing that for any $i \in [3, d]$, $w_i^2$ cannot grow too quickly. 
From Lemma~\ref{lemma:pop1} item \ref{Pwinf}, with $X = (\xi - e_i\xi_i)^Tw$, we have that
\begin{align}
    -w_i^T\nabla_{w_i} L_0 = \frac{|a_w||w_i|}{4}\left(\mathbb{P}_{\xi}\left[X \in [\sqrt{2}\|\wopp\| - |w_i|, \sqrt{2}\|\wopp\| + |w_i|]\right] - \mathbb{P}_{\xi}\left[X \in [\sqrt{2}\|\wsig\| - |w_i|, \sqrt{2}\|\wsig\| + |w_i|]\right]\right).
\end{align}

Now employing Lemma~\ref{lemma:boolean2} twice with $v = \wpe^{(0)}$, $\Delta = \wpe^{(t)} - \wpe^{(0)} - e_iw_i$ both times, and $[a\|v\|, b\|v\|]$ as the intervals $\left[\sqrt{2}\|\wsig\| - |w_i|, \sqrt{2}\|\wsig\| - |w_i|\right]$ and $\left[\sqrt{2}\|\wopp\| - |w_i|, \sqrt{2}\|\wopp\| - |w_i|\right]$, we obtain that 

\begin{align}
    -w_i^T\nabla_{w_i} L_0 \leq |a_w||w_i|P_{\frac{|w_i|}{\|\wpe^{(0)}\|}}\left(\sqrt{2\eta t\zeta^{1/4} + \frac{\log(d)^2}{\sqrt{d}}} + 2|a_w|\frac{\|w_{1:2}\|^2 + \|\wpe\|^2_{\infty}}{\|\wpe^{(0)}\|}\right) + 200\cbe d^{-1/2}|a_w||w_i|,
\end{align}
where $P_c := \mathbb{P}_{G \sim \mathcal{N}(0, 1)}[|G| \leq c]$. Here we have used the fact that $\frac{\|\wpe^{(t)} - \wpe^{(0)} - e_iw_i\|}{\|\wpe^{(0)}\|} \leq 2\eta t\zeta^{1/4} + \frac{\log(d)^2}{\sqrt{d}}$ by \ref{Cwpe} and \ref{Cwinf}, which serves as the role of $\zeta$ in Lemma~\ref{lemma:boolean2}. 

Simplifying this expression, and observing that the second term in the parenthesis becomes insignificant by \ref{Csig}-\ref{Cwinf}, we have that 
\begin{align}
    -w_i^T\nabla_{w_i} L_0 \leq - \frac{|a_w||w_i|^2}{\|\wpe\|}\zeta^{1/9} + 200\cbe d^{-1/2}|a_w||w_i|.
\end{align}
Thus since $|a_w| \leq \|w\| \leq 2\theta$, and $\|\wpe\| \geq \theta/2$ (by \ref{Ca} and \ref{Cwpe}), we have
\begin{align}
    w_i^T\nlrz{w_i} \leq 2|w_i|^2\zeta^{1/9} +  200\cbe d^{-1/2}\theta |w_i| \leq 2\zeta^{1/9}|w_i|^2 + \frac{Q_0}{\log(d)}|w_i|.
\end{align}
Now we proceed via Lemma~\ref{lemma:helper}. Observe that for any $i \in [3, d]$, we have:
\begin{enumerate}
    \item $-w_i^T\nabla_w L_0 \leq  (Q_t^2 + \theta B_t^2)\left(2\zeta^{1/9} + \frac{1}{\log(d)}\right)$ by the calculation above and the assumption that $w$ is controlled.
    \item $\|\nabla_{w_{i}} L_0 - \nabla_{w_{i}} \hat{L}_{\rho}\| \leq \theta B_t$ with probability $1 - d^{-\omega(1)}$. This follows from combining Lemma~\ref{lemma:pop1a} item \ref{Awinf} with Lemma~\ref{lemma:emp}.
    \item $\|\nabla_{w_i} L_0\| \leq B_t/2$ by Lemma~\ref{lemma:pop1a} item \ref{A12}.
\end{enumerate}
Then calling Lemma~\ref{lemma:helper} with $u_t = \wopp$, $G = \nabla_{w_{1:2}} L_0$, $\hat{G} = \nabla_{w_{1:2}} \hat{L}_{\rho}$, and $Q_t$ and $B_t$ as in Definition~\ref{def:control}, we achieve with probability $1 - d^{-\omega(1)}$,
\begin{align}
    \|w_i^{(t+1)}\|^2 \leq (Q_0^2 + \theta B_t^2)\left(1 + 5\eta \left(2\zeta^{1/9} + \frac{1}{\log(d)}\right)\right) \leq Q_{t+1}^2 + \theta B_{t + 1}^2,
\end{align}
as desired.

Finally we check the growth of $\|\wsig\|$, which we will use to show that either \ref{Csig}  or \ref{Wwsig} holds at step $t + 1$, and that a strong neuron stays strong. We have by Lemma~\ref{lemma:pop1a} item \ref{Psig} that
\begin{align}\label{eq:sigcheck}
    \|\wsig^{(t + 1)}\|^2 &= \|\wsig^{(t)} - \eta \mu\mu^T\nabla \hat{L}_{\rho}(w^{(t)})\|^2\\
    &= \|\wsig\|^2 -2\eta \wsig^T\nabla_w \hat{L}_{\rho} + \eta^2\|\mu\mu^T\nabla_w \hat{L}_{\rho}\|^2\\
    &= \|\wsig\|^2 -2\eta \wsig^T\nabla_w L_{0} - 2\eta \wsig^T(\nabla_w \hat{L}_{\rho} - \nabla_w L_0) + \eta^2\|\mu\mu^T\nabla_w \hat{L}_{\rho}\|^2\\
    &=  \|\wsig\|^2 + 2\eta\frac{\sqrt{2}}{4}|a_w|\|\wsig\|X - 2\eta \wsig^T(\nabla_w \hat{L}_{\rho} - \nabla_w L_0) + \eta^2\|\mu\mu^T\nabla_w \hat{L}_{\rho}\|^2,
\end{align}
where $X := \mathbb{E}_{\xi}\mathbf{1}(|w^T\xi| \leq \sqrt{2}\|\wsig\|)$.

Now with probability $1 - d^{-\omega(1)}$, we can control the last two terms as follows. 
\begin{align}\label{eq:C1}
    |- 2\eta \wsig^T(\nabla_w \hat{L}_{\rho} - \nabla_w L_0) + \eta^2\|\mu\mu^T\nabla_w \hat{L}_{\rho}\|^2| &\leq 2\eta \|\wsig\|\|\nabla_{w_{1:2}} \hat{L}_{\rho} - \nabla_{w_{1:2}} L_{0}\| + \eta^2\|\mu\mu^T\nabla_w \hat{L}_{\rho}\|^2\\
    &\leq 2\eta \|\wsig\|\|\nabla_{w_{1:2}}\hat{L}_{\rho}  - \nabla_{w_{1:2}} L_{0}\|\\
    &\qquad+ 2\eta^2\|\nabla_{w_{1:2}} \hat{L}_{\rho} - \nabla_{w_{1:2}} L_{0}\|^2 +  2\eta^2\|\nabla_{w_{1:2}} L_0\|^2.
\end{align}
Now we have
\begin{align}
    \|\nabla_{w_{1:2}}\hat{L}_{\rho}  - \nabla_{w_{1:2}} L_{0}\| &\leq  \|\nabla_{w_{1:2}}\hat{L}_{\rho}  - \nabla_{w_{1:2}} L_{\rho} \| +  \|\nabla_{w_{1:2}}L_{\rho}  - \nabla_{w_{1:2}} L_{0}\|\\
    &\leq \theta B_t/2 + \theta B_t/2 \leq \theta B_t,
\end{align}
by Lemma~\ref{lemma:emp} and Lemma~\ref{lemma:pop1} \ref{Awinf}. Second, we have $\|\nabla_{w_{1:2}} L_0\|^2 \leq B_t^2$ by Lemma~\ref{lemma:pop1} \ref{A12}. Plugging these two bounds back into Eq.~\ref{eq:C1} yields 
\begin{align}\label{eq:errterms}
    |- 2\eta \wsig^T(\nabla_w \hat{L}_{\rho} - \nabla_w L_0) + \eta^2\|\mu\mu^T\nabla_w \hat{L}_{\rho}\|^2| &\leq 2\eta \|\wsig\| \theta B_t + 4\eta^2 B_t^2 \leq 6\eta \theta B_t^2,
\end{align}
where we have used the fact that $\|\wsig\| \leq B_t$ (property \ref{Csig}) and that $\eta \leq \theta$. Now by Lemma~\ref{lemma:boolean2}, we have 
\begin{align}\label{eq:boolapp}
    X &\leq (1 + 3\zeta^{1/10})\mathbb{P}_{G \sim \mathcal{N}(0, 1)}\left[|G| \leq \frac{\sqrt{2}\|\wsig\|}{\|\wpe^{(0)}\|}\right] + 200\cbe d^{-1/2}\\
    &\leq (1 + 3\zeta^{1/10})\frac{2\sqrt{2}}{\sqrt{2\pi}}\frac{\|\wsig\|}{\|\wpe^{(0)}\|} + 200\cbe d^{-1/2} \tagblue{Upper bound Gaussian density by density at $0$.}\\
    &\leq \frac{B_t}{\theta}\frac{2}{\sqrt{\pi}}\left(1 + 3\zeta^{1/10} + \frac{1}{\log^{1.1}(d)}\right) \tagblue{Definition \ref{def:control} of $B_t$ and \ref{Csig}}\\
    &\leq \frac{B_t}{\theta}\frac{2}{\sqrt{\pi}}\left(1 + \frac{2}{\log^{1.1}(d)}\right)  \tagblue{$\zeta = \log^{-c}(d)$ for $c$ large enough.}
\end{align}
Here we have used Lemma~\ref{lemma:boolean2} with $v = \wpe^{(0)}$, $\Delta = \wpe - \wpe^{(0)}$, and $\zeta = \frac{\wpe - \wpe^{(0)}}{\wpe^{(0)}} \leq \zeta^{1/5}$ (by \ref{Cwpe}), and $b = -a = \frac{\|\wsig\|\sqrt{2}}{\|\wpe^{(0)}\|}$. 


To check that either \ref{Csig} in the definition of controlled continues to hold at time $t + 1$, so that $w^{(t + 1)}$ is controlled (Def.~\ref{def:controlled}) \em or \em that item \ref{Wwsig} in the definition of weakly controlled (Def.~\ref{def:weakcontrol}), it suffices to check that $\|\wsig^{(t + 1)}\|^2 \leq B_{t + 1}^2$. Indeed from combining Eqs.~\ref{eq:sigcheck}, \ref{eq:errterms}, \ref{eq:boolapp} we have with probability $1 - d^{-\omega(1)}$,
\begin{align}
    \|\wsig^{(t + 1)}\|^2 &\leq  \|\wsig\|^2 + \frac{1}{\sqrt{2}}\eta \|\wsig\||a_w|X + 6\eta \theta B_t^2\\
    &\leq B_t^2 + 2\eta \tau B_t^2\left(1 + \frac{2}{\log^{1.1}(d)}\right) + 6\eta \theta B_t^2\\
    &\leq B_t^2 \left(1 + 2\eta \tau \left(1 + \frac{1}{\log(d)}\right) \right)\\
    &\leq B_{t + 1}^2.
\end{align}
Here the second inequality we used the fact that $\frac{|a_w|}{\theta} \leq 1 + \Theta(\log(d)\zeta)$ by \ref{Ca}.

Now we check that if the neuron is strong, it stays strong. To do this, we need to lower bound $X = \mathbb{E}_{\xi}\mathbf{1}(|w^T\xi| \leq \sqrt{2}\|\wsig\|)$. For the case that $\frac{\|\wsig\|}{\|\wpe\|}$ is small (ie., $t$ is small), we use Lemma~\ref{lemma:infinity}, which yields that
\begin{align}
    X \geq \frac{1}{2\exp(100\cws^8)\sqrt{d}}.
\end{align}
Indeed, we can apply Lemma~\ref{lemma:infinity} wit $v = \wpe^{(0)}$ and $\Delta = \wpe - \wpe^{(0)}$, which by \ref{Cwpe} yields $\frac{\|\Delta\|}{\|v\|} \leq \zeta^{1/5}$.

Recall the definition of $S_t$ from Definition~\ref{def:strong}. For $t \leq \Ts$, we have $S_t = \frac{\theta}{\sqrt{d}}\left(1 + 2\tau \eta/\cs\right)^{t/2} \leq \frac{\theta}{\sqrt{d}} e^{t \tau \eta /\cs} = \frac{\theta}{\sqrt{d}}800\cbe$, and thus
\begin{align}
    X \geq \frac{1}{2\exp(100\cws^8)}\frac{1}{800\cbe}\frac{S_t}{\theta} = \frac{2}{\cs}\left(\frac{S_t}{\theta}\frac{2}{\sqrt{\pi}}\right),
\end{align}
where we recall that $\cs = \frac{6400}{\sqrt{\pi}}\exp(100\cws^8)$. Further, from Lemma~\ref{lemma:boolean2}, applied as in Eq.~\ref{eq:boolapp}, we have 
\begin{align}
    X &\geq (1 - 3\zeta^{1/10})\mathbb{P}_{G \sim \mathcal{N}(0, 1)}\left[|G| \leq \frac{\sqrt{2}\|\wsig\|}{\|\wpe^{(0)}\|}\right]- \frac{200\cbe}{\sqrt{d}}\\
    &\geq (1 - 3\zeta^{1/10})\left(\frac{2}{\sqrt{\pi}}\frac{\|\wsig\|}{\|\wpe^{(0)}\|}\right)e^{-\frac{2\|\wsig\|^2}{2\|\wpe^{(0)}\|^2}}- \frac{200\cbe}{\sqrt{d}} \tagblue{Lower bound Gaussian density by endpoints of interval}\\
    &\geq (1 - 3\zeta^{1/10})\left(\frac{2}{\sqrt{\pi}}\frac{\|\wsig\|}{\|\wpe^{(0)}\|}\right)\left(1-\frac{\|\wsig\|^2}{\|\wpe^{(0)}\|^2}\right)- \frac{200\cbe}{\sqrt{d}} \tagblue{$e^{-x} \geq 1 - x$}\\
    &\geq (1 - 4\zeta^{1/10})\left(\frac{2}{\sqrt{\pi}}\frac{\|\wsig\|}{\|\wpe^{(0)}\|}\right) - \frac{200\cbe}{\sqrt{d}}. \tagblue{$\frac{\|\wsig\|^2}{\|\wpe^{(0)}\|^2} \leq \zeta^2$ by \ref{Csig}.}
\end{align}
Thus from the definition of strong neurons, since $\eps_s$ is decreasing for $s \leq \Ta$ (see Def.~\ref{def:strong}), we have $\frac{\|\wsig\|^2}{d\|\wpe\|^2} \geq \frac{\|\wsig\|^2}{2d\theta^2} \geq \frac{1}{2}(1 + 2\eta \tau /\cs)^t$, and so 
\begin{align}
    X &\geq \left(1 - 4\zeta^{1/10} - \frac{200\cbe}{\sqrt{d}} \frac{\sqrt{\pi}\|\wpe\|}{2\|\wsig\|}\right)\left(\frac{2}{\sqrt{\pi}}\frac{\|\wsig\|}{\|\wpe\|}\right)\\
    &\geq \left(1 - 4\zeta^{1/10} - \frac{200\cbe \sqrt{\pi}}{(1 + 2\eta \tau/\cs)^{t/2}}\right)\left(\frac{2}{\sqrt{\pi}}\frac{\|\wsig\|}{\|\wpe\|}\right).
\end{align}

Returning to Eq.~\ref{eq:sigcheck}, we have that 
\begin{align}
    \|\wsig^{(t + 1)}\|^2 &\geq \|\wsig\|^2 + 2\eta \frac{\sqrt{2}}{4}|a_w|\|\wsig\|X - 6\eta \theta B_t^2.
\end{align}
To complete this computation, observe that for $t \leq \Ta$, by Lemma~\ref{lemma:St}, we have $B_t^2 \leq S_t^2 \log^3(d) \leq S_t^2\theta^{-1/2}$.
Thus since $\|\wsig\| \geq S_t$, using the previous lower bounds on $X$, we have 
\begin{align}
     \|\wsig^{(t + 1)}\|^2 &\geq S_t^2\left(1 - 6\eta \theta^{1/2}\right) + 2\eta \frac{\sqrt{2}}{4}|a_w|\|\wsig\|X \\
     &\geq \begin{cases}
         S_t^2\left(1 - 6\eta \theta^{1/2} + 2\eta \tau \frac{2}{\cs} \frac{|a_w|}{\theta}\right) &t \leq \Ts;\\
          S_t^2\left(1 - 6\eta \theta^{1/2} + 2\eta \tau \left(1 - 4\zeta^{1/10} - \frac{200\cbe \sqrt{\pi}}{(1 + 2\eta \tau/\cs)^{t/2}}\right) \frac{|a_w|}{\|\wpe\|}\right) & t > \Ts; 
     \end{cases}\\
    &\geq \begin{cases}
         S_t^2\left(1 + \frac{2\eta \tau}{\cs} \right) &t \leq \Ts;\\
          S_t^2\left(1 + 2\eta \tau \left(1 - 5\zeta^{1/10} - \frac{200\cbe \sqrt{\pi}}{(1 + 2\eta \tau/\cs)^{t/2}}\right)\right) & t > \Ts; \\
     \end{cases}\\
     &= S_t^2(1 + 2\eta \tau (1 - \epsilon_t)),
\end{align}
which means that the neuron stays strong.

\end{proof}

We are now ready to prove Lemma~\ref{lemma:inductive1b}.

\begin{proof}[Proof of Lemma~\ref{lemma:inductive1b}]
By Lemma~\ref{lemma:inductive1aALT}, it suffices to prove that with probability $1 - d^{-\omega(1)}$, all weakly controlled neurons stay weakly controlled, and that all weakly controlled and strong neurons stay strong at iteration $t + 1$.

Suppose $(w^{(t)}, a_w^{(t)})$ is weakly controlled.  We begin by showing that with high probability, the five properties \ref{Wwsig}, \ref{Wwopp}, \ref{Wa}, \ref{Wwpe}, \ref{Winf} of controlled neurons hold at iteration $t + 1$. We will omit the proof that \ref{Wwopp} continues to hold, as this proof is nearly identical to the proof for \ref{Copp} in Lemma~\ref{lemma:inductive1aALT}, which proceeds by applying Lemma~\ref{lemma:helper}

We begin with \ref{Wa}. We have with probability $1 - d^{-\omega(1)}$,
\begin{align}
    |a_w^{(t + 1)}|^2 - \|w^{(t + 1)}\|^2 &\geq |a_w|^2 - \|w\|^2 - 4\eta^2 |a_w|^2 \\
    &\geq |a_w|^2 - \|w\|^2 - 4\eta^2\|w\|^2 \\
    &\geq |a_w|^2 - \|w\|^2 - 8\eta^2\theta^2\zeta^{-600},
\end{align}
Here the first and second inequalities are from Lemma~\ref{lemma:layer_balance} item \ref{Sb2} and \ref{Ca} respectively, and the third is from combining \ref{Wwsig}, \ref{Wwopp}, and \ref{Wwpe} to bound $\|w\|$. This, in addition to Lemma!\ref{lemma:layer_balance} \ref{Sb1}, yield the desired conclusion since the gap between $a_w^2$ and $\|w\|^2$ cannot grow by more than $8\eta^2\theta^2\zeta^{-600}$ at each step.

Next we show \ref{Wwpe} holds at step $t + 1$ with high probability. Observe that we have: 
\begin{enumerate}
    \item $-\wpe^T\nabla_w L_0 \leq 0$ by Lemma~\ref{lemma:pop1}, since by the definition of weakly controlled, $\|\wsig\| \geq \|\wopp\|$.
    \item $\|\nabla_{w} L_0 - \nabla_{w} \hat{L}_{\rho}\| \leq \theta \zeta$ with probability $1 - d^{-\omega(1)}$. This follows from combining Lemmas~\ref{lemma:graddiff1} and Claim~\ref{claim:implication} (which together yield $\|\nabla_{w} L_0 - \nabla_{w} L_{\rho}\| \leq \theta\zeta$ ) with Lemma~\ref{lemma:emp} (which yields $\|\nabla_{w} L_{\rho} - \nabla_{w} \hat{L}_{\rho}\| \leq \zeta \theta/2)$.
    \item $\|\nabla_{w} L_{\rho}\| \leq 3B_t$ with probability $1 - d^{-\omega(1)}$ by Lemma~\ref{lemma:layer_balance} (\ref{Sw}) and Lemma~\ref{lemma:graddiff1}.
\end{enumerate}
Note the all the above approximations are very loose. Thus we have 
\begin{align}
    \|\wpe^{(t+1)}\|^2 &\leq \|\wpe^{(t)}\|^2 + 2\eta \|\wpe^{(t)}\|\|\nabla_{w} L_0 - \nabla_{w} \hat{L}_{\rho}\| + \eta^2 \|\nabla_{w} \hat{L}_{\rho}\|^2 \\
    &\leq 2\theta^2(1 + 3\eta \zeta)^t + 2\eta (2\theta)\theta \zeta  + 9\eta^2 B_t^2\\
    &\leq 2\theta^2(1 + 3\eta \zeta)^t\left(1 + 3\eta \zeta\right) \leq 2\theta^2(1 + 3\eta \zeta)^{t+1},
\end{align}
as desired.

We can carry out an almost identical computation to bound $\|\wopp^{(t + 1)}$ to prove that \ref{Wwopp} continues to hold, so we omit the details. 

Next we show that property \ref{Wwsig} holds at step $t + 1$ with high probability. Let $\mu$ be the direction of $\wsig := \wsig^{(t)}$. Following the same steps Lemma~\ref{lemma:inductive1aALT}, we have
\begin{align}\label{eq:bsigcheck}
    \|\wsig^{(t + 1)}\|^2 &= \|\wsig - \eta \mu\mu^T\nabla_w \hat{L}_{\rho}\|^2\\
    &= \|\wsig\|^2 -2\eta \wsig^T\nabla_w \hat{L}_{\rho} + \eta^2\|\mu\mu^T\nabla_w \hat{L}_{\rho}\|^2\\
    &\in \|\wsig\|^2 - 2\eta \wsig^T\nabla_w L_0 \pm 2\eta \|\wsig\|\|\nabla_w \hat{L}_{\rho} - \nabla_w L_{0}\| \pm \eta^2\|\mu\mu^T\nabla_w \hat{L}_{\rho}\|^2\\
    &\in \|\wsig\|^2 - 2\eta \wsig^T\nabla_w L_0 \pm 6\eta \theta B_t^2.
\end{align}
Here the final inequality follows from Eq.~\ref{eq:errterms} from the proof of Lemma~\ref{lemma:inductive1aALT}; the difference in assumption that the neurons are weakly controlled and not controlled does not affect this computation.

Now plugging in the $L_0$ population gradient from Lemma~\ref{lemma:pop1}, we have
\begin{align}\label{eq:sigupdate}
    \|\wsig^{(t + 1)}\|^2 &\in \|\wsig\|^2 - 2\eta \wsig^T\nabla_w L_0 \pm \eta 6\eta \theta B_t^2\\
    &\in \|\wsig^{(t)}\|^2\left(1 + 2\eta \frac{\sqrt{2}}{4}\frac{|a_w|}{\|\wsig\|}\mathbb{P}_{\xi}[|w^T\xi| \leq \sqrt{2}\|\wsig\|]\right)\pm 6\eta \theta B_t^2.
\end{align}

Now to prove that \ref{Wwsig} holds at time $t + 1$, we upper bound $\mathbb{P}_{\xi}[|w^T\xi| \leq \sqrt{2}\|\wsig\|]$ by $1$ and consider two cases:
\begin{enumerate}
    \item Case 1: $\|\wsig^{(t)}\| \leq B_t/2$. In this case, since $|a_w| \leq \|w\| \leq 2B_t$ (since the neuron is weakly controlled), we have $\|\wsig^{(t + 1)}\|^2 \leq B_t^2 \leq B_{t + 1}^2$.
    \item Case 2: $\|\wsig^{(t)}\| \geq B_t/2$. In this case, since $t \geq \Ta$, we have $B_t \geq \theta (1 - o(1))$ (see Definition~\ref{def:T}), and so $\|\wsig\| \geq |a_w|/4$. Thus we have 
    \begin{align}
        \|\wsig^{(t + 1)}\|^2 &\in \|\wsig^{(t)}\|^2\left(1 + 2\eta \frac{\sqrt{2}}{4}4\right)\pm 6\eta \theta B_t^2 \leq \|\wsig^{(t)}\|^2(1 + 4\eta) \leq Q_{t + 1}^2.
    \end{align}
\end{enumerate}

We now show that the if the neurons have the additional strong property at step $t$ (Definition~\ref{def:strong}), they continue to have this property at step $t + 1$. To do this, we need to lower bound the probability $\mathbb{P}_{\xi}[|w^T\xi| \leq \sqrt{2}\|\wsig\|]$.
\begin{claim}\label{claim:strongprob}
If $(w, a_w)$ is weakly controlled and strong, then 
\begin{align}
    \mathbb{P}_{\xi}[|w^T\xi| \leq \sqrt{2}\|\wsig\|] \geq \frac{\|\wsig\|}{5\|\wpe + \wsig\|}.
\end{align}
\end{claim}
\begin{proof}
We will use the Berry-Esseen Inequality (stated in Theorem~\ref{thm:be}). We have
\begin{align}\label{be:app}
    \mathbb{P}_{\xi}[|w^T\xi| \leq \sqrt{2}\|\wsig\|] &\geq \mathbb{P}_{G \sim \mathcal{N}(0, 1)}\left[|G| \leq \frac{\sqrt{2}\|\wsig\|}{\|\wpe\|}\right] - \cbe \frac{\|\wpe\|_3^3}{\|\wpe\|_2^3}\\
    &\geq \mathbb{P}_{G \sim \mathcal{N}(0, 1)}\left[|G| \leq \frac{\sqrt{2}\|\wsig\|}{\|\wpe\|}\right] - \cbe \frac{\|\wpe\|_{\infty}}{\|\wpe\|_2}\\
    &\geq \frac{\|\wsig\|}{4\|\wsig + \wpe\|}- \cbe \frac{\|\wpe\|_{\infty}}{\|\wpe\|_2}.
\end{align}
Now by \ref{Winf}, either we have $\|\wpe\| \leq \|\wsig\|$, or $\frac{\|\wpe\|_{\infty}}{\|\wpe\|_2} \leq \zeta^{1000}$.

We first consider the latter case when $\frac{\|\wpe\|_{\infty}}{\|\wpe\|_2} \leq \zeta^{3}$. By definition of a strong neuron and Lemma~\ref{lemma:St}, we have 
\begin{align}
    \|\wsig\| \geq S_t \geq \theta \zeta^{700}.
\end{align}
Thus we have from Eq.~\ref{be:app}
\begin{align}
    \mathbb{P}_{\xi}[|w^T\xi| \leq \sqrt{2}\|\wsig\|] &\geq \frac{\|\wsig\|}{5\|\wsig + \wpe\|}.
\end{align}
Now if $\|\wpe\| \leq \|\wsig\|$, then we have by Chebychev's inequality,
\begin{align}
     \mathbb{P}_{\xi}[|w^T\xi| \leq \sqrt{2}\|\wsig\|] &\geq 1 -  \mathbb{P}_{\xi}[|w^T\xi| \geq \sqrt{2}\|\wsig\|]\\
     &\geq 1 -  \frac{\mathbb{E}_{\xi}[(\wpe^T\xi)^2]}{2\|\wsig\|^2}\\
     &= 1 - \frac{\|\wpe\|^2}{2\|\wsig\|^2} \geq \frac{1}{2} \geq \frac{\|\wsig\|}{5\|\wsig + \wpe\|}.
\end{align}

\end{proof}
It follows from Claim~\ref{claim:strongprob}, Eq.~\ref{eq:sigupdate}, and the definition of weakly controlled that 
\begin{align}
    \|\wsig^{(t + 1)}\|^2 &\geq  \|\wsig^{(t)}\|^2\left(1 + 2\eta \frac{\sqrt{2}}{4}\frac{|a_w|}{\|\wsig\|}\frac{\|\wsig\|}{5\|\wsig + \wpe\|}\right) -  6\eta \theta B_t^2\\
    &\geq \|\wsig^{(t)}\|^2\left(1 + 2\eta \frac{\sqrt{2}}{4}\frac{\|\wsig + \wpe\|/2}{5\|\wsig + \wpe\|}\right) -  6\eta \theta B_t^2 \tagblue{$|a_w| \geq \|w\|/2$ by \ref{Wa}}\\
    &\geq \|\wsig^{(t)}\|^2\left(1 + \eta \frac{\sqrt{2}}{20}\right) -  \eta \theta^{1/2} \|\wsig^{(t)}\|^2  \tagblue{$\|\wsig\| \geq S_t \geq B_t \zeta^{\Theta(1)}$ by Lemma~\ref{lemma:St}, and $C$ is large enough in terms of $c$.}\\
    &\geq \|\wsig^{(t)}\|^2\left(1 + \frac{\eta}{20}\right),
\end{align}
which means that the neuron stays strong.

We now check that \ref{Winf} continues to hold. If $\|\wpe^{(t)}\| \leq \|\wsig^{(t)}\|$, then it is a routine calculation to verify from the above computation and Lemma~\ref{lemma:pop1} (and the associated [high probability] bounds on the gradients in Lemmas~\ref{lemma:layer_balance}, \ref{lemma:emp} that with probability $1 - d^{-\omega(1)}$, $\|\wpe^{(t)}\|$ grows at a slower rate than $\|\wsig^{(t)}\|$, and thus $\|\wpe^{(t + 1)}\| \leq \|\wsig^{(t + 1)}\|$. We omit the details of the calculation. 

If $\|\wpe^{(t)}\| \geq \|\wsig^{(t)}\|$, then we must consider the growth of $\|\wpe\|_{\infty}$. Fix any $i \geq 3$. $M_t := \chi^{\cbe 10000} \theta (1 + 21\cbe \eta)^{t - \Ta}$ is the bound guaranteed on on $\|\wpe\|_{\infty}$ in Definition~\ref{def:weakcontrol}, item \ref{Winf}.

We will show that $w_i^2$ cannot grow too quickly. Recall that $\xii$ denotes $\xi - e_i \xi_i$. Then symmetrizing over the pair $(z + \xii + e_i\xi_i, z + \xii - e_i\xi_i)$, we have
\begin{align}\label{eq:infgrowth}
    -w_i^T\nlrz{w_i} &= a_w \mathbb{E}_x y(x)\sigma'(w^Tx)\xi_i \\
    &= a_w \frac{1}{2}\mathbb{E}_{x} y(z)\left(\sigma'(w^Tz + w^T\xii + w_i\xi_i) - \sigma'(w^Tz + w^T\xii - w_i\xi_i)\right)w_i\xi_i\\
    &= a_w \frac{1}{2}\mathbb{E}_{x} y(z)\mathbf{1}(|w^Tz + w^T\xii| \leq |w_i|)|w_i| 
\end{align}

Now explicitly evaluating the expectation over $z$, we have
\begin{align}
    a_w\mathbb{E}_{x} &y(z)\mathbf{1}(|w^Tz + w^T\xii| \leq |w_i|) \\
    &= |a_w|\frac{1}{4}\mathbb{E}_{\xii} \left[\mathbf{1}(|\sqrt{2}\|\wsig\| + w^T\xii| \leq |w_i|) + \mathbf{1}(|-\sqrt{2}\|\wsig\| + w^T\xii| \leq |w_i|)\right]\\
    &\qquad - |a_w|\frac{1}{4}\mathbb{E}_{\xii} \left[\mathbf{1}(|\sqrt{2}\|\wopp\| + w^T\xii| \leq |w_i|) + \mathbf{1}(|-\sqrt{2}\|\wopp\| + w^T\xii| \leq |w_i|)\right]\\
\end{align}
Since the neuron is weakly controlled and thus $\|\wsig\| \geq \|\wopp\|$, this equals
\begin{align}
    -|a_w|\frac{1}{4}\mathbb{P}_{\xii}\left[|w^T\xii| \in |w_i| - \sqrt{2}\|\wsig\|, |w_i| - \sqrt{2}\|\wopp\|\right] + |a_w|\frac{1}{4}\mathbb{P}_{\xii}\left[|w^T\xii| \in |w_i| + \sqrt{2}\|\wopp\|, |w_i| + \sqrt{2}\|\wsig\|\right],
\end{align}

By Berry-Esseen (Theorem~\ref{thm:be}), we have 
\begin{align}
-\mathbb{P}_{\xii}&\left[|w^T\xii| \in |w_i| - \sqrt{2}\|\wsig\|, |w_i| - \sqrt{2}\|\wopp\|\right] + \mathbb{P}_{\xii}\left[|w^T\xii| \in |w_i| + \sqrt{2}\|\wopp\|, |w_i| + \sqrt{2}\|\wsig\|\right] \\
&\leq 4\cbe \frac{\|\wpe\|_3^3}{\|\wpe - e_iw_i\|_2^3}\\
&\leq 4\cbe \frac{\|\wpe\|_{\infty}}{\|\wpe - e_iw_i\|_2}\\
&\leq 5\cbe \frac{M_t}{\|\wpe\|_2}.\\
\end{align}
Here the the first inequality following because the Guassian analog of the first probability will be greater than the Gaussian analog of the second probability, since the intervals in question are of the same length, but the first one is closer to $0$. The last inequality follows from \ref{Winf}, since the neuron is weakly controlled.

Now returning to Equation~\ref{eq:infgrowth}, we have
\begin{align}
    -w_i^T\nabla_{w_i} L_0 \leq 5\cbe|a_w|\frac{M_t}{\|\wpe\|}|w_i| \leq 5\cbe|a_w|\frac{M_t^2}{\|\wpe\|}.
\end{align}

Since we are in the case that $\|\wpe\| \geq \|\wsig\|$ (which is also at least $\|\wopp\|$), we have $\|\wpe\| \geq \|w\|/\sqrt{3}$, and thus since $|a_w| \leq \|w\|$ (recall Lemma~\ref{lemma:layer_balance} \ref{Sb1}), we have 
\begin{align}
    -w_i^T\nabla_{w_i} L_{0} \leq 10\cbe M_t^2.
\end{align}

We can show via the same calculation performed in this lemma for $\wsig$, that with probability $1 - d^{-\omega(1)}$, the approximation error due to using the population gradient $\nabla L_0$ instead of $\nabla \hat{L}_{\rho}$ and due to the second order term in $\eta$ are sufficiently small. We omit the details; as before, this uses Lemmas~\ref{lemma:emp} and \ref{lemma:layer_balance}. Thus with probability $1 - d^{-\omega(1)}$, 

\begin{align}
    (w_i^{(t + 1)})^2 \leq M_t^2\left(1 + \eta 21\cbe\right) \leq M_{t + 1}^2,
\end{align}
as desired.

\end{proof}

\section{Phase 2}\label{apx:phase2}
Throughout Phase 2, we will show that we can maintain the following invariant. Recall that we have defined the following notation (summarized in Table~\ref{table:notation}):
\begin{align}
    \bmu &:= f_{\rho}(\mu)y(\mu) \\
    \bmin &:= \min_{\mu \in \{\pm \mu_1, \pm \mu_2\}} \bmu \qquad \gmin := \min_{\mu \in \{\pm \mu_1, \pm \mu_2\}} |\ell'_{\rho}(\mu)| = \frac{\exp(-\bmax)}{1 + \exp(-\bmax)}\\
    \bmax &:= \max_{\mu \in \{\pm \mu_1, \pm \mu_2\}} \bmu \qquad \gmax := \max_{\mu \in \{\pm \mu_1, \pm \mu_2\}} |\ell'_{\rho}(\mu)| = \frac{\exp(-\bmin)}{1 + \exp(-\bmin)}
\end{align}

\heavydef*

We will additionally use/recall the following definitions.
\begin{definition}\label{def:hmargin}
    For a $(\zeta, H)$-signal heavy network $\rho$ with heavy set $S$, we define the \em heavy-margin: \em 
    \begin{align}
        \hmu := \mathbb{E}[\mathbf{1}(w \in S_{\mu})a_w\sigma(w^T\mu)],
    \end{align}
    and $S_{\mu} := S \cap \{w : \wsig^T\mu > 0\}$. Let $\hmin := \min_{\mu \in \muset} \hmu$, and $\hmax :=  \max_{\mu \in \muset} \hmu$.
\end{definition}

Throughout this section, we define the rate parameter $$\tau := \frac{\sqrt{2}}{4}.$$ We additionally define the following quantities.
\begin{definition}[Parameters for Phase 2]\label{def:2param}
    Define $\zeta_{T_1} := \log^{-c/3}(d)$, where $c$ is the constant in Lemma~\ref{lemma:phase1formal}, and $H := -\log(\zeta_{T_1})/20$.
\end{definition}

Our main inductive lemma for this phase is as follows:
\begin{restatable}[Phase 2 Inductive Lemma; Formal]{lemma}{phaseee}\label{lemma:main2}
Suppose $t \leq T_1 + \frac{\zeta_{T_1}^{-1/160}}{\eta}$. If a network $\rho_t$ is $(\zeta, H)$-signal heavy with heavy set $S$ and $\zeta \leq \zeta_{T_1} (1 + 10\eta \zeta H)^{t - T_1}$, then after one minibatch gradient step with step size $\eta \leq \zeta^3$, with probability $1 - d^{-\omega(1)}$,
\begin{enumerate}
    \item $\rho_{t + 1}$ is $(\zeta(1 + 10\eta \zeta H), H)$-signal heavy.
    \item $\hmin^{(t + 1)}  \geq \left(1 + 2\eta \tau (1 - o(1))\gmax\right)\hmin^{(t)}$
    \item $\hmax^{(t + 1)}  \leq \left(1 + 2\eta \tau (1 + o(1))\gmin\right)\hmax^{(t)}$.
\end{enumerate}
Here $\zeta_{T_1}$ and $H$ are defined in Definition~\ref{def:2param}, and $\tau = \frac{\sqrt{2}}{4}$.

\end{restatable}

\begin{lemma}[Base Case from Phase 1]\label{lemma:reduction}
Assume the conclusion of Lemma~\ref{lemma:phase1formal} holds for the network $\rho_{T_1}$ after $T_1$ steps. Then $\rho_{T_1}$ is $(\zeta_{T_1}, H)$ signal-heavy for the parameters $\zeta_{T_1}$ and $H$ define in Definition~\ref{def:2param}. 

Further, we have $\hmin^{(T_1)} \geq \log^{-\Theta(1)}(d)$, and $\hmax^{(T_1)} \leq 1$.

\end{lemma}
\begin{proof}
Let $\rho := \rho_{T_1}$, and we will likewise drop the superscript $T_1$ on all other variables. First observe that $\hmax \leq 1$ since for any $\mu \in \{\pm \mu_1, \pm \mu_2\}$, we have $\hmu \leq \mathbb{E}_{\rho}[\|a_w w_{1:2}\|] \leq 1$. 

For $\mu \in \muset$, define
\begin{align}
    S_{\mu} := \{w : \zeta_{T_1} w^T\mu \geq \exp(6H)\|\wpe\| + \|\wopp\| , a_w y(\mu) > 0\},
\end{align}
and let  $S := S_{\mu_1} \cup S_{-\mu_1} \cup S_{\mu_2} \cup S_{-\mu_2}$. 

First observe that for any $\mu \in \muset$, by the third item of Lemma~\ref{lemma:phase1formal}, we have
\begin{align}
    \hmu &= \mathbb{E}_{\rho}[\mathbf{1}(w \in S_{\mu})\|a_w \wsig\|]\\
    &\geq \frac{1}{20}\log^c(d)\theta^2\\
    &\geq \log^{-\Theta(1)}(d).
\end{align}

fFinally, we have 
\begin{align}
    \mathbb{E}_{\rho}[\mathbf{1}(w \notin S)\|w\|^2] &= \mathbb{E}_{\rho}\left[\mathbf{1}(w \notin S)\left(\|\wns\|^2 + \|\wsig\|^2\right)\right]\\
    &\leq \mathbb{E}_{\rho}\left[\mathbf{1}(w \notin S)\|\wns\|^2\left(1 + (\exp(6H)\zeta_{T_1}^{-1})^2\right)\right]\\
    &\leq \mathbb{E}_{\rho}\left[\|\wns\|^2\right]\left(1 + (\zeta_{T_1}^{-2/3})^2\right)\\
    &\leq 4\theta^2 (\zeta_{T_1}^{-4/3} + 1)\\
    &\leq 8\zeta_{T_1}^{-4/3} (20\log^{-c}\hmin)\\
    &\leq 160 \zeta_{T_1}^{-4/3} \zeta_{T_1}^3 \hmin \\
    &\leq \zeta_{T_1} \hmin.
\end{align}
Finally, observe that the last clause of Lemma~\ref{lemma:phase1formal} yields the final condition of Definition~\ref{def:heavy} abound $\mathbb{E}[\|w\|^2]$. 
This yields the lemma.
\end{proof}
We recall that in Phase 2, our main analysis tool is to compare to the ``clean'' gradients $\ncl$, which are defined in Equation~\ref{eq:cldef} as
\begin{align}
      &\ncl_w L_{\rho} := a_w\mathbb{E}_{x} \ell_{\rho}'(z)\sigma'(w^Tx)x \quad \text{and} \quad \ncl_{a_w} L_{\rho} := \mathbb{E}_{x} \ell_{\rho}'(z)\sigma(w^Tx).
\end{align}

Similarly to Phase 1, the main building blocks of Phase 2 are computations of the gradients $\ncl L_{\rho}$, and bounds on the distance $\|\ncl_w L_{\rho} - \nabla L_{\rho}\|$. We state these main lemmas here. 

Throughout, we define $$H_{\rho} := \mathbb{E}_{\rho}[\|a_w w\|],$$
such that we have
\begin{align}\label{eq:gbd}
   \exp(-H_{\rho}) \leq \gmin \leq \gmax \leq 2.
\end{align}
\subsection{Bounding Distance to Clean Gradients.}

\begin{lemma}\label{claim:approxerror}

For any neuron $w$, we have
\begin{align}
    \|\ncl_w L_{\rho} - \nabla_w L_{\rho}\|_2 &\leq 4|a_w|\zeta H_{\rho};\\
    \|\ncl_{a_w} L_{\rho} - \nabla_{a_w} L_{\rho}\|_2 &\leq 4\|w\| \zeta H_{\rho}.
\end{align}

\end{lemma}
\begin{proof}
Letting $\Delta_x := (\ell'_{\rho}(x) - \ell'_{\rho}(z))\sigma'(w^Tx)$, we have
\begin{align}
    \|\ncl_w L_{\rho} - \nabla_w L_{\rho}\|_2 & = |a_w |\|\mathbb{E}_x \Delta_x x\| \\
    &=|a_w |\sup_{v : \|v\| = 1} \mathbb{E}_x \Delta_x \langle{v, x\rangle} \\
    &\leq |a_w |\sup_{v : \|v\| = 1} \sqrt{\mathbb{E}_x \Delta_x^2} \sqrt{\mathbb{E}_x\langle{v, x\rangle}^2}\\
    &= |a_w |\sqrt{\mathbb{E}_x \Delta_x^2}.
\end{align}
Now 
\begin{align}
    \mathbb{E}_x[\Delta_x^2] &\leq \mathbb{E}_x[(\ell_{\rho}'(x) - \ell'_{\rho}(z))^2]\\
    &= \mathbb{E}_z(\ell'_{\rho}(z))^2\mathbb{E}_{\xi}\left(\frac{\ell_{\rho}'(x)}{\ell_{\rho}'(z)} - 1\right)^2\\
    &\leq \mathbb{E}_z(\ell'_{\rho}(z))^2\mathbb{E}_{\xi}\left(\exp(|f_{\rho}(z) - f_{\rho}(x)|) - 1\right)^2\\
    &\leq \gmax^2 \mathbb{E}_{\xi}\left(\exp(\mathbb{E}_{\rho}|a_w w^T\xi|) - 1\right)^2\\
    &\leq \gmax^2 \mathbb{E}_{\xi}\left(\exp(\mathbb{E}_{\rho}[\|a_w\wpe\|]|v^T\xi|) - 1\right)^2,\\
\end{align}
where $v$ is any unit vector. Here the last line holds because the expression will be maximized when all neurons are in the same direction (up to sign).

Then since $v^T\xi$ is subgaussian, defining $N := \mathbb{E}_{\rho}[\|a_w\wpe\|] \leq \zeta H_{\rho}$, we have 
\begin{align}
    \mathbb{E}_{\xi}\left(\exp(|v^T\xi|) - 1\right)^2 &\leq (\exp(2N + 2N^2) -1)^2 \leq (\exp(\zeta H_{\rho} + \zeta^2 H_{\rho}^2) - 1)^2 \leq 4\zeta^2 H_{\rho}^2. 
\end{align}
(Explicitly, we can verify this by upper bounding the  moments of $|v^T\xi|$ by moments of a Gaussian, and then using the moment generating function of a half-Gaussian distribution.)

Thus plugging this back in and recalling that $\gmax \leq 2$ always, we achieve
\begin{align}
     \|\ncl_w L_{\rho} - \nabla_w L_{\rho}\|_2 \leq 4|a_w|\zeta H_{\rho} < 1,
\end{align}
as desired. Similarly, we have
\begin{align}
    \|\ncl_{a_w} L_{\rho} - \nabla_{a_w} L_{\rho}\|_2 & = \mathbb{E}_x [\|\Delta_x w^Tx\|] \leq \|w\|\sqrt{\mathbb{E}_x \Delta_x^2},
\end{align}
and thus 
\begin{align}
     \|\ncl_{a_w} L_{\rho} - \nabla_{a_w} L_{\rho}\|_2 \leq 4\|w\| \zeta H_{\rho}.
\end{align}
\end{proof}

\subsection{Clean Gradients}

\subsubsection{Neurons in $S$}

\begin{claim}[Clean Gradients in Signal Direction]\label{claim:clean}
If $w \in S$ and $\mu^T\wsig > 0$, then 
\begin{align}
    \mu^T\ncl_w L_{\rho} = -|a_w|\tau \gmu (1 \pm o(1)),
\end{align}
and 
\begin{align}
    -y(\mu)\ncl_{a_w} L_{\rho} = (1 \pm o(1))\|\wsig\|\tau \gmu.
\end{align}
\end{claim}
\begin{proof}
First, we compute 
\begin{align}
     \mu^T\ncl_w L_{\rho} &= a_w \mathbb{E}_z \ell'_{\rho}(z)\sigma'(w^Tz)z^T\mu - \mathbb{E}_x \ell'_{\rho}(z)(\sigma'(w^Tx) - \sigma'(w^Tz))z^t\mu\\
     &\in -\tau a_w y(\mu)\gmu \pm \sqrt{2}\mathbb{E}_x|\ell'_{\rho}(z)|\mathbf{1}(|w^T\xi| \geq |w^Tz|)\\
     &\in -\tau |a_w|\gmu \pm \sqrt{2}|a_w|\gmax \frac{\|\wpe\|^2}{\|\wsig^2\|} \tagblue{Chebychev's inequality} \\
     &\in -\tau |a_w|\gmu \pm \sqrt{2}|a_w|\gmax \zeta^2 \tagblue{$w \in S$}\\
     &= -|a_w|\tau \gmu (1 \pm o(1)).
\end{align}
Second, we compute
\begin{align}
    \ncl_{a_w} L_{\rho} &= \mathbb{E}_z \ell'_{\rho}(z)\sigma(w^Tx)\\
    &= -\tau y(\mu)\left(\gmu\|\wsig\| \pm \Theta(\gmax\|\wns\|)\right)\\
    &= -\tau y(\mu)\left(\gmu\|\wsig\| \pm \Theta(\gmax \zeta \|\wsig\|)\right)\\
    &= -(1 \pm o(1))\frac{\tau y(\mu)}\gmu\|\wsig\|.
\end{align}
\end{proof}

In the following lemma, we lower bound the size of the gradient in the $\wpe$ and $\wopp$ directions.
\begin{lemma}\label{lemma:cleanns}
For a neuron $(w, a_w )$, let $X := \mathbb{P}_{\xi}[|\wpe^T\xi| \geq \sqrt{2}\|\wopp\|]$. Then
    \begin{align}
        \frac{1}{|a_w |}\wpe^T\ncl_w L_{\rho} &\geq \frac{1}{8}\gmin X\|\wpe\| -\zeta\|\wpe\|.
    \end{align}
    and 
    \begin{align}
        \frac{1}{|a_w |}\wopp^T\ncl_w L_{\rho} &\geq \frac{1}{8}\gmin\sqrt{2}\|\wopp\| -\frac{\gmax \sqrt{2}}{4}X\|\wopp\|.
    \end{align}
\end{lemma}
\begin{proof}[Proof of Lemma~\ref{lemma:cleanns}]
We compute 
\begin{align}
    \on{sign}(a_w )&\mathbb{E}_x \ell'_{\rho}(z)\sigma'(w^Tx)x^T\wpe = \mathbb{E}_{\xi}\mathbb{E}_z \ell'_{\rho}(z)\sigma'(w^T\xi + w^Tz)\xi^T\wpe\\
    &= \frac{1}{2}\mathbb{E}_{\xi}\mathbb{E}_z \ell'_{\rho}(z)\left(\sigma'(w^T\xi + w^Tz) - \sigma'(-w^T\xi + w^Tz)\right)\xi^T\wpe\\
    &=\frac{1}{2}\mathbb{E}_z \ell'_{\rho}(z)  \mathbb{E}_{\xi}\mathbf{1}(|w^T\xi| \geq |w^Tz|)|\xi^T\wpe|\\
    &\geq -\frac{1}{4}\gmax\mathbb{E}_{\xi}\mathbf{1}(|w^T\xi| \geq \sqrt{2}\|\wsig\|)|\xi^T\wpe| + \frac{1}{4}\gmin\mathbb{E}_{\xi}\mathbf{1}(|w^T\xi| \geq \sqrt{2}\|\wopp\|)|\xi^T\wpe|\\
    &\geq -\frac{1}{4}\gmax\sqrt{\mathbb{E}_{\xi}[\mathbf{1}(|w^T\xi| \geq \sqrt{2}\|\wsig\|)]}\sqrt{\mathbb{E}_{\xi}[\xi^T\wpe|^2]} + \frac{1}{4}\gmin\mathbb{E}_{\xi}\mathbf{1}(|w^T\xi| \geq \sqrt{2}\|\wopp\|)\mathbb{E}_{\xi}|\xi^T\wpe|\\
    &\geq -\gmax \sqrt{\mathbb{P}[|w^T\xi| \geq \sqrt{2}\|\wsig\|]} \|\wpe\| + \frac{1}{4}\gmin X\|\wpe\|\\
    &\geq -\zeta\|\wpe\| + \frac{1}{4}\gmin X\|\wpe\|.
\end{align}

where here in the last line, we used the inductive hypothesis that $\|\wpe\| \leq \|\wns\| \leq \zeta \|\wsig\|$.
We also compute the gradient in the $\wopp$ direction: 
\begin{align}
     \on{sign}(a_w )\mathbb{E}_x \ell'_{\rho}(z)\sigma'(w^Tx)x^T\wopp &\geq -\frac{1}{4}\mathbb{E}_{\xi} \gmax \mathbf{1}(|w^T\xi| \geq \sqrt{2}\|\wopp\|)\sqrt{2}\|\wopp\| +  \frac{1}{8}\gmin\sqrt{2}\|\wopp\|\\
     &= -\frac{\gmax \sqrt{2}}{4}X\|\wopp\|   + \frac{1}{8}\gmin\sqrt{2}\|\wopp\|.
\end{align}

\end{proof}
We derive the following corollary of Lemma~\ref{lemma:cleanns}, which will be used to show that a weighted average of $\|\wpe\|^2$ and $\|\wopp\|^2$ decreases under the clean gradients. 
\begin{corollary}[Clean Population Gradients for $\wns$]\label{cor:clean}
If $w \in S$, then
\begin{align}
    -\ncl_w L_{\rho}^T\wopp - \exp(6H)\ncl_w L_{\rho}^T\wpe \leq -\zeta^{2/3} \left(\|\wopp\| + \exp(6H)\|\wpe\|\right)|a_w|
\end{align}

\end{corollary}
\begin{proof}
The gist of the proof of this claim is to show that if $\wopp$ is large relative to $\wpe$, then the gradient will be large in the $\wopp$ direction, thereby decreasing $\wopp$. Conversely, if  $\wpe$ is large relative to $\wopp$, then the gradient will be large in the $\wpe$ direction. Because however we are working on the Boolean hypercube, where $\xi$ is not rotationally invariant, the exact condition of ``$\wopp$ being large relative to $\wpe$'' is slightly nuanced. 

Let $$X := \Pr[|\wpe^T\xi| \geq \sqrt{2}\|\wopp\|].$$ If $X$ is large, then we will show that $\wpe^T \ncl_w L_{\rho}$ is sufficiently large to yield the desired result. If $X$ is small, we will show that $\wopp^T \ncl_w L_{\rho}$ is sufficiently large to yield the desired result.

Now 
\begin{align}
    \frac{1}{|a_w|}&\left(\wopp^T\ncl_w L_{\rho} + \exp(6H)\wpe^T\ncl_w  \right)\\
    &\geq \left(\frac{1}{8}\gmin\sqrt{2}\|\wopp\| -\frac{\gmax \sqrt{2}}{4}X\|\wopp\| + \frac{\exp(6H)}{8}\gmin X\|\wpe\| -\zeta \exp(6H)\|\wpe\|\right)\\
    &\geq \frac{1}{8}\gmin\sqrt{2}\|\wopp\| - \frac{\exp(H)\gmin \sqrt{2}}{4}X\|\wopp\| + \frac{\exp(6H)}{8}\gmin X\|\wpe\| -\zeta \gmin \exp(7H)\|\wpe\|
\end{align}

First we will consider the case that $\|\wpe\|$ is large relative to $\|\wopp\|$. We will need the following claim.
\begin{claim}
If $\|\wpe\| \geq 10\|\wopp\|$, then $X \geq \frac{1}{2}$.
\end{claim}
\begin{proof}
If $\|\wpe\|_{\infty} \geq \sqrt{2}\|\wopp\|$, then with $i := \arg \max |(\wpe)_i|$, with probability $1/2$ condtional on $\{\xi_j\}_{j \neq i}$, $\xi_i$ is such that $|\wpe^T\xi| \geq |(\wpe)_i|$, and thus
\begin{align}
    X &= \mathbb{P}[|\wpe^T\xi| \geq \sqrt{2}\|\wopp\|] \geq \frac{1}{2}.
\end{align}
Otherwise, by Berry-Essen (Theorem~\ref{thm:be}), we have
\begin{align}
     \mathbb{P}[|\wpe^T\xi| \geq \sqrt{2}\|\wopp\|] &\geq \mathbb{P}_{G \sim N(0, 1)}\left[|G| \geq \frac{\sqrt{2}\|\wopp\|}{\|\wpe\|}\right] - \frac{\|\wpe\|_{\infty}}{\|\wpe\|}\\
     &\geq \mathbb{P}_{G \sim N(0, 1)}\left[|G| \geq \frac{\sqrt{2}\|\wopp\|}{\|\wpe\|}\right] - \frac{\sqrt{2}\|\wopp\|}{\|\wpe\|}\\
     &\geq  \mathbb{P}_{G \sim N(0, 1)}\left[|G| \geq \frac{\sqrt{2}}{10}\right] - \frac{\sqrt{2}}{10}\\
     &\geq \frac{1}{2}.
\end{align}
\end{proof}
Thus if $\|\wpe\| \geq 10\|\wopp\|$, by bounding $\frac{1}{2} \leq X \leq 1$, we have 
\begin{align}
    \frac{1}{|a_w|}&\left(\wopp^T\ncl_w L_{\rho} + \exp(6H)\wpe^T\ncl_w L_{\rho} \right)\\
    &\geq \left(\frac{1}{8}\gmin\sqrt{2}\|\wopp\| -\frac{\exp(H)\gmin \sqrt{2}}{4}\|\wopp\| + \frac{\exp(6H)}{16}\gmin \|\wpe\| -\zeta \gmin \exp(7H)\|\wpe\|\right)\\
    &\geq \frac{\exp(6H)}{16}\gmin \|\wpe\| - \frac{\exp(H)\gmin \sqrt{2}}{40}\|\wpe\| - \zeta  \gmin \exp(7H)\|\wpe\| \tagblue{$\|\wpe\| \geq 10\|\wopp\|$}\\
    &\geq \frac{\exp(6H)}{20}\gmin \|\wpe\| \tagblue{$\zeta \leq \exp(-10H)$}\\
    &\geq \frac{\gmin}{30}\left(\|\wopp\| + \exp(6H)\|\wpe\|\right) \tagblue{$\|\wpe\| \geq 10\|\wopp\|$}.
\end{align}

Now if $\|\wpe\| \leq 10\|\wopp\|$, we have
\begin{align}
    \frac{1}{|a_w|}&\left(\wopp^T\ncl_w L_{\rho} + \exp(6H)\wpe^T\ncl_w  L_{\rho}\right)\\
    &\geq \left(\frac{1}{8}\gmin\sqrt{2}\|\wopp\| -\frac{\exp(H)\gmin \sqrt{2}}{4}X\|\wopp\| + \frac{\exp(6H)}{8}\gmin X\|\wpe\| -\zeta \exp(6H)\|\wpe\|\right)\\
    &\geq \left(\frac{1}{10}\gmin\sqrt{2}\|\wopp\| -\frac{\exp(H)\gmin \sqrt{2}}{4}X\|\wopp\| + \frac{\exp(6H)}{8}\gmin X\|\wpe\|\right) \tagblue{$\zeta \leq \exp(-10H)$, $\|\wpe\| \leq 10\|\wopp\|$}\\
    &\geq \frac{1}{10}\gmin\sqrt{2}\|\wopp\| \tagblue{$2\sqrt{2}\exp(-5H)\|\wopp\| \leq \|\wpe\|$}\\
    &\geq \gmin \frac{\sqrt{2}}{200}\exp(-6H)\left(\|\wopp\| + \exp(6H)\|\wpe\|\right) \tagblue{$\|\wpe\| \leq 10\|\wopp\|$}\\
    &\geq \zeta^{2/3}\gmin \left(\|\wopp\| + \exp(6H)\|\wpe\|\right).\tagblue{$\zeta \leq \exp(-10H)$}
\end{align}

Here the third inequality follows from the fact that if $\|\wpe\| \geq 2\sqrt{2}\exp(-5H)\|\wopp\|$, then the positive term with an $X$ exceeds the negative term with an $X$. Alternatively, if $\|\wpe\| \leq 2\sqrt{2}\exp(-5H)\|\wopp\|$, then by Chebychev's inequality that $X \leq \frac{\|\wpe\|^2}{2\|\wopp\|^2} \leq 4\exp(-10H)$, so we can bound the negative term with an $X$.

These two cases prove the lemma.

\end{proof}

\subsubsection{Neurons Not in $S$}
Finally, we need to show that the neurons not in $S$ don't grow too large. To do this, we use the following claim, which states that no neuron can grow more at the rate $\tau \gmax$, which is the rate of growth of the neuron in the direction $\text{arg}\min_{\mu} \mu^T$.
\begin{lemma}[Clean Gradient Bound for all Neurons]\label{lemma:cleanall}
For any neuron,
    $$|\ncl_{a_w} L_{\rho}| = |w^T\ncl_w L_{\rho}| \leq \tau \gmax \|w\|.$$
\end{lemma}
\begin{proof}
First recall that 
\begin{align}
    |w^T\ncl_w L_{\rho}| = |\mathbb{E}_z \ell'_{\rho}(z)\mathbb{E}_{\xi}\sigma(w^Tz + w^T\xi)| = |\ncl_{a_w} L_{\rho}|.
\end{align}
We compute
    \begin{align}
&|\mathbb{E}_z \ell'_{\rho}(z)\mathbb{E}_{\xi}\sigma(w^Tz + w^T\xi)|\\
        &\qquad \leq \sup_{\mu} \frac{1}{4}\left(\gmax \mathbb{E}_{\xi}\sigma(w^T\mu + w^T\xi)+ \gmin \mathbb{E}_{\xi}\sigma(-w^T\mu + w^T\xi)\right)\\
        &\qquad\leq \sup_{\mu} \frac{1}{4}\gmax\left( \mathbb{E}_{\xi}\sigma(w^T\mu + w^T\xi) + \sigma(-w^T\mu + w^T\xi)\right)\\
        &\qquad= \sup_{\mu} \frac{1}{8}\gmax\left( \mathbb{E}_{\xi}\sigma(w^T\mu + w^T\xi) + \sigma(-w^T\mu + w^T\xi)+ \sigma(w^T\mu - w^T\xi)+ \sigma(-w^T\mu - w^T\xi)\right)\\
        &\qquad\leq \frac{\sqrt{2}}{4}\|w\|\gmax = \tau \gmax\|w\|.
    \end{align}
    Indeed, the last inequality follows from the fact that the expression is maximized when $w$ is in the direction of $\mu$.  
\end{proof}

We additionally use the following lemma.
\begin{lemma}\label{claim:allneuron}
    For any neuron $w$, with high probability
    \begin{align}
        \|w^{(t + 1)}\|^2 \leq \|w^{(t)}\|^2\left(1 + 2\eta(1 + 2\zeta H)\tau \gmax \right).
    \end{align} 
\end{lemma}

\begin{proof}
With high probability,
    \begin{align}
        \|w^{(t + 1)}\|^2 &= \|w^{(t)}\|^2 - 2\eta w^T\nabla_w \hat{L} + \eta^2 \|\nabla_w \hat{L}\|^2\\
        &\leq \|w^{(t)}\|^2 - 2\eta w^T \ncl_w L_{\rho} + 2\eta \|w\| \|\ncl_w L_{\rho} - \nabla_w \hat{L}\| + \eta^2 \|\nabla_w \hat{L}\|^2\\
        &\leq \|w^{(t)}\|^2 - 2\eta w^T \ncl_w L_{\rho} + 2\eta \|w\|^2 \zeta H\gmax + \eta^2 \|\nabla_w \hat{L}\|^2\\
        &\leq \|w^{(t)}\|^2 - 2\eta w^T \ncl_w L_{\rho} + 4 \eta H \zeta \gmax \|w\|^2,
    \end{align}
    where in the second inequality we used Claim~\ref{claim:approxerror}. Plugging in Lemma~\ref{lemma:cleanall} yields the claim.
\end{proof}

\subsection{Proof of Inductive Lemma}

We break the proof of Lemma~\ref{lemma:main2} up into three main lemmas. The first lemma shows the growth of $\hmin$. The second ensures that $\hmax$ doesn't grow too fast. The third lemma ensures that the network stays signal-heavy.

\begin{lemma}[$\hmin$]\label{lemma:heavygrowth}
Suppose $\rho_t$ is $(\zeta, H)$-signal heavy for some signal-heavy set $S$. Then if $\eta \leq \zeta^3$, with probability $1 - d^{-\omega(1)}$,
\begin{align}
    \hmin^{(t + 1)}  \geq \left(1 + 2\eta \tau(1 - o(1))\gmax\right)\hmin^{(t)}.
\end{align}
Further, for any neuron for which $\|\wsig^{(t)}\| \leq \exp(6H)\|\wpe^{(t)}\| + \|\wopp^{(t)}\|$, we have $\|\wsig^{(t + 1)}\| \leq \exp(6H)\|\wpe^{(t + 1)}\| + \|\wopp^{(t + 1)}\|$.
\end{lemma}

\begin{lemma}[$\hmax$]\label{lemma:bmax}
Suppose $\rho_t$ is $(\zeta, H)$-signal heavy for some signal-heavy set $S$. Then if $\eta \leq \zeta^3$, with probability $1 - d^{-\omega(1)}$,
\begin{align}
    \hmax^{(t + 1)}  \leq \left(1 + 2\eta \tau(1 + o(1))\gmin\right)\hmax^{(t)}.
\end{align}
\end{lemma}

To prove these two lemmas, we will also need the following lemma which states the the network doesn't change too much at each iteration.
\begin{lemma}\label{lemma:small_step}
If $\rho_t$ satisfies Definition~\ref{def:heavy} and $\eta \leq \zeta^2$, then with probability $1 - d^{-\omega(1)}$, we have
\begin{align}
    |\hmu^{(t + 1)} - \hmu^{(t)}| &\leq \sqrt{\eta}.
\end{align}  
Further for any $\mu$,
\begin{align}\label{hgap}
    |\bmu - \hmu| \leq 2\zeta H.
\end{align}
\end{lemma}
\begin{proof}
Using Lemma~\ref{lemma:layer_balance}, and \ref{lemma:emp} 
we have with probability $1 - d^{-\omega(1)}$
\begin{align}
    |\hmu^{(t + 1)} - \hmu^{(t)}| &\leq \eta \mathbb{E}_{\rho}[\|\nabla_{a_w} \hat{L}_{\rho}\|\|w\| + \|\nabla_{w}\hat{L}_{\rho}\|\|a_w\|]\\
    &\leq 2\eta \mathbb{E}_{\rho}[\|w\|^2 + \|a_w\|^2] \tagblue{Lemma~\ref{lemma:layer_balance} and \ref{lemma:emp} }\\
    &\leq 8\eta H \tagblue{Definition~\ref{def:heavy}}\\
    &\leq \sqrt{\eta} \tagblue{$H \leq \log(\zeta^{-1})/10 \leq \log(\eta^{-1/2})/10 \leq \eta^{-1/2}/8$}.
\end{align}
For the second statement, we have
\begin{align}
    |\bmu - \hmu| &\leq \mathbb{E}_{\rho}[\mathbf{1}(w \notin S)\|a_w w\|] + \mathbb{E}_{\rho}[\mathbf{1}(w \in S)\|a_w \wopp\|]\\
    &\leq \mathbb{E}_{\rho}[\mathbf{1}(w \notin S)\|w\|^2] + \zeta \mathbb{E}_{\rho}[\mathbf{1}(w \in S)\|a_w w\|]\\
    &\leq 2\zeta H.
\end{align}
\end{proof}

\begin{proof}[Proof of Lemma~\ref{lemma:heavygrowth}]
Our approach here will be to show that for any $\mu \in \{\pm \mu_1, \pm \mu_2\}$ from which $\hmu \leq \hmin + 2\sqrt{\eta}$, we have 
\begin{align}\label{eq:smallBincrease}
    \hmu^{(t + 1)} \geq \left(1 + \eta \frac{\sqrt{2}}{2}(1 + o(1))\gmax\right)\hmin^{(t)}.
\end{align}
Then by Lemma~\ref{lemma:small_step} for any $\mu$ for which $\hmu^{(t)} \geq \hmin^{(t)} + 2\sqrt{\eta}$, we have 
\begin{align}
    \hmu^{(t + 1)}  \geq \hmu^{(t)} - \sqrt{\eta} \geq \hmin^{(t)} + \sqrt{\eta} \geq \left(1 + \eta\right)\hmin^{(t)} \geq \left(1 + \frac{\sqrt{2}}{2}\eta \gmax^{(t)}\right)\hmin^{(t)}.
\end{align}

Let us prove Equation~\ref{eq:smallBincrease}. Fix any $\mu$ with $\hmu \leq \hmin + 2\sqrt{\eta}$. We first define a set of neurons on which the growth of signal is large. Let
\begin{align}
    S_{\mu} = \{w : \zeta w^T\mu \geq \|\wpe + \wopp\| , a_w y(\mu) > 0\},
\end{align}
that is, $S_{\mu} = S \cap \{w : \wsig^T\mu > 0\}$, where $S$ is the signal-heavy set from Definition~\ref{def:heavy}.

\begin{claim}\label{claim:aw}
For any $w \in S_{\mu}$,
\begin{align}
    \|a_w^{(t+1)}\wsig^{(t + 1)}\| \geq \|a_w^{(t)}\wsig^{(t)}\|\left(1 + 2\eta \tau (1 - o(1))\gmu^{(t)}\right)
\end{align}
\end{claim}
\begin{proof}[Proof of Claim~\ref{claim:aw}]
Observe that (with $y = y(\mu)$ and $X := \tau \gmu^{(t)}$, we have
\begin{align}
     y a^{(t+1)}_w & \|\wsig^{(t + 1)}\| =  y(a^{(t)}_w - \eta \nabla_{a_w} \hat{L}_{\rho})(\|\wsig^{(t)}\| - \eta \mu^T\nabla_w \hat{L}_{\rho})\\
     &\geq y(a_w - \eta \ncl_{a_w} L_{\rho})(\|\wsig\| - \eta \mu^T\ncl_w L_{\rho}) - O\left(\eta |a_w|\|\ncl_w L_{\rho} - \nabla_w \hat{L}_{\rho}\| + \eta \|\wsig|\|\ncl_{a_w} L_{\rho} - \nabla_{a_w} \hat{L}_{\rho}\|\right) - O(\eta^2\|a_w w\|) \tagblue{Lemma~\ref{lemma:layer_balance}}\\
     &\geq y(a_w - \eta \ncl_{a_w} L_{\rho})(\|\wsig\| - \eta \mu^T\ncl_w L_{\rho}) - O\left(\eta \zeta H_{\rho} (a_w^2 + \|\wsig\|^2)\right) - O(\eta^2\|a_w w\|) \tagblue{Lemma~\ref{claim:approxerror}}\\
     &\geq y a_w \|\wsig\|  + \eta (1 - o(1))\left(X\|\wsig\|^2 + X a_w^2 \right) - O\left(\eta \zeta H_{\rho}(a_w^2 + \|w\|^2)\right) - O(\eta^2 \|a_w w\|) \tagblue{Lemma~\ref{claim:clean}}\\
     &\geq y a_w \|\wsig\| + 2X\eta (1 - o(1))y a_w \|\wsig\| + O(\eta^2 \|a_w w\|) \tagblue{AM-GM, and $\zeta H_{\rho} = o(\gmin)$ by Eq.~\ref{eq:gbd}, $\zeta \leq \exp(-10H)$}\\
     & \geq  y a_w^{(t)}\|\wsig^{(t)}\|\left(1 + 2\eta(1 - o(1))X\right),
\end{align}
as desired.
\end{proof}

Now we have
\begin{align}
    \frac{\gmu}{\gmax} &= \frac{\exp(-\bmu + \bmin)(1 + \exp(-\bmin))}{1 + \exp(-\bmu)}\\
    &\geq \exp(-\bmu + \bmin) \tagblue{$\bmu \geq \bmin$}\\
    &\geq \exp(-\hmu + \hmin - 4\zeta H) \tagblue{Eq~\ref{hgap}}\\
    &\geq \exp(- 2\sqrt{\eta} - 4\zeta H) \tagblue{Lemma~\ref{lemma:small_step}}\\
    &\geq 1 - o(1).
\end{align}
Plugging this in to the previous equation yields
\begin{align}
    y a^{(t+1)}_w \|\wsig^{(t + 1)}\| \geq y a_w^{(t)}\|\wsig^{(t)}\|\left(1 + 2\tau\eta(1 - o(1))\gmax\right)
\end{align}

Now it remains to check that if a neuron is in $S_{\mu}$ at step $t$, then that neuron still satisfies $\zeta \|\wsig^{(t + 1)}\| \geq \exp(6H)\|\wpe^{(t + 1)}\| + \|\wopp^{(t + 1)}\|$ at time $t + 1$. Observe that for every neuron in $S$, we have:
\begin{enumerate}
    \item $\|\wsig^{(t + 1)}\| \geq \|\wsig^{(t)}\|$. This is easy to show (as in the calculation above) by plugging in the lower bound on $-\wsig^T\ncl_w L_{\rho}$, and the upper bound on $\|\ncl_w L_{\rho} - \nabla \hat{L}_{\rho}\|$.
    \item Since $\eta$ is small enough (relative to $\zeta$), if $\exp(6H)\|\wpe^{(t)}\| + \|\wopp^{(t)} \| \leq \zeta \|\wsig^{(t)}\|/2$, then $\exp(6H)\|\wpe^{(t + 1)}\| + \|\wopp^{(t + 1)} \| \leq \zeta \|\wsig^{(t)}\| \leq \zeta \|\wsig^{(t + 1)}\|$.
    \item If $\exp(6H)\|\wpe^{(t)}\| + \|\wopp^{(t)} \| \geq \zeta \|\wsig^{(t)}\|/2$, then 
    \begin{align}
        &\exp(6H)\|\wpe^{(t + 1)}\|^2 + \|\wopp^{(t + 1)}\|^2 - \left(\exp(6H)\|\wpe^{(t)}\|^2 + \|\wopp^{(t)}\|^2\right)\\
        &\quad\leq - 2\eta\exp(6H)(\wpe^{(t)})^T\nabla_w \hat{L}_{\rho} - 2\eta(\wopp^{(t)})^T\nabla_w \hat{L}_{\rho} + \exp(6H)\eta^2\|\nabla_w \hat{L}_{\rho}\|^2\\
        &\quad\leq -2\eta \left(\exp(6H)\wpe^T\ncl_w L_{\rho} + 2\eta\wopp^T\ncl_w L_{\rho}\right) + 2\eta (\exp(6H)\|\wpe\| + \|\wopp\|)\|\ncl_w L_{\rho} -  \nabla_w \hat{L}_{\rho}\| + \exp(6H)\eta^2\|\nabla_w \hat{L}_{\rho}\|^2\\
    \end{align}
    Now we can use Corollary~\ref{cor:clean} to bound the first term, and Lemma~\ref{claim:approxerror}, Lemma~\ref{lemma:emp}, and Lemma~\ref{lemma:layer_balance} to bound the second and third terms. Thus we have 
    \begin{align}
        &\exp(6H)\|\wpe^{(t + 1)}\|^2 + \|\wopp^{(t + 1)}\|^2 - \left(\exp(6H)\|\wpe^{(t)}\|^2 + \|\wopp^{(t)}\|^2\right)\\
        &\quad \leq -2\eta\zeta^{2/3} \left(\|\wopp\| + \exp(6H)\|\wpe\|\right)|a_w| + 2\eta (\exp(6H)\|\wpe\| + \|\wopp\|)\zeta H_{\rho} |a_w| + O(\exp(6H)\eta^2\|a_w\|^2)\\
        &\quad \leq -\eta\zeta^{2/3} \left(\|\wopp\| + \exp(6H)\|\wpe\|\right)|a_w| + O(\exp(6H)\eta^2\|a_w\|^2) \tagblue{$\zeta \leq \exp(-10H)$}\\
        &\quad \leq -\eta\zeta^{2/3} \left(\|\wopp\| + \exp(6H)\|\wpe\|\right)|a_w| + O(\exp(6H)\eta^2\|w\||a_w|) \tagblue{$|a_w| \leq \|w\|$}\\
        &\quad \leq -\eta\zeta^{2/3}(\zeta/4)\|w\||a_w| + O(\exp(6H)\eta^2\|w\||a_w|) \tagblue{$\zeta \|\wsig\|/2 \leq \exp(6H)\|\wpe\| + \|\wopp\| \leq \zeta \|\wsig\|$} \\
        &\quad \leq 0. \tagblue{$\eta \leq \zeta^3$.}
    \end{align}
    Thus it follows that $\exp(6H)\|\wpe^{(t + 1)}\| + \|\wopp^{(t + 1)}\| \leq \exp(6H)\|\wpe^{(t)}\| + \|\wopp^{(t)}\| \leq \|\wsig^{(t)}\| \leq \zeta \|\wsig^{(t + 1)}\|$.
\end{enumerate}

\end{proof}

\begin{proof}[Proof of Lemma~\ref{lemma:bmax}]
Our approach here is similar to the previous lemma. We will show that for any $\mu \in \{\pm \mu_1, \pm \mu_2\}$ from which $\hmu \geq \hmax -2\sqrt{\eta}$, we have 
\begin{align}\label{eq:bigBincrease}
    \hmu^{(t + 1)} \leq \left(1 + \eta \frac{\sqrt{2}}{2}(1 + o(1))\gmin\right)\hmax^{(t)}.
\end{align}
Then by Lemma~\ref{lemma:small_step}, for any $\mu$ for which $\hmu^{(t)} \leq \hmax^{(t)} - 2\sqrt{\eta}$, we have 
\begin{align}
    \hmu^{(t + 1)}  \leq \hmu^{(t)} + \sqrt{\eta} \leq \hmax^{(t)} - \sqrt{\eta} \leq \hmax^{(t)}.
\end{align}

For any neurons $w \in S_{\mu}$, using Lemma~\ref{claim:clean} to bound $\ncl_w L_{\rho}$, Lemma~\ref{claim:approxerror} to bound $\|\ncl_w L_{\rho} - \nabla_w L_{\rho}\|$, and Lemma~\ref{lemma:emp} to bound $\|\nabla_w \hat{L}_{\rho} - \nabla_w L_{\rho}\|$, we have with probability $1 - d^{-\omega(1)}$, 
\begin{align}
    \|\wsig^{(t + 1)}\|^2 & \leq \|\wsig\|^2\left(1 + 2\eta\frac{|a_w|}{\|\wsig\|}\tau \gmu (1 + o(1))\right).
\end{align}

Similarly, by the same 3 lemmas, we have with probability $1 - d^{-\omega(1)}$,
\begin{align}
    (a_w^{(t + 1)})^2 = a_w^2\left(1 + 2\eta\frac{\|\wsig\|}{|a_w|}\tau \gmu (1 + o(1))\right).
\end{align}
Thus
\begin{align}\label{eq:ins}
    \mathbb{E}_{\rho_{t + 1}}&\mathbf{1}(w^{(t)} \in S_{\mu})\|a_w^{(t + 1)}\wsig^{(t + 1)}\| \\
    &\leq \mathbb{E}_{\rho}\mathbf{1}(w^{(t)} \in S_{\mu})\|a_w\wsig\| \left(1 + \eta\frac{|a_w|}{\|\wsig\|}\tau \gmu (1 + o(1))\right)\left(1 + \eta\frac{\|\wsig\|}{|a_w|}\tau \gmu (1 + o(1))\right)\\
    &\leq \mathbb{E}_{\rho}\mathbf{1}(w \in S_{\mu})\|a_w\wsig\| + \eta \tau \gmu (1 + o(1))\mathbb{E}_{\rho}\mathbf{1}(w \in S_{\mu}){\mu})\left(a_w^2 + \|\wsig\|^2\right)\\
    &\leq \mathbb{E}_{\rho}\mathbf{1}(w \in S_{\mu})\|a_w\wsig\| + \eta \tau \gmu (1 + o(1))\left(\mathbb{E}_{\rho}[\mathbf{1}(w \in S_{\mu})2a_w^2] + O(\log(d)\eta H)\right)\\
    &\leq \mathbb{E}_{\rho}\mathbf{1}(w \in S_{\mu})\|a_w\wsig\|\left(1 + 2\eta \tau \gmu (1 + o(1))\right) \tagblue{$|a_w| \leq \|w\| \leq (1 + o(1))\|\wsig\|$ since $w \in S$.}\\
\end{align}
Next we show that $\frac{\gmu}{\gmin}$ is small. We have
\begin{align}
    \frac{\gmu}{\gmin} &= \frac{\exp(-\bmu + \bmax)(1 + \exp(-\bmax))}{1 + \exp(-\bmu)}\\
    &\leq \exp(-\bmu + \bmax) \tagblue{$\bmu \leq \bmax$}\\
    &\leq \exp(-\hmu + \hmax - 4\zeta H) \tagblue{Eq.\ref{hgap}}\\
    &\leq \exp(2\sqrt{\eta} + 4\zeta H) \tagblue{Assumption that $\hmu \geq \hmax - 2\sqrt{\eta}$}\\
    &\leq 1 + o(1).
\end{align}
Plugging this into Eq.~\ref{eq:ins} yields
\begin{align}
    \hmu^{(t + 1)} \leq \hmu^{(t)}\left(1 + 2\eta \tau \gmin (1 + o(1))\right),
\end{align}
as desired.

\end{proof}

\begin{proof}[Proof of Lemma~\ref{lemma:main2}]
The second and third items of Lemma~\ref{lemma:main2} follow immediately from Lemmas~\ref{lemma:heavygrowth} and \ref{lemma:bmax}. To prove the first item, we first need to control the growth of $\mathbb{E}_{\rho}[\mathbf{1}(w \notin S)\|w\|^2]$. We compute 
\begin{align}
    \mathbb{E}_{\rho_{t + 1}}[\|w\|^2\mathbf{1}(w \notin S)] &\leq \mathbb{E}_{\rho_{t}}[\|w\|^2\mathbf{1}(w \notin S)]\left(1 + 2\eta(1 + 2\zeta H)\tau \gmax\right)\\
    &\leq \zeta \hmin^{(t)}\left(1 + 2\eta(1 + 2\zeta H)\tau \gmax \right) \tagblue{Lemma~\ref{claim:allneuron}}\\
    &\leq \zeta \hmin^{(t + 1)} \frac{1 + 2\eta(1 + 2\zeta H)\tau \gmax}{1 + 2\eta(1 - 2\zeta H)\tau \gmax} \tagblue{Lemma~\ref{lemma:heavygrowth}}\\
    &\leq \zeta \hmin^{(t + 1)}\left(1 + 10 \eta \zeta H\right).
\end{align}

Letting $\zeta' := \zeta\left(1 + 10 \eta \zeta H\right)$ be the new signal-heavy parameter, it follows that 
\begin{align}
    \zeta' &\leq \zeta\left(1 + 10 \eta \zeta H\right)\\
    &\leq \zeta_{T_1}\left(1 + 10 \eta \zeta H\right)^{t - T_1}\\
    &\leq \zeta_{T_1}\left(1 + 10 \eta \zeta  H\right)^{\frac{\zeta_{T_1}^{-1/2}}{\eta}} \tagblue{$t - T_1 \leq \frac{\zeta_{T_1}^{-1/2}}{\eta}$ by assumption}\\
    &\leq \zeta_{T_1}e^{10\zeta H \zeta_{T_1}^{-1/2}}\\
    &\leq \zeta_{T_1}e \\
    &\leq \exp(-10H) \tagblue{Lemma~\ref{lemma:reduction}}. 
\end{align}

Further, by Lemma~\ref{lemma:layer_balance}, we have that with probability $1 - d^{-\omega(1)}$, for all neurons, $|a_w^{(t + 1)}| \leq \|w^{(t + 1)}\|$, and 
\begin{align}
    \mathbb{E}_{\rho_{t + 1}}[\|w^{(t + 1)}\|^2] - \mathbb{E}_{\rho_{t + 1}}[(a_w^{(t + 1)})^2] &\leq \mathbb{E}_{\rho_{t}}[\|w^{(t)}\|^2] - \mathbb{E}_{\rho_{t}}[(a_w^{(t)})^2] + 4\eta^2 \mathbb{E}_{\rho_t}(a_w^{(t)})^2\\
    &\leq 2\zeta H + 2\eta^2 H\\
    &\leq 2\zeta' H.
\end{align}

Finally, to bound $\mathbb{E}_{\rho_{t + 1}}[(a_w^{(t + 1)})^2]$, we have 
\begin{align}\label{abd}
    \mathbb{E}_{\rho_{t + 1}}[(a_w^{(t + 1)})^2] &\leq \mathbb{E}_{\rho_{t + 1}}[\|a_w^{(t + 1)}w^{(t + 1)}\|]\\
    &\leq \mathbb{E}_{\rho_{t + 1}}[\mathbf{1}(w^{(t + 1} \in S)\|a_w^{(t + 1)}w^{(t + 1)}\|] + \mathbb{E}_{\rho_{t + 1}}[\mathbf{1}(w^{(t + 1} \notin S)\|a_w^{(t + 1)}w^{(t + 1)}\|]\\
    &\leq \sum_{\mu \in \muset}\hmu^{(t + 1)} + \zeta' \hmin^{(t + 1)}\\
    &\leq 5\hmax^{(t + 1)}.
\end{align}

We bound $\hmax^{(t + 1)}$ in the following claim.
\begin{claim}
    $$\hmax^{(T_1 + \zeta_{T_1}^{-1/2}/\eta)} \leq -\log(\zeta_{T_1})/20.$$
\end{claim}
\begin{proof}
    Let $t^*$ be the last time at which $\hmax$ is at most $-\log(\zeta_{T_1})/40$. If $t^* \geq T_1 + \zeta_{T_1}^{-1/160}/\eta$, then we are done. Suppose $t^* \leq T_1 + \zeta_{T_1}^{-1/160}/\eta$. Then by Lemma~\ref{lemma:bmax}, we have
    \begin{align}
        \hmax^{(T_1 + \zeta_{T_1}^{-1/160}/\eta)} &\leq \hmax^{(t^*)}\left(1 + 2\eta (1 + o(1))\tau \max_{t^* \leq s \leq T_1 + \zeta_{T_1}^{-1/160}/\eta} \gmin^{(s)} \right)^{\zeta_{T_1}^{-1/160}/\eta}\\
        &\leq \frac{-\log(\zeta_{T_1})}{40}\left(1 + 2\eta (1 + o(1))\tau \max_{t^* \leq s \leq T_1 + \zeta_{T_1}^{-1/160}/\eta} \gmin^{(s)} \right)^{\zeta_{T_1}^{-1/2}/\eta}\\
        &\leq \frac{-\log(\zeta_{T_1})}{40}\exp\left(1.5\zeta_{T_1}^{-1/160} \max_{t^* \leq s \leq T_1 + \zeta_{T_1}^{-1/160}/\eta} \gmin^{(s)} \right)\\
        &\leq \frac{-\log(\zeta_{T_1})}{40}\exp\left(2\zeta_{T_1}^{-1/160} \exp(-\hmax^{(t^*)}) \right)\\
        &\leq \frac{-\log(\zeta_{T_1})}{40}\exp\left(2\zeta_{T_1}^{-1/160} \zeta_{T_1}^{\frac{1}{40}}) \right)\\
        &\leq  \frac{-\log(\zeta_{T_1})}{20}.
    \end{align}

Here we have used Lemma~\ref{lemma:small_step} to bound
\begin{align}
    \gmin^{(s)} \leq 2\exp(-\bmax^{(s)}) \leq 2\exp(-\hmax^{(s)} + 2\zeta H) , 
\end{align}
and thus
\begin{align}
    \max_{t^* \leq s \leq T_1 + \zeta_{T_1}^{-1/160}/\eta} \gmin^{(s)} &\leq 2.1 \exp(-\hmax^{(t^*)}).
\end{align}
\end{proof}
Plugging this claim into Eq.~\ref{abd} yields
\begin{align}
    \mathbb{E}_{\rho_{t + 1}}[(a_w^{(t + 1)})^2] \leq 5\hmax^{(t + 1)} \leq -\log(\zeta_{T_1})/20 \leq 2H.
\end{align}
The above computations, in addition to Lemma~\ref{lemma:heavygrowth} proves that $\rho_{t + 1}$ is $(\zeta', H)$ signal-heavy with the heavy set $S$.
\end{proof}

\section{Proof of Theorem~\ref{thm:main}}\label{apx:thmpf}
We now prove Theorem~\ref{thm:main} from Lemmas~\ref{lemma:phase1formal} and Lemmas~\ref{lemma:main2}. We restate the theorem and these two lemmas for the readers convenience. 

\begin{theorem}
There exists a constant $C > 0$ such that the following holds for any $d$ large enough. Let $\theta := 1/\log(d)^C$. Suppose we train a 2-layer neural network with minibatch SGD as in Section~\ref{sec:training} with a minibatch size of $m \geq d/\theta$, width $1/\theta \leq p \leq d^C$, step size $\eta \leq \theta$, and initialization scale $\theta$. Then for some $t \leq C\log(d)/\eta$, with probability $1 - d^{-\omega(1)}$, we have
\begin{align}
    \mathbb{E}_{x \sim P_d}[\ell_{\rho_t}(x)] \leq o(1).
\end{align}
\end{theorem}

\lemphase*

\phaseee*

\begin{proof}[Proof of Theorem~\ref{thm:main}]
Let $\rho_{T_1}$ be the network output by Lemma~\ref{lemma:phase1formal}. By Lemma~\ref{lemma:reduction}, we have that $\rho_{T_1}$ is $(\zeta_{T_1}, H)$-signal-heavy, where $\zeta_{T_1}$ and $H$ are defined in Definition~\ref{def:2param}. Further, we have that $\hmin^{(T_1)} \geq \zeta_{T_1}^{1/200}$.


Let us iterate Lemma~\ref{lemma:main2} for $T_2 := -\zeta_{T_1}^{-1/160}/\eta$ steps. We will show that $\hmin^{(T_1 + T_2)} = \omega(1)$.

If $\hmin^{(t)}$ for $t \in [T_1, T_1 + T_2]$ ever exceeds $\log \log \log (\zeta_{T_1}^{-1})$, then we are done since $\hmin$ always increases. Suppose it does not exceed this value. Then we can show that $\gmin$ is relatively large, and thus by the second item of Lemma~\ref{lemma:main2}, $\hmin$ will grow quickly. Indeed for $d$ large enough:
\begin{align}
    \hmin^{(T_1 + T_2)} &\geq \hmin^{(T_1)} \left(1 + 2\eta \tau (1 - o(1))\min_{t \in [T_1, T2] } \gmin^{(t)}\right)\\
    &\geq \hmin^{(T_1)} \left(1 + \eta \exp(-\log \log \log (\zeta_{T_1}^{-1}))\right)^{T_2}\\
    &\geq \hmin^{(T_1)} \exp\left(\zeta_{T_1}^{-1/160}\log \log (\zeta_{T_1}^{-1})^{-1}\right)\\
    &\geq \zeta_{T_1}^{\Theta(1)} \exp\left(\zeta_{T_1}^{-1/161}\right) \tagblue{Lemma~\ref{lemma:reduction}}\\
    & = \omega(1) \tagblue{$\zeta_{T_1} = \log^{-\Theta(1)}(d)$}.
\end{align}
Here in the second inequality we have lower bounded $\gmin^{(t)}$ by $(1 - o(1))\exp(-\hmin^{(t)})$ using Lemma~\ref{lemma:small_step}.

Finally, we check the loss guarantee of the network $\rho_{T}$ for $T = T_1 + T_2$. Since $\rho_T$ is $(\zeta', H)$-signal heavy for $\zeta' \leq 2\zeta_{T_1}$ (see the proof of Lemma~\ref{lemma:main2}), and by Definition~\ref{def:2param}, $\zeta_{T_1}H = o(1)$, we have

\begin{align}
     \mathbb{E}_x \ell_{\rho_T}(x) &= -2\log\left(\frac{1}{1 + \exp(-f_{\rho}(x)y(x))}\right) \\
     &\leq \mathbb{E}_x 2\exp(-f_{\rho}(x)y(x)) \\
     &\leq \mathbb{E}_x 2\exp(-\bmin + |f_{\rho}(x) - f_{\rho}(z)|)\\
     &\leq 2\exp(-\bmin)\mathbb{E}_x\exp(|f_{\rho}(x) - f_{\rho}(z)|)\\
     &\leq 2\exp(-\bmin)\mathbb{E}_{\xi}\exp(\mathbb{E}_{\rho}|a_w \wpe^T\xi|)\\
     &\leq 2\exp(-\hmin + 2\zeta' H)\mathbb{E}_{\xi}\exp(\mathbb{E}_{\rho}|a_w \wpe^T\xi|) \tagblue{Lemma~\ref{lemma:small_step}}\\
     &\leq 3\exp(-\hmin)\mathbb{E}_{\xi}\exp(\mathbb{E}_{\rho}|a_w \wpe^T\xi|)
\end{align}
Since $a_w \wpe^T\xi$ is subguassian with norm $\Theta(\|a_w \wpe\|)$, we have 
\begin{align}
    \mathbb{E}_{\xi}\exp(\mathbb{E}_{\rho}|a_w \wpe^T\xi|) \leq \exp(\Theta(\mathbb{E}_{\rho}[|a_w\wpe\|^2])) &\leq \exp\left(\Theta\left(\sqrt{\mathbb{E}_{\rho}[|a_w\|^2]}\sqrt{\mathbb{E}_{\rho}[|\wpe\|^2]}\right)\right) \leq \exp(\Theta(H\zeta')),
\end{align}
so 
\begin{align}
    \mathbb{E}_x \ell_{\rho_T}(x) \leq 3\exp(-\hmin) \exp(\Theta(H\zeta)) \leq 4\exp(-\hmin) = o(1).
\end{align}
This yields the theorem.   
\end{proof}

\section{Lower Bound of $\tilde{\Theta}(d)$ for Learning the XOR function with Rotationally Invariant Algorithm.}\label{sec:stat}

\begin{proposition}
Suppose $A : (\{\pm 1\}^d)^n \times \{\pm 1\}^n \times \{\pm 1\}^d \rightarrow \Delta(\{\pm 1\})$ is an algorithm, which given $n$ labeled samples $X = (x_1, \ldots, x_n)$  from the Boolean hypercube, and an additional unlabeled sample, outputs a distribution over labels. Suppose additionally that $A$ is rotationally invariant, that is, for an orthonormal rotation $U$, we have $A(UX, y, Ux) = A(X, y, x)$, so long as $UX$ and $Ux$ are on the hypercube.

Then if $n \leq d/\log^2(d)$, if $X = (x_1, \ldots, x_n)$ and $x$ are sampled uniformly at random from the hypercube, and $i, j$ are sampled uniformly without replacement from $[d]$, we have 
\begin{align}
\mathbb{P}_{X, x, i, j}\mathbb{P}_{\hat{y} \sim A(X, y(X), x)}[\hat{y} \neq y(x)] \geq 0.03,
\end{align}
where $y(x) := -(x^Te_i)(x^Te_j)$, and $y(X) \in \{\pm 1\}^n$ denotes the labels of the entire set $X$.
\end{proposition}
\begin{proof}
Let $H = \{\pm 1\}^d$ denote the Boolean Hypercube. 
Let $Q := \{h \in H: h^Tx_i = x^Tx_i \forall i \in [n]\}$. Because the geometry of the set of points $(x_1, \ldots, x_n, x)$ is the same as the geometry of the points $(x_1, \ldots, x_n, q)$, for each $q \in Q$, for each such $q$, there exists (at least 1) rotation $U$ such that $Ux = q$ and $Ux_i = x_i$ for $i \in [n]$.

Thus by the rotational invariance of $A$, for all $q \in Q$, we must have that $A(X, y, q)$ is the same distribution. Let us denote this distribution by $D_{X, y, Q}$. 

We make the following claim:
\begin{claim}
For some sufficiently small constant $c$, if $|Q| \geq 2^{d - cd}$, we have that with probability at least $0.25$ over $i, j$, for any distribution $D$, $\mathbb{P}_{q \sim Q, \hat{y} \sim D}[y(q) \neq \hat{y}] \geq 0.15$.
\end{claim}
\begin{proof}
Without loss of generality let $i = 1$. It suffices to prove that with probability $0.25$ over $j$, at most a $0.85$ fraction of the points in $Q$ have the same label. 

Suppose this was not true. For $j \in [d]$, let $v_j$ denote the majority of $y(q)$ over $q \in Q$ (break ties arbitrarily), where the labeling function $y$ is determined by $i$ and $j$. Let $P \subset [d]$ be the set of all $j$ such that $\mathbb{P}_{q \sim Q}[y(q) = v_j] > 0.85$, such that we must have $|P| \geq 0.75 d$. To abbreviate, for $j \in P$, define $y_j(q) := -(q^Te_1)(q^Te_j)$. Then for at least a $0.1$ fraction of points in $q \in Q$, we must have $\mathbb{P}_{j \sim P}[y_j(q) = v_j] \geq 0.8$. Indeed, if not, 
\begin{equation}
\mathbb{E}_{j \sim P}\mathbb{E}_{q \sim Q}\mathbf{1}(y_j(q) = v_j) = \mathbb{E}_{q \sim Q}\mathbb{E}_{j \sim P}\mathbf{1}(y_j(q) = v_j) \leq 0.1*1.0 + 0.9*0.8 < 0.85.
\end{equation}
Now we consider how many points $h$ there are in $H$ satisfying $\mathbb{P}_{j \sim P}[y_j(h) = v_j] \geq 0.8$. Consider first the number of such points over $H_1 := \{h \in H: e_1 = 1\}$. Having $\mathbb{P}_{j \sim P}[y_j(h) = v_j] \geq 0.8$ implies that $-h_j = v_j$ for at least $0.8 |P| \geq 0.55 d$ coordinates.  By a standard Chernoff bound, the number of such points is smaller that $2^{d-1}\exp(-Cd)$ for some constant $C$. The same holds for the set of all points where $h^Te_1 = -1$. Thus the total number of points in $H$ satisfying $\mathbb{P}_{j \sim P}[y_j(h) = v_j] \geq 0.8$ is at most $2^{d}\exp(-Cd)$. Thus if $|Q| > \frac{1}{0.1}2^{d}\exp(-Cd)$ we have reached a contradiction. This holds for $|Q| \geq 2^{d - cd}$ for $c$ small enough.
\end{proof}

Now fix $X$ and partition $H$ into $K$ disjoint sets $Q_1, Q_2, \ldots Q_K$ such that that for any $k$, for all $q \in Q_k$, the projection of $q$ onto $X$ is the same. We have
\begin{align}
\mathbb{P}_{x \sim H, i, j}[\hat{y} \neq y(x)] &\geq \sum_{k = 1}^K\frac{|Q_k|}{2^d} \mathbb{P}_{i, j} \mathbb{P}_{q \sim Q_k} [\hat{y} \neq y(q)] \\
&\geq \sum_{k = 1}^K{\frac{|Q_k|}{2^d} (0.25)(0.15) \mathbf{1}(|Q_k| \geq 2^{d(1 - c)})} 
\end{align}
Now since $K \leq (d + 1)^n = 2^{o(d)}$ (indeed, the projection onto each $x_i$ can only take on $d + 1$ different integer values $-2d, -2(d - 1), \ldots, 2d$), we have that
\begin{align}
\sum_{k = 1}^K |Q_k| \mathbf{1}(|Q_k| \geq 2^{d(1 - c)}) \geq 2^d - 2^{d(1 - c)} K \geq (1 - o(1))2^d.
\end{align}
Thus for any $X$, we have that 
\begin{align}
\mathbb{P}_{x \sim H, i, j}[\hat{y} \neq y(x)] \geq (0.25)(0.15)(1 - o(1)) \geq 0.03.
\end{align}
If follows that $\mathbb{P}_{\hat{y} \sim A(X, y(X), x)}[\hat{y} = y(x)] \geq 0.03$.
\end{proof}

\end{document}